\documentclass[11pt,a4paper]{article}
\usepackage{microtype}
\usepackage{graphicx}
\usepackage{subcaption}
\usepackage{booktabs}
\usepackage{hyperref}
\usepackage[left=2.5cm,right=2.5cm]{geometry}

\usepackage[shortlabels]{enumitem}

\usepackage{amsmath}
\usepackage{amssymb}
\usepackage{mathtools}
\usepackage{amsthm}

\usepackage[capitalize,noabbrev]{cleveref}

\theoremstyle{plain}
\newtheorem{theorem}{Theorem}[section]

\newtheorem{lemma}[theorem]{Lemma}

\theoremstyle{definition}
\newtheorem{definition}[theorem]{Definition}

\theoremstyle{remark}
\newtheorem{remark}[theorem]{Remark}
\newtheorem{example}[theorem]{Example}

\crefname{assumption}{Assumption}{Assumptions}
\crefname{equation}{Eq.}{Eqs.}
\crefname{figure}{Fig.}{Figs.}
\crefname{table}{Table}{Tables}
\crefname{section}{Sec.}{Secs.}
\crefname{theorem}{Thm.}{Thms.}
\crefname{lemma}{Lemma}{Lemmas}
\crefname{corollary}{Cor.}{Cors.}
\crefname{example}{Example}{Examples}
\crefname{appendix}{Appendix}{Appendixes}
\crefname{remark}{Remark}{Remark}

\newlist{enumthm}{enumerate}{1} %
\setlist[enumthm]{label=\upshape(\alph*),ref=\upshape\thetheorem(\alph*)}
\crefalias{enumthmi}{theorem} %

\usepackage{natbib}

\usepackage{macro}
\usepackage{xspace}

\newcommand{\Diffy}{DID\xspace}
\newcommand{\name}{\Diffy\xspace}
\title{Measuring dissimilarity with diffeomorphism invariance}
\author{Théophile Cantelobre$^{1,3}$ \and Carlo Ciliberto$^{2}$ \and Benjamin Guedj$^{2,3,4}$ \and Alessandro Rudi$^{1}$}
\date{%
	\small{
	$^1$Inria, École normale supérieure, CNRS, PSL Research University, Paris, France\\%
	$^2$Centre for AI, Department of Computer Science, University College London, London, UK\\%
	$^3$The Inria London Programme, London, UK\\%
	$^4$Inria, Lille - Nord Europe Research Centre, Lille, France\\%
	Correspondance: \texttt{first.last at inria.fr}, \texttt{c.ciliberto at ucl.ac.uk}}%
}

\begin{document}

\maketitle
\begin{abstract}
Measures of similarity (or dissimilarity) are a key ingredient to many machine learning algorithms. We introduce \Diffy, a pairwise dissimilarity measure applicable to a wide range of data spaces, which leverages the data's internal structure to be invariant to diffeomorphisms. We prove that \Diffy enjoys properties which make it relevant for theoretical study and practical use. By representing each datum as a function, \Diffy is defined as the solution to an optimization problem in a Reproducing Kernel Hilbert Space and can be expressed in closed-form. In practice, it can be efficiently approximated via Nyström sampling. Empirical experiments support the merits of \Diffy.
\end{abstract}

\section{Introduction}
One of the overarching goals of most machine learning algorithms is to generalize to unseen data. Ensuring and quantifying generalization is of course challenging, especially in the high-dimensional setting. One way of reducing the hardness of a learning problem is to study the invariances that may exist with respect to the distribution of data, effectively reducing its dimension. Handling invariances in data has attracted considerable attention over time in machine learning and applied mathematics more broadly. Two notable examples are image registration \citep{registration1,registration2} and time series alignment \citep{dtw,soft-dtw,vayer,diff-dtw,align-incomparable}.

In practice, data augmentation is a central tool in the machine learning practitioner's toolbox. In computer vision for instance, images are randomly cropped, color spaces are changed, artifacts are added. Such heuristics enforce some form of invariance to transformations that are chosen by hand, often parametrically, and comes at the cost of a more demanding learning step (\emph{e.g.}, more data to store and process, more epochs, to name but a few).

Learning under invariances has also spurred significant theoretical interest. 
\citet{decoste2002training}, \citet{haasdonk2007invariant}, \citet{kondor2008group} and \citet{mroueh2015} algebraically studied how kernels invariant to group actions behave for learning. \citet{bruna2013invariant} takes inspiration in signal processing (with wavelet spaces) to build scattering networks, which can present good properties with respect to invariances. Focusing on neural networks, \citet{CKN} and \citet{CKNInvariance} introduced and analyzed a model aiming to mimic Convolutional Neural Networks (CNNs). More recently, \citet{bietti2021} studied the sample complexity of learning in the presence of invariances, with invariant kernels. Conversely, the hypothesis that invariance can be a proxy for neural network performance has been put to test empirically by \citet{wyartdiffeo}.

\paragraph{Contributions.}
We introduce a dissimilarity called \Diffy (standing for Diffeomorphism Invariant Dissimilarity) which is invariant to smooth diffeomorphisms present between data points. Although \Diffy is somewhat less sophisticated than most of the models presented above, it is considerably more generic and can be seen as a building block for devising practical machine learning algorithms in the presence of invariances.

In order to exploit the internal structure of a data point (\emph{e.g.} of an image), we cast them as functions between two spaces (\emph{e.g.} coordinate and color spaces). This unlocks the potential of using the change of variable formula to eliminate diffeomorphisms between functions (\emph{i.e.} transformations between data points) when they exist. \Diffy is based on a generic method for identifying if a function $g$ is of the form $f\circ Q$, without any parametric model over $Q$, founded on the change of variable formula. This makes \Diffy a promising and flexible tool for image processing (\emph{e.g.}, image registration), time-series analysis (\emph{e.g.}, dynamic time warping) and machine learning (\emph{e.g.}, nearest neighbors).

\Diffy is defined as an optimisation problem in a Reproducing Kernel Hilbert Space (RKHS), of which we present a closed form solution (\Cref{theorem:closed-form}). We then show how it can be approximated in practice, using Nyström sampling techniques (\Cref{lm:widehatD}, \cref{thm:appr-error-widehatD}). By relying on standard matrix linear algebra, this approximation can be efficiently implemented with batch techniques, and accelerated hardware.

A key aspect is that \Diffy has very few ``hyper-parameters'' which can easily be chosen by a domain expert: the kernels on the input and output space and a regularization parameter. We provide guidance on choosing the regularization parameter in \cref{sec:approximation}.

\begin{figure*}[ht!]
\centering
    \begin{subfigure}{0.30\textwidth}
\includegraphics[width=\textwidth]{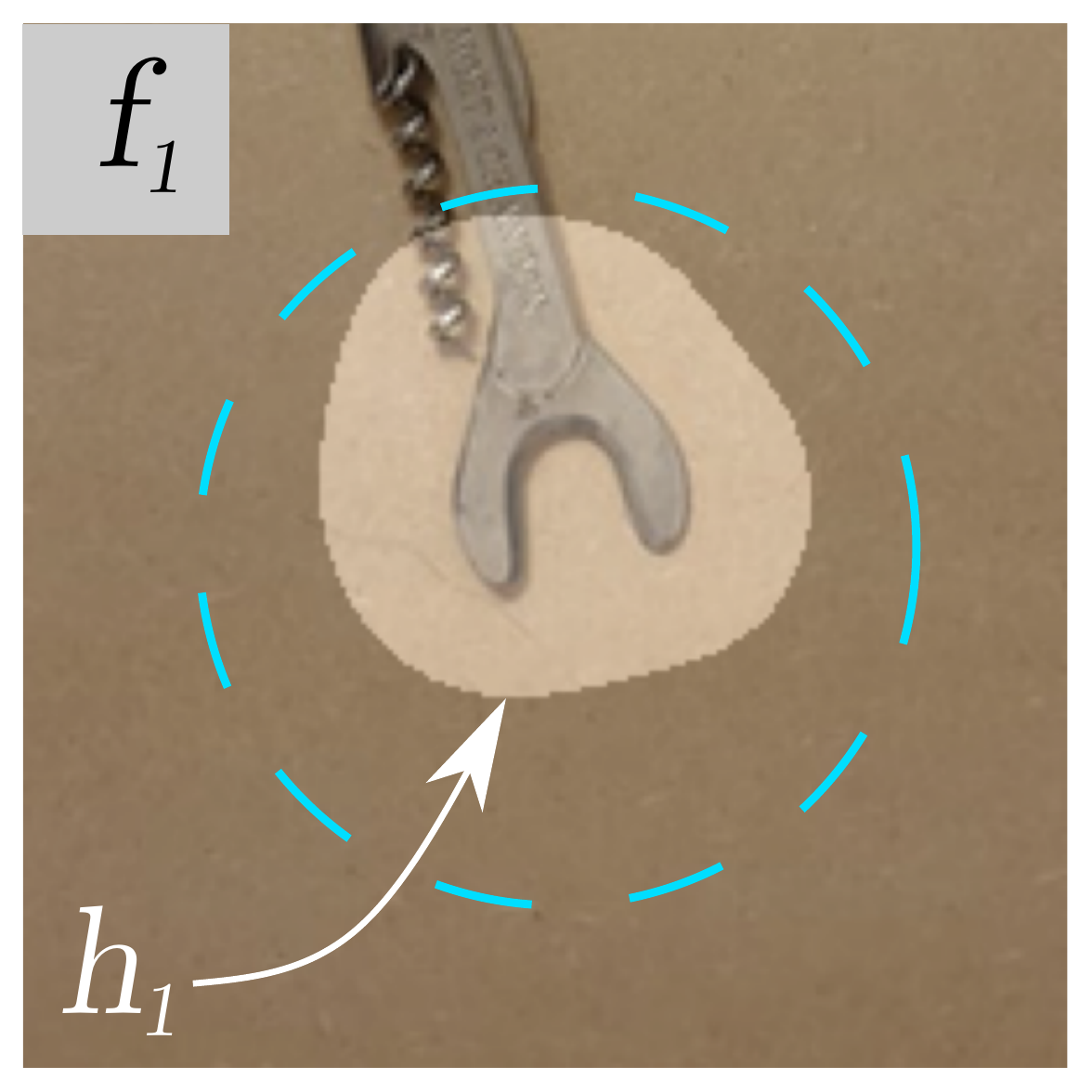}
\caption{$f_1$ (close-up)}
\label{fig:sceneh}
    \end{subfigure}\hfil
    \begin{subfigure}{0.30\textwidth}
\includegraphics[width=\textwidth]{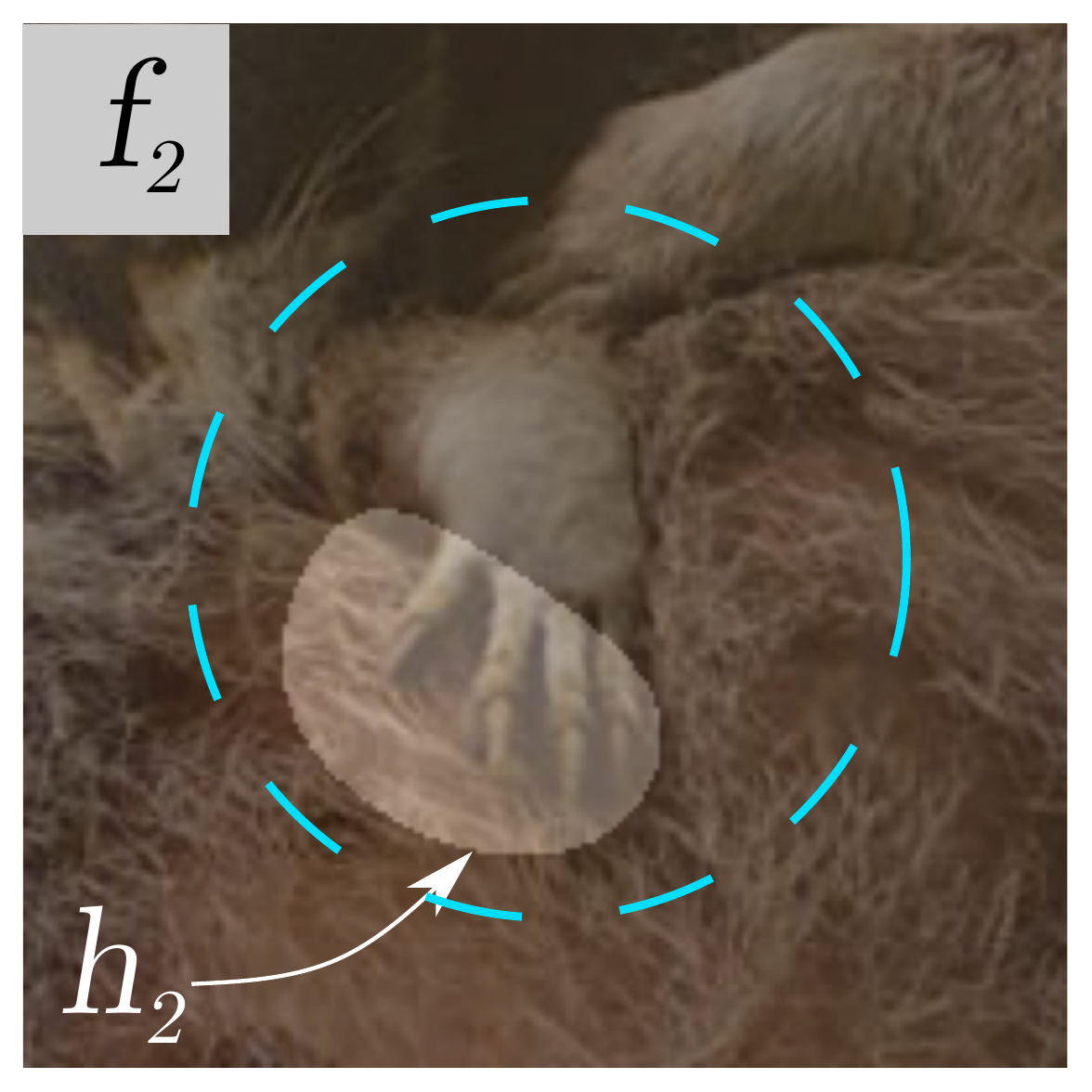}
\caption{$f_2$ (raccoon's paw)}
\label{fig:raccoonh}    \end{subfigure}\hfil
     \begin{subfigure}{0.30\textwidth}
\includegraphics[width=\textwidth]{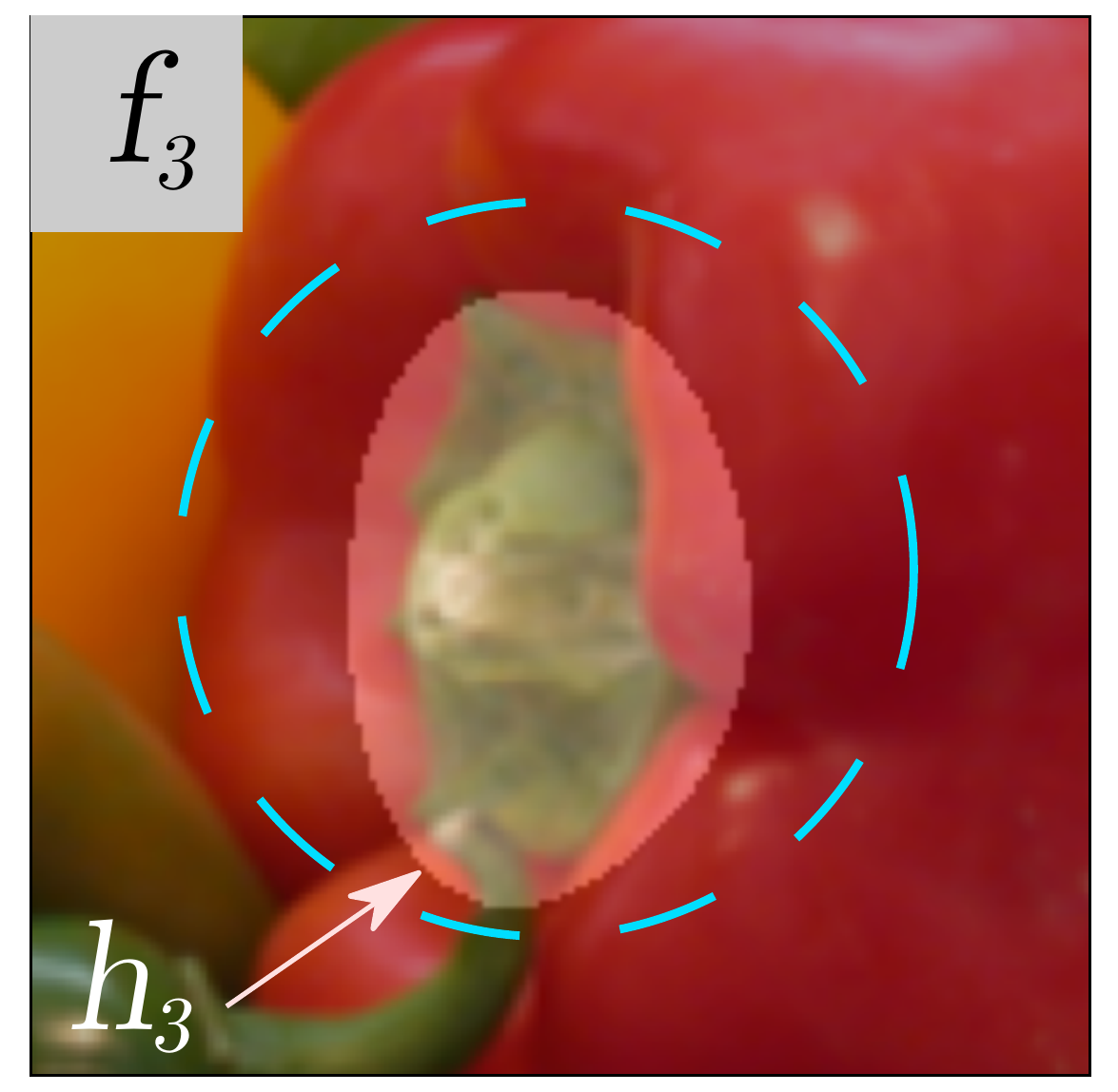}
\caption{$f_3$ (pepper close-up)}
\label{fig:flowersh}
    \end{subfigure}
\smallskip
     \begin{subfigure}{0.30\textwidth}
\includegraphics[width=\textwidth]{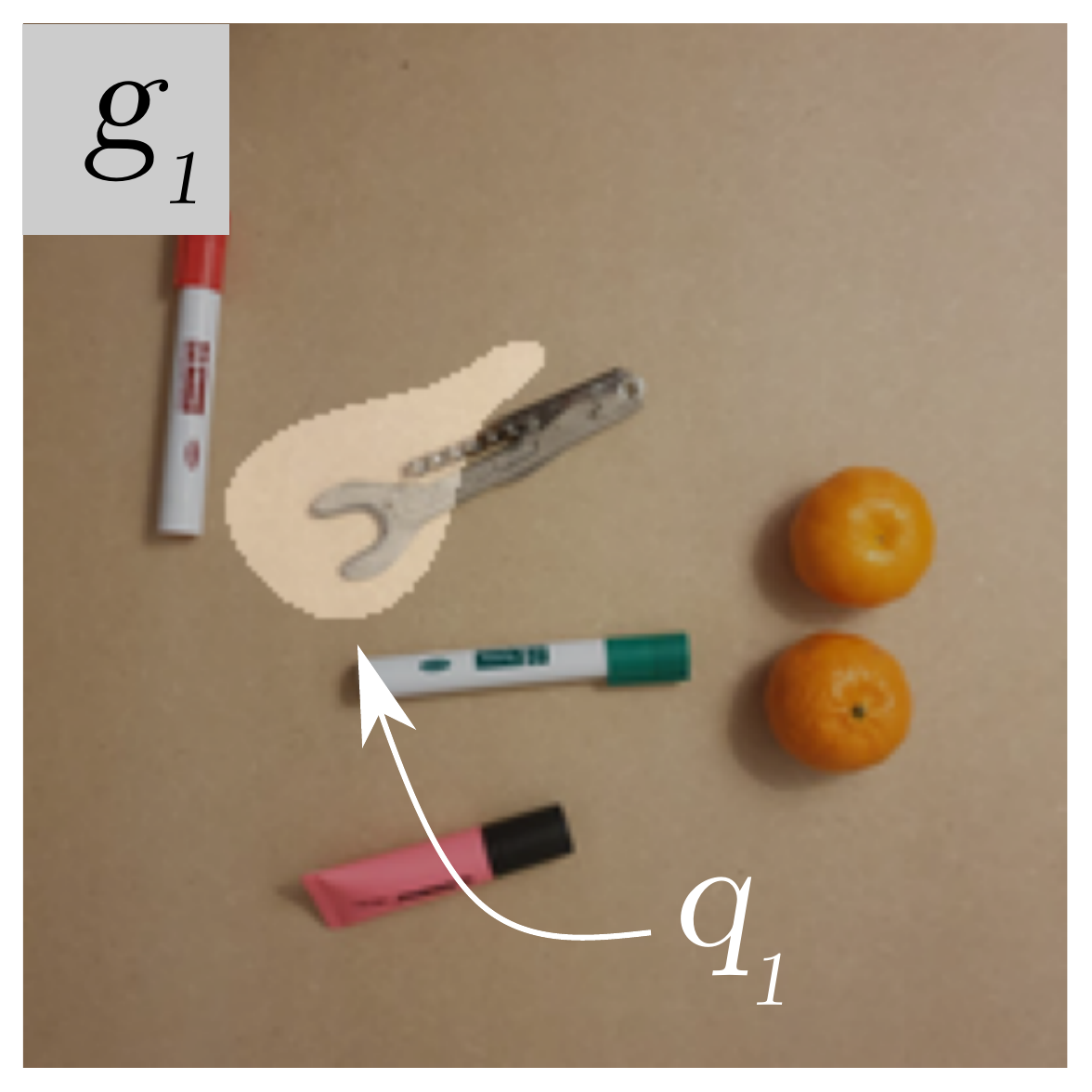}
\caption{$g_1$ (scene with multiple objects)}
\label{fig:sceneq}
    \end{subfigure}\hfil
     \begin{subfigure}{0.30\textwidth}
\includegraphics[width=\textwidth]{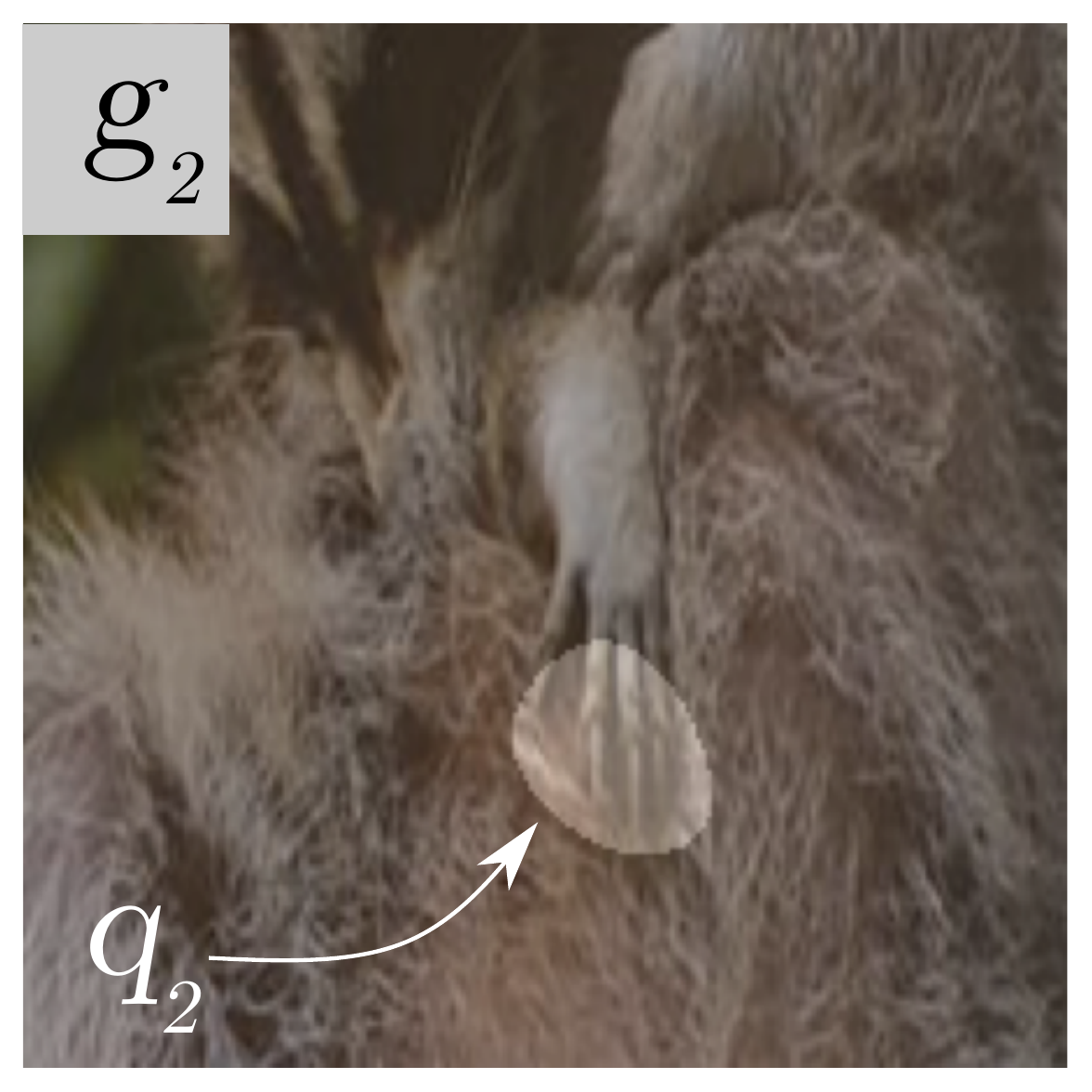}
\caption{$g_2$ (raccoon's abdomen, with natural keystone deformation.)}
\label{fig:raccoonq}
    \end{subfigure}\hfil
     \begin{subfigure}{0.30\textwidth}
\includegraphics[width=\textwidth]{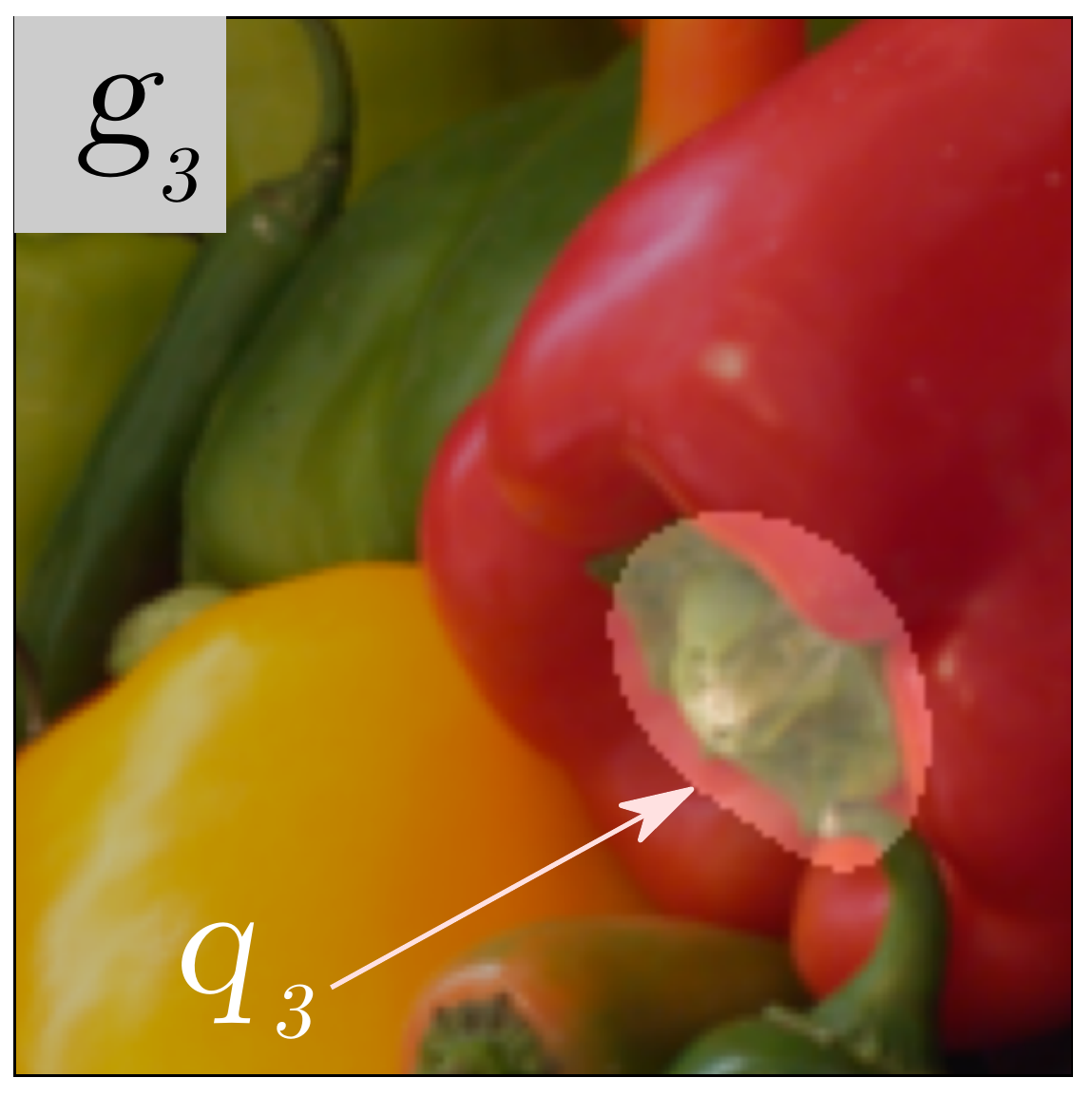}
\caption{$g_3$ (peppers among vegetables)}
\label{fig:flowersq}
    \end{subfigure}
\caption{\emph{Illustration of $\widehat D_\lambda$ on images.} We compute $\widehat D_\lambda(f_i, g_i)$ and materialize the optimal $h_i$ and $q_i$ selected by \Diffy (with thresholding for visualisation purposes), for $f_i$ and $g_i$ images taken with a smartphone. The mask $\mu$ is illustrated by the dashed circle (in light blue), see \cref{sec:details} for details. The images are taken from different views so as to provide different angles and lighting. \cref{fig:sceneh,fig:sceneq} show a scene of \texttt{objects} and a close-up on one of them (bottle opener). \cref{fig:raccoonh,fig:raccoonq} are taken from images of a raccoon (\texttt{raccoon}). \cref{fig:flowersh,fig:flowersq} are sub-patches from \texttt{peppers}, a scene with vegetables. Notice that $q_i$ visually matches the area highlighted by $h_i$, despite the perspective and scale changes. Additional details are gathered in \Cref{sec:illustration}.}
\label{fig:example}
    \end{figure*}

Using tools from functional analysis, we prove that \Diffy behaves as expected when comparing $f$ and $f\circ Q$ (\emph{i.e.}, that it considers these two functions to be close) in the limit of vanishing regularization (\cref{theorem:main-theorem}). We support our theoretical claims with numerical experiments on images.

\paragraph{Outline.} We introduce our new dissimilarity in \Cref{sec:dissimilarity}. In \cref{sec:robustness}, we prove that the intuition leading to the definition in \cref{sec:dissimilarity} is well founded and discuss the theoretical properties of the dissimilarity. We then show how to compute the dissimilarity in \cref{sec:computing}: we first show that it has a closed-form expression; we then present and justify an approximation scheme based on Nyström sampling. We illustrate the behavior of the dissimilarity with experiments on images in \cref{sec:experiments}.

\section{The dissimilarity}\label{sec:dissimilarity}

\subsection{Informal derivation of the dissimilarity} \label{sec:informal}

The dissimilarity we describe in this work relies on the internal structure of the objects it compares. An efficient way of encoding this structure is to, whenever possible, view objects as maps between an input space and an output space. In this way, we can consider both the values taken by the function and the locations at which these values were taken.

Consider $f$ and $g$ two maps between $\R^d$ and $\R^p$. Our goal is to determine whether there exists a diffeomorphism $Q:\R^d \to \R^d$ such that $g = f \circ Q$. In practice, such transformations could be rigid-body transformations, a non-singular projective transformation or more generally a mild distorsion of the space (such as warping). The goal is thus to find a measure of dissimilarity between $f$ and $g$ that is robust to such diffeomorphisms. We derive such a measure informally in this section around three key ideas.

\paragraph{Change of variable formula.} Integrals offer a natural way to ``eliminate'' a diffeomorphism from a function, via the change of variable formula
\eqals{
\int f(Q(x)) |\nabla Q(x)| \dd x = \int f(x) \dd x. 
}
As $|\nabla Q|$ (the determinant of the Jacobian of $Q$) is unknown, we can approximate the above formula with:
\eqal{\label{eq:prob}
\min_{q} \left|\int g(x) q(x) \dd x - \int f(x) \dd x\right|,
} where $q$ lies in a space of functions.

Indeed, when $g = f\circ Q$ for some diffeomorphism $Q$, choosing $q = \vert\nabla Q\vert$ minimizes \eqref{eq:prob}. However, this solution is not unique. Indeed there exist trivial solutions as $q = f / g$, that are irrespective of the existence of a $Q$ such that $g=f\circ Q$ or not.

\textbf{Range of statistics.} One way of reducing the class of solutions to ones that are relevant to our original question is to study not only how well the weighted integral of $g$ can approximate the integral of $f$, but also require that the same weight approximate a wide class of transformations of $f$. A natural example with inspiration in probability theory is to be able to approximate all moments of $f$, \emph{i.e.}, the integrals of the moments $v_1(f) = f, v_2(f) = f^2, \dots$, or more general statistics. The function $q = f / g$ may match the integral of $v_1(g)$ with that of $v_1(f)$, but cannot work for $v_2$. However, if $g = f\circ Q$, $q = |\nabla Q(x)|$ (the solution we are seeking), satisfies that for any continuous function $v: \R^p \to \R$,
\eqals{
\int v(f(Q(x))) |\nabla Q(x)| \dd x = \int v(f(x)) \dd x.
}
Problem \eqref{eq:prob} is thus replaced by the following one, which also has $q$ as a solution when $g = f \circ Q$:
\eqal{\label{eq:prob2}
\min_q \max_{v \in V} \left|\int v(g(x)) q(x) \dd x - \int v(f(x)) \dd x\right|,
} where $V$ is a rich set of statistics, \emph{e.g.}, continuous integrable functions on $\R^p$.

\textbf{Uniformity over regions.} Problem \eqref{eq:prob2} averages the statistics uniformly over the whole space irrespectively of the fact that a witness of $g \neq f \circ Q$ could live in a lower dimensional region. For instance $g$ and $f$ might be non-zero only on a small region of the space, consequently yielding to a relatively small value for \cref{eq:prob2}. To enhance such regions, we choose to integrate with respect to a smooth function $h$, that is chosen adversarially to maximize the dissimilarity between $f$ and $g$.  %
In other words, we arrive at the following optimization problem:
\eqals{
\max_{h \in \hh_1} \min_{q} \max_{v \in V} \left|\int v(g(x)) q(x) \dd x - \int v(f(x)) h(x) \dd x\right|, 
}
where $\hh_1$ is a suitable set of smooth functions.
Again, if $g = f\circ Q$, the above is solved by $q \,=\, h\circ Q \, |\nabla Q|$.

Note that the smoothness of $h$ and $q$ is crucial. On the one hand, the smoothness of $h$ ensures that the considered regions are of interest with respect to the underlying metric on $\R^d$, i.e. they cannot be too close to diracs on pathological sets. On the other, the smoothness of $q$ ensures that the transformations are not matched by ``cherry-picking'' dispersed points on the domain such that the integrals match.

\subsection{Definition of the dissimilarity}
Now that we have given the motivation as well as the intuition behind our method, we can formally introduce it.

Let $X \subseteq \R^p$ and $Y \subseteq \R^p$ where $d, p \geq 1$. In this paper, the objects of interest are maps from $X$ to $Y$. Note that this is quite flexible and can reflect many of the rich types of data considered in image processing, time-series modelling and machine learning. Indeed, an image can be seen as a map from $\R^2$ (the coordinates of the pixels) to $\R^3$ (the color space, RGB for instance). A time-series can be seen as a map from $\R^1$ (time) to $\R^n$ (the space of the sample values).

Let $k_X$ be a reproducing kernel on $X$ with Reproducing Kernel Hilbert Space (RKHS) $\mathcal H$  \cite{aronszajn1950theory} and $k_Y$ be a reproducing kernel on $Y$ with RKHS $\mathcal F$. In particular, we assume that they are bounded.

To conclude, the formalization below allows naturally for the presence of a mask function $\mu$, with the role of focusing the matching process on a subregion of interest of $f$. The role of the mask will be discussed after the definition.

\begin{definition}[Dissimilarity $D_\lambda(f, g)$]\label{def:D}
    Let $\la \geq 0$ and bounded integrable $\mu:\X \to \R$.  For any $f,g:X\to Y$, we define the dissimilarity
\eqal{\label{eq:def-D}
    D_\lambda(f, g) := \max_{\norh{h}\leq 1}\min_{q \in \mathcal H} \Delta_{f,g}(h, q) + \lambda \norh{q}^2
}
where $\Delta_{f, g}(h, q)$ is defined as follows,
\begin{align*}
\Delta_{f, g}(h, q) := \max_{\|v\|_{{\cal F}} \leq 1} &\Big|  \int_X v(g(x))q(x)\dd x\\ 
 & -  \int_X v(f(x))\mu(x)h(x)\dd x\Big|^2.
\end{align*}
\end{definition}

When clear from context, we write $D$ instead of $D_\la$ for the sake of conciseness.

The reader should easily recognize the construction described in \cref{sec:informal}. We have made the space in which we search for $q$ and $h$ precise: $\hh$, the RHKS of $k_X$. Regularity of $h$ is enforced by searching in the unit ball, while regularity of $q$ is enforced with Tikhonov regularization. This choice allows at the same time to compute the dissimilarity in closed form (see \cref{sec:computing}), while not sacrificing its expressivity (see \cref{sec:approximation}). In particular, to allow a very rich set of statistics that is also manageable from a computational viewpoint, we choose $V$ to be the unit ball of ${\cal F}$. For example, if $Y$ is a bounded subset of $\R^p$ and $k_Y$ is chosen as the Laplace kernel $k_Y(y,y') = \exp(-\|y-y'\|)$, then a rescaled version of any infinitely smooth function belongs to $V$ (including in particular all polynomials, smooth probabilities, Fourier basis -- see \cref{app:background} for more details). For the same reason we choose $\hh_1$ to be the unit ball of $\hh$ and we choose also $q \in \hh$.
\cref{fig:example} shows a few examples of the role played by the $h$ and $q$ optimizing the problem in \cref{eq:def-D}.

\begin{figure*}
    \centering
    \begin{subfigure}[t]{0.48\textwidth}
        \includegraphics[width=\textwidth]{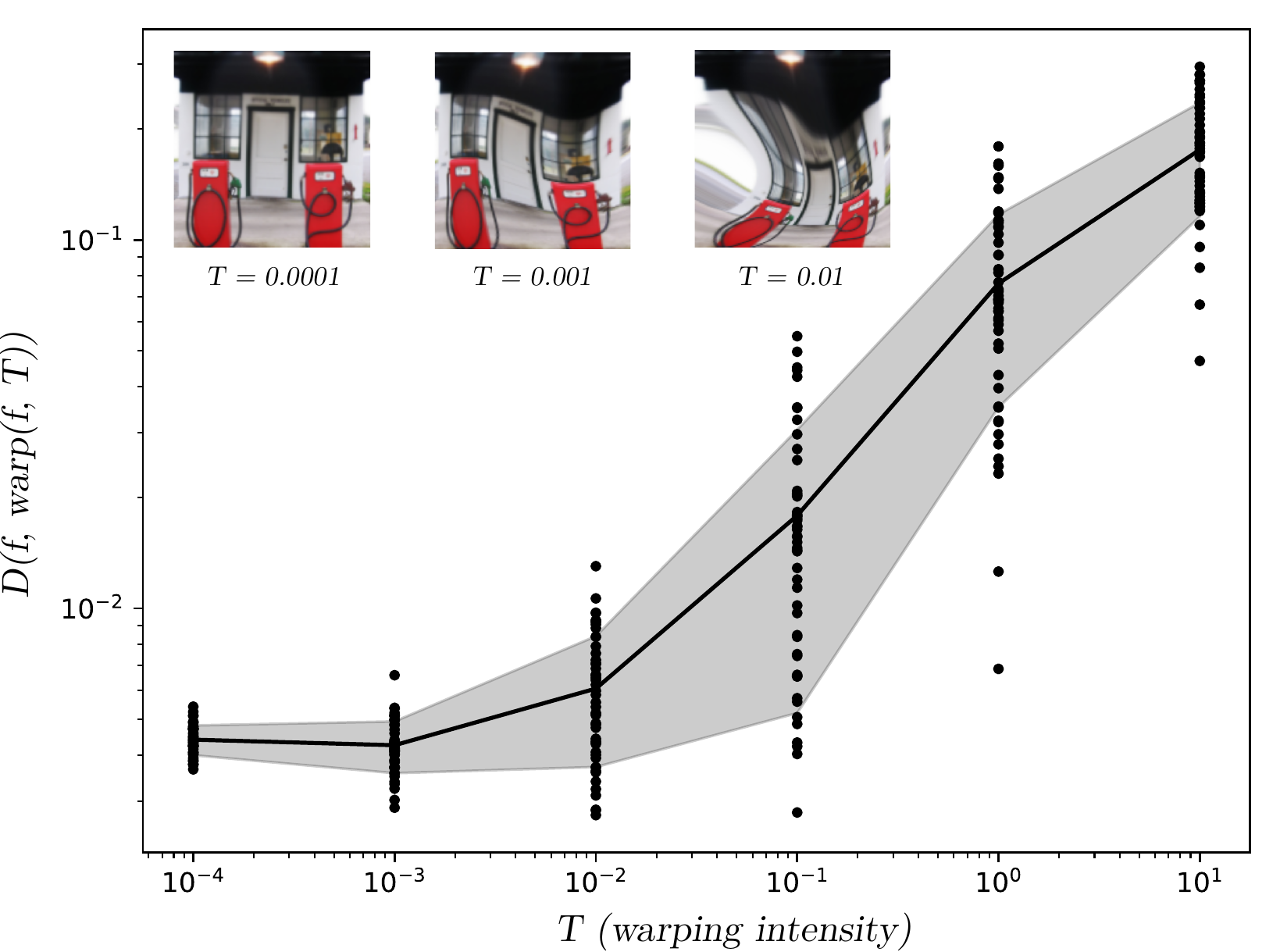}
        \label{fig:warping-left}
        \caption{\emph{$\widehat D_\lambda(f, \text{warp}(f, T))$ as a function of $T$.} We repeat the warp $50$ times ($+$ markers) on the same image and represent average values $\pm$ standard deviation (in grey).}
    \end{subfigure}
    \begin{subfigure}[t]{0.48\textwidth}
        \includegraphics[width=\textwidth]{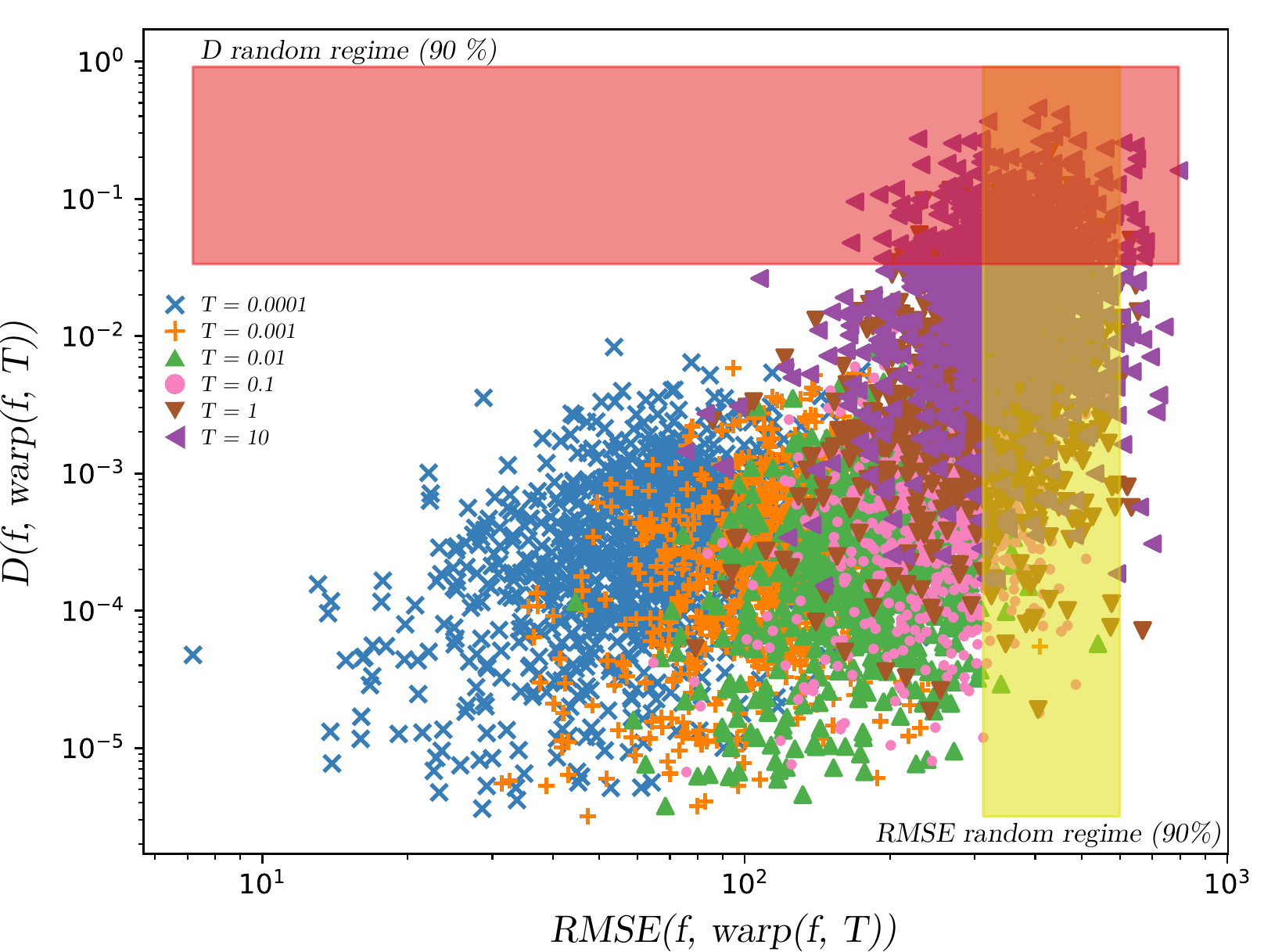}
        \caption{\emph{$\textrm{RMSE}(f, \textrm{warp}(f, T))$ against $\widehat D_\lambda(f, \textrm{warp}(f, T))$ for various values of $T$ and images $f$.} $1000$ images are each warped once for each value of $T$. We represent the random regime for each metric.} 
    \end{subfigure}
    \caption{\emph{Invariance to general diffeomorphisms (warping).} Warping is randomly generated, its intensity controlled by a temperature parameter $T$ (higher $T$ produces, on average, warps with higher displacement norm). In (a): $\widehat D_\la$ stays constant (i.e. invariant to the warps) as long as their norm is not too strong (small T), while RMSE increases exponentially. When $T$ becomes large the transformations become intense (indeed they are non-diffeomorphic) and $\widehat D_\la$ grows to reflect this fact.  In (b): we see that $\widehat D_\la$ stays invariant to warps as long as $T \leq 0.1$ (far from the random regime interval), while the Euclidean distance increases exponentially with $T$, even for small $T$. See \cref{sec:invariance-warping} for more details.}
    \label{fig:warping}
\end{figure*}

\paragraph{The role of the mask $\mu$.} We introduced a function $\mu:X \to \R$ which applies to the term depending on $f$ and $h$. This function is meant to be a mask which focuses the distance on a certain region of $f$, discounting other regions. Such presence is useful in practice, since typically the space $X$ is given by the problem. For example, if we want to use the dissimilarity to check if the content of a given image ($f$) is contained in an image ($g$), the shape $X$ is typically rectangular, while the region of interest is the interior of the image. In this case the mask is useful to avoid the artifacts introduced by the corners. Notice that this addition further breaks the symmetry between $f$ and $g$: $f$ becomes a \emph{reference}, and we search in $g$ for matching statistics. In the experiments, we relied for example on a Blackman Window, a classical window function in signal and image processing (see \cref{sec:details}) to reduce the impact of the corners.

\section{Robustness to diffeomorphisms}\label{sec:robustness}

The dissimilarity $D$ is designed (consistently with the derivation in \cref{sec:informal}) to be small when $f$ and $g$ are equal up to a diffeomorphism. The ideal result would be something along the lines of the following \emph{informal} theorem:

\begin{theorem}[Ideal]\label{thm:ideal}
    For any $f: X \to Y$ and any $Q$ diffeomorphism over $X$, $D(f, f\circ Q) \approx 0.$ 
\end{theorem}

This is of course too much to ask. Indeed, the regularization over the choice of $q$ (which in turn controls the regularity of the jacobian of a hypothetical $Q$) introduces a bias:  even if $g = f \circ Q$, the dissimilarity is not $0$. This bias vanishes if and only if $q^*=0$ (by definiteness of the norm).

However, when the RKHS is assumed to be rich enough and $Q$ and $\mu$ are regular enough, then we have the following result. Before stating it, we recall that the Laplace kernel $k_X(x,x') = \exp(-\|x-x'\|)$ belongs to the more general family of Sobolev kernels \cite{wendland2004scattered}. In particular, it corresponds to the Sobolev kernel of smoothness $m$, with $m = (d+1)/2$, where $d$ is the dimension of $X \subset \R^d$. 
\begin{theorem}\label{theorem:main-theorem}
Let $X \subset \R^d$ be an open bounded set with Lipschitz boundary. Let $\mu \in C^\infty(\R^d)$ with compact support $\Omega \subset X$. Choose $k_X$ to be a Sobolev kernel of smoothness $m$, with $m > d/2$. Then for any $C^{m+1}$ diffeomorphism $Q$ on $\R^d$ satisfying $Q^{-1}(\Omega) \subset X$, we have that
\eqals{
 D_\la(f, f \circ Q) ~~\leq~~ \lambda ~ C^2_{\mu} C^2_{Q} \qquad \forall f\, \textrm{measurable}, 
}
where $C_\mu = \|\mu\|_{H^m(\R^d)}$, and $C_Q$ is defined in \cref{eq:def-CQ} in \cref{sec:general-theorem} and depends only on $\Omega, X, Q, d, m$.
\end{theorem}
The theorem above is a special case of \cref{thm:general-theorem}, presented in \cref{sec:general-theorem}. There we prove the more general result: $D_\la(f,g) \leq \lambda C^2_{\mu} C^2_{Q}$, for all measurable $f, g$ that satisfy $g(x) = (f \circ Q)(x)$, only in the region not canceled by the mask, \emph{i.e.}, $\forall x \in Q^{-1}(\Omega)$.

A first consequence of \cref{theorem:main-theorem} is that the regularization parameter $\lambda$ controls the threshold to decide whenever $g \approx f \circ Q$. In particular, we easily see that $D_\la(f, f \circ Q) \to 0$, when $\la \to 0$. This result confirms that in the limit where the regularization vanishes, so does the bias and we have the result we would have imagined à la \cref{thm:ideal}. We will see in \cref{sec:approximation} that $\la$ has also an important role in controlling the approximation error of $D(f,g)$. This shows that $\la$ controls a similar bias-variance trade-off as in classical kernel supervised learning \cite{shawe-taylor2004}. 

To show concretely the dependence of $C_Q$ with respect to $Q$, in the following example we show $C_Q$ explicitly for an interesting class of diffeomorphisms.

\begin{example}[Magnitude of $C_Q$ for rigid transforms]\label{ex:diffeo}
Let $X$ be the unit ball in $\R^d$ and let $\mu$ be a mask supported on $\Omega$, the ball of radius $r < 1$. We consider the diffeomorphisms $Q(x) = \alpha R x$, with $R$ a unitary matrix and $r < \alpha < 1/r$. We use the Laplace kernel $k_X(x,x') = \exp(-\|x-x'\|)$ for $k_X$ (analogously for $k_Y$).
Then, we compute explicitly the bound in \cref{eq:def-CQ}, since the Laplace kernel corresponds to Sobolev kernel with exponent $m = (d+1)/2$, obtaining 
$$C_Q ~~\leq~~ C_0 \left(\frac{\alpha}{\min(\alpha,1) - r}\right)^{m+d/2}\alpha^d (1 + \alpha + \alpha^m).$$
\end{example}
\subsection{Discussion on the discriminatory power of $D_\la$} \label{sec:discussion-selectivity}
In \cref{theorem:main-theorem}, we proved that when $g = f \circ Q$ for some diffeomorphism $Q$, then $D(f,g)$ is small, i.e. that the dissimilarity is essentially invariant to the diffeomorphisms. However, to fully characterize the properties of the proposed dissimilarity it would be interesting to study also its discriminatory power, i.e. the fact that $D(f,g)$ is small {\em only if} there exists a diffeomorphism $Q$ such that $g = f \circ Q$. \cref{fig:warping} investigates this question from the empirical perspective. The details of these experiments are reported in \cref{sec:experiments} (and further explored in \cref{app:additional-experiments}). They show that, \name is very robust to significant transformations $f\circ Q$ of the original signal $f$. Additionally we observe that $D_\la$ is very discriminative, in contrat to less diffeomorphism invariant metrics such as the euclidean distance, when comparing $D_\la(f,f\circ Q)$ with $D_\la(f,g)$ for a random signal $g$. We care to point out however, that the theoretical analysis of \name's the discriminative abilities is beyond the scope of this work (whose aim is to introduce the discrepancy and study its invariance properties) and we postpone it to future research.

\section{Computing the dissimilarity}\label{sec:computing}

Before deriving the closed form solution for $D_\la$, we need to recall some basic properties of kernels. Reproducing kernels and RKHSs satisfy the so called {\em reproducing property}, i.e. There exists a map $\psi:X \to \hh$ such that, for any $f \in \hh$ and $x \in X$, it holds that $f(x) = \scal{f}{\psi(x)}_\hh$, where $\scal{\cdot}{\cdot}_\hh$ is the scalar product associated to the RKHS $\hh$. Moreover $k_X(x,x') = \scal{\psi(x)}{\psi(x')}_{\hh}$, for all $x,x' \in X$. The same holds for $k_Y$ and $\cal F$. i.e., there exists $\Phi:Y \to {\cal F}$ such that $v(y) = \scal{v}{\Phi(y)}_{\cal F}$ for all $v \in {\cal F}, y \in Y$ and $k_Y(y,y') = \scal{\Phi(y)}{\Phi(y')}$ for all $y,y' \in Y$, where $\scal{\cdot}{\cdot}_{\cal F}$ is the inner product associated to ${\cal F}$. In particular, note that, since we assumed that $k_X, k_Y$ are bounded kernels, then there exist two constants $\kappa_X, \kappa_Y$ such that $\sup_{x \in X} \|\psi(x)\|_{\hh} \leq \kappa_X$ and analogously $\sup_{y \in Y} \|\Phi(y)\|_{\cal F} \leq \kappa_Y$.

\subsection{Closed form solution}
In \cref{def:D}, we define $D_\lambda(f,g)$ as an optimization problem in $\hh \times \hh$. In fact, this optimization problem has a closed-form solution, as the solution of an eigenvalue problem of an operator between $\hh$ and $\mathcal F$ as derived in \cref{eq:cl-form-eig-Dla}. We introduce the relevant objects and prove \cref{theorem:closed-form}.

\begin{definition}[Operators $F_\mu$, $G$]
Given $f, g: X \to Y$, the feature map $\Phi: Y\to \mathcal F$ and the mask function $\mu:X \to \R$, define the linear operators $F_\mu, G:\hh \to {\cal F}$ as follows:
\eqal{
F_\mu &= \int_X \Phi(f(x)) \otimes \psi(x) \mu(x) \dd x, \\ G~&= \int_X \Phi(g(x)) \otimes \psi(x) \dd x.
}
\end{definition}

Now, when $X$ is a bounded set, the two operators above are trace class and, by the representer property  $\scal{v}{F_\mu h}_{\cal F} = \int_X v(f(x)) h(x) \mu(x) dx$ and also $\scal{v}{G q}_{\cal F} = \int_X v(f(x)) q(x) dx$ (see \cref{lm:explicit-Fmu-G} in \cref{sec:additional-proofs} for a detailed proof). Using this result and considering the linearity of $\scal{\cdot}{\cdot}_{\cal F}$ and the variational characterization of the Hilbert norm (i.e. $\|u\|_{\cal F} = \max_{\|v\|_{\cal F} \leq 1} |\scal{v}{u}_{\cal F}|$), we have
\eqals{
\Delta_{f,g}(h,q) & = \|F_\mu h - G q\|_{\cal F}^2,
}
for any $h, q \in \hh$. From which we characterize $D_\lambda$ as
\eqal{
D_\lambda(f, g) = \max_{\norh{h}\leq 1} \min_{q\in\hh} \norf{F_\mu h - G q}^2 + \lambda \norh{q}^2.
}
To conclude, note that the optimization problem in the equation above has a closed-form expression in terms of the operatorial norm of an operator depending on $F_\mu$ and $G$. All the reasoning above is formalized below. 
\begin{theorem}[Closed-form solution]\label{theorem:closed-form}
Let $X \subset \R^d$ be an open bounded set. Using the notations above, we have:
\eqal{\label{eq:cl-form-eig-Dla}
D_\lambda(f, g) = \lambda\norop{(GG^* + \lambda I)^{-1/2}F_\mu}^2.
}
\end{theorem}
The proof of \cref{theorem:closed-form} comes from identifying the inner optimization problem as a linear regression problem, and the the outer maximization problem as an eigenproblem. The complete proof is presented in \cref{sec:proof-closed-form}.

\begin{figure}[t]
    \centering
    \includegraphics[width=0.45\textwidth]{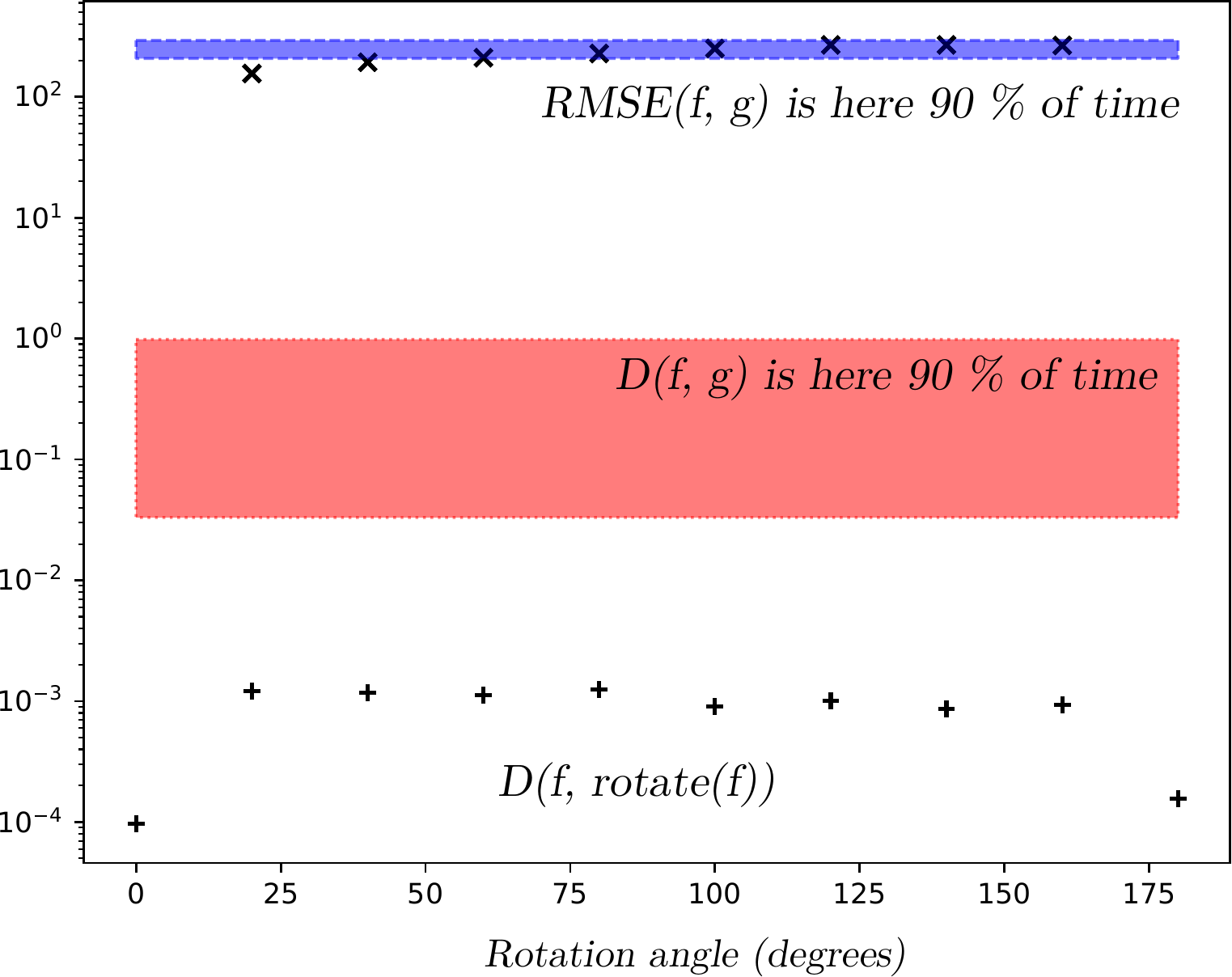}
    \caption{\emph{Invariance to rotation.} We consider a patch $f$ (size $100\times 100$) of a larger scene (\texttt{peppers.jpeg}) and compare it to its rotated versions $\text{rotate}(f, \alpha)$, where $\alpha$ is an angle. $D(f, \text{rotate}(f, \alpha))$ is represented with $+$ symbols and $\textrm{RMSE}(f, \text{rotate}(f, \alpha))$ with $\times$ symbols. Random regimes are represented by shaded areas (see text). Both $\widehat D_\la$ and the RMSE seem constant as a function of $\alpha$ (although with $\alpha=0$ or $\alpha=180$ a smaller value is achieved). However, the RMSE of the rotated patches falls in (or close to) the confidence interval, making them indistinguishable from random patches from the same image. $\widehat D_\la$ takes values that are over $10 \times$ smaller for rotated patches that for random patches. Hence, \Diffy is invariant to rotation, whereas RMSE is constant (for $\alpha > 0$). Here $\la = 10^{-6}$. See \cref{sec:invariance-rotation} for more details.}
    \label{fig:rotation}
\end{figure}

\subsection{Approximate computation}\label{sec:approximation}

Although \cref{theorem:closed-form} gives a closed-form expression of $D_\lambda(f,g)$, $F_\mu$ and $G$ are defined as integral operators between infinite dimensional Hilbert spaces. In practice, we only have access to a discretization of $f$ and $g$ (e.g. an image is a discretized spatial signal represented by $N$ pixels). A first natural approximation is thus to replace the integral with an empirical counterpart. This estimate is then a sum of rank-one operators between $\mathcal H$ and $\mathcal F$. To reduce the computational cost, while keeping good accuracy, we can then further approximate it using Nyström methods for kernels. The resulting estimator is $\widehat{D}_\la$ presented in \cref{eq:widehatD-finite-dimensional}, its convergence to $D_\la$ is studied in \cref{thm:appr-error-widehatD}.

\paragraph{Quadrature approximation.}

We replace $F_\mu$ with an estimator $F_{\mu, N}$ and $G$ with an estimator $G_N$:
\eqal{
F_{\mu, N} &= \frac{v_X}{N}\sum_{i=1}^N \Phi(f(x_i)) \otimes \psi(x_i) \mu(x_i)\label{eq:F-sum}\\
G_{N} &= \frac{v_X}{N}\sum_{i=1}^N \Phi(g(x_i)) \otimes \psi(x_i),\label{eq:G-sum}
}
where $v_X := \int_X \dd x$ is the volume of the domain $X$.
Note that the set of $\lbrace x_1, \ldots, x_N \rbrace$ can be chosen at random or arbitrarily to best approximate the integrals. $F$ and $G$ can be approximated using different points. In practice, they are often given as the positions of pixels of images, sample times of a time series.

\paragraph{Nyström approximation.}

From the previous section, it is clear that $\rank{F_{\mu, N}} \leq N$ and $\rank{G_N} \leq N$. This justifies using the low-rank approximations we introduce in this section. It is possible to further reduce the rank of the matrices, while keeping a good accuracy, by using the so-called {\em Nystr\"om approximation} \cite{williams2001using,drineas2005nystrom,rudi2015less}. 
Let $M_X, M_Y \in \N$ and choose the set of points $\widetilde{X} = \lbrace \widetilde x_1, \ldots, \widetilde x_{M_X}\rbrace \subset X$ and $\widetilde{Y} = \lbrace \widetilde y_1, \ldots, \widetilde y_{M_Y} \} \subset Y$. The Nystr\"om approximation of a vector $v \in \hh$ consists in the projected vector $P_{\tilde{X}} v$, where $P_{\tilde{X}} :\hh \to \hh$ is the projection operator with range corresponding to $\textrm{span} \{\psi(\tilde{x}_1),\dots,\psi(\tilde{x}_{M_X})\}$. Note, in particular, that $P_{\tilde{X}}$ has rank $M_X$ when $k_X$ is universal. $P_{\tilde{Y}}: {\cal F} \to {\cal F}$ on $\tilde{Y}$ in defined analogously.

\paragraph{Combining the two approximations.}
Assume in this section, that the kernels of choice are universal (as, e.g., the Laplacian or the Gaussian kernel).
Let $P_{\tilde{X}}:\hh \to \hh$ be the projection operator associated to the Nystr\"om points on $X$ and $P_{\tilde{Y}}$ the one associated to the Nystr\"om point on $Y$. Combining the two approximations, we define the following estimator for $D_\la$,
\eqals{
\widehat{D}_\la(f,g) := \la \|(P_{\tilde{Y}} G_N P_{\tilde{X}} G_N^*P_{\tilde{Y}} + \la)^{-\frac{1}{2}} P_{\tilde{Y}} F_{\mu, N} P_{\tilde{X}} \|^2_{op}.
}
Note, however, that $\widehat{D}_\la(f,g)$ has a finite dimensional characterization that we are going to derive now.
Let $K_{\widetilde Y f} \in \mathbb R^{M_Y \times N}$ the matrix defined by $(K_{\widetilde Y f})_{i, j} = k_Y(\widetilde y_i, f(x_j))$, $K_{\widetilde Y g}$ in an analogous way, and finally, $K_{X \widetilde X}\in\R^{N \times M_X}$ the matrix defined by $(K_{X\widetilde X})_{i, j} = K_X(x_i, \widetilde x_j)$. Let $\widehat \mu = \left[ \mu(x_1), \ldots, \mu(x_{M_X})\right]$ and $R_{\widetilde X} \in\R^{M_X \times M_X}$ be the upper-triangular Cholesky decomposition of $K_{\widetilde X \widetilde X}$ defined by $(K_{\widetilde X \widetilde X})_{ij} = k_X(\widetilde x_i, \widetilde x_j)$. Analogously, define $K_{\widetilde Y \widetilde Y}$ and define $R_{\widetilde Y}\in\R^{M_Y \times M_Y}$ its Cholesky decomposition. Note that the decomposition exists since the kernel $k_X$ is universal and so the kernel matrix $K_{\tilde{X},\tilde{X}}$ is invertible.

We introduce the following operators $\widehat{A}, \widehat{B} \in \R^{M_Y \times M_X}$, which are the finite dimensional representations in appropriate spaces $\R^{M_X}, \R^{M_Y}$ of, respectively, $\tilde{P}_Y F_{\mu, N} \tilde{P}_X$ and $\tilde{P}_Y G_N \tilde{P}_X$:
\eqal{
\widehat{A} ~&=~ \frac{v_X}{N} R_{\widetilde Y}^{-T} K_{\widetilde Y f} \diag{\widehat \mu} K_{X\widetilde X} R_{\widetilde X}^{-1},\label{eq:widehatA}\\
\widehat{B} ~&=~ \frac{v_X}{N} R_{\widetilde Y}^{-T} K_{\widetilde Y g} K_{X\widetilde X} R_{\widetilde X}^{-1}. \label{eq:widehatB}
}
In particular, we have the following characterization for $\widehat{D}_\la$.
\begin{lemma}\label{lm:widehatD}
With the notation above,
\eqal{\label{eq:widehatD-finite-dimensional}
\widehat{D}_\la(f,g) = \la \|\widehat{A}^* (\widehat{B} \widehat{B}^* + \la I)^{-1} \widehat{A}\|_{op}.
}
\end{lemma}
Note that, in practice, $\widetilde X$ and $\widetilde Y$ can be chosen either deterministically, e.g. on a grid, or randomly. Now, we provide a bound on the approximation error associated to $\widehat{D}_\la(f,g)$. We assume that the $N$ points in $x_1,\dots, x_N$ are sampled independently and uniformly at random in $X$ and, moreover, that the $M_X$ points in $\widetilde{X}$ and the $M_Y$ points in $\widetilde{Y}$ are sampled independently and uniformly at random in, respectively, $X, Y$ (similar result can be derived for a grid). %
\begin{theorem}\label{thm:appr-error-widehatD}
Let $\delta \in (0,1)$. Let $X \subset \R^d, Y \subset \R^p$ be bounded sets and $k_X, k_Y$ be Sobolev kernels with smoothness, respectively, $s + d/2$ and $z+p/2$, for some $s,z > 0$. There exists two constants $c_1, c_2$ s. t., when $M_X \geq c_1$ and $M_Y \geq c_2$, then the following holds with probability $1-\delta$,
$$|\widehat{D}_\la(f,g) - D_\la(f,g)| \leq c\big(\tfrac{\log\frac{1}{\delta}}{\la\sqrt{N}} + \tfrac{(\log \frac{M_X}{\delta})^{\alpha}}{\la M_X^{s/d}} + \tfrac{(\log\frac{M_Y}{\delta})^{\beta}}{\la M_Y^{z/p}}\big),$$
for any measurable $f,g : X \to Y$, where $\widehat{D}_\la(f,g)$ is defined as in \cref{eq:widehatD-finite-dimensional}. Here $c_1, c_2, c$ depend only on $X, Y, \mu, s, z, d, p$, while $\alpha = s/d+1/2$, $\beta = z/p+1/2$.
\end{theorem}

The theorem above shows that the estimation error of $\widehat{D}_\la$ with respect to $D_\la$ goes to $0$ when $N, M_X, M_Y \to \infty$. On the contrary, the error diverges in $\la$. This is in accordance with the fact that the error is of variance type and shows that $\la$ plays the role of a regularization parameter. The bound shows also that, when $s \gg d$ and $z \gg p$, i.e. when we are choosing very smooth Sobolev kernels, the decay rate of the error in $M_X$ and $M_Y$ is faster. For example, if we choose $s = r d$, $z = r p$, for some $r > 0$, then choosing $M_X = M_Y =  O(N^{r/2})$ leads to the rate
$$ |\widehat{D}_\la(f,g) - D_\la(f,g)| = O\left(\frac{1}{\la \sqrt{N}}\right).$$

\paragraph{On the choice of $\la$.}
To conclude, a choice of $\la$ as $\la = N^{-1/4}$ guarantees a final convergence rate of $\widehat{D}_\la$ to $D_\la$ in the order of $N^{-1/4}$ and, together with \cref{theorem:main-theorem} a level of invariance to diffeomorphism for $\widehat{D}_\la$ of the order
$$\widehat{D}_\la(f, f \circ Q) = O(N^{-1/4} C_\mu^2 C_Q^2),$$
which can become a statistically significant threshold to decide if, in practice $f \approx g$ up to diffeomorphism.
Clearly the choice of $r,s$ while reducing the number of Nystr\"om points required in the approximation (with important computational implications that we see below) increases the constant $C_Q$ as shown, e.g., in \cref{ex:diffeo}, where $m = s + d/2$.

\paragraph{Algorithm and computational complexity.}
The final form of the empirical estimator $\widehat{D}_\la$ is \cref{eq:widehatD-finite-dimensional}. An efficient algorithm to compute it consists in (1) first computing the matrices $\widehat{A}$ and $\widehat{B}$, then the inverse $\widehat{C} = (\widehat{B}\widehat{B}^* + \la)^{-1}$ and finally compute the largest eigenvalue of $\widehat{A}^* \widehat{C} \widehat{A}$ via a power iteration method \cite{trefethen1997numerical}. %
Assuming that (a) the cost of one kernel computation in $\R^d$ is $O(d)$ (as in the case of any translation invariant kernel as the Laplace kernel) (b) $M_X \leq N$ and $M_Y \leq N$ (which is reasonable in light of \cref{thm:appr-error-widehatD}), then the cost of computing $\widehat{D}_\la$ with the algorithm above is
$O(d N M_X + p N M_Y + M_XM_Y^2 + M_X^3 + M_Y^3)$.
Choosing the parameters, as in the discussion after \cref{thm:appr-error-widehatD}, with $r = 1$, would lead to a total computational cost of 
$O(N^{1.5}(p+d)).$

\begin{figure}[t]
    \centering
    \includegraphics[width=0.75\textwidth]{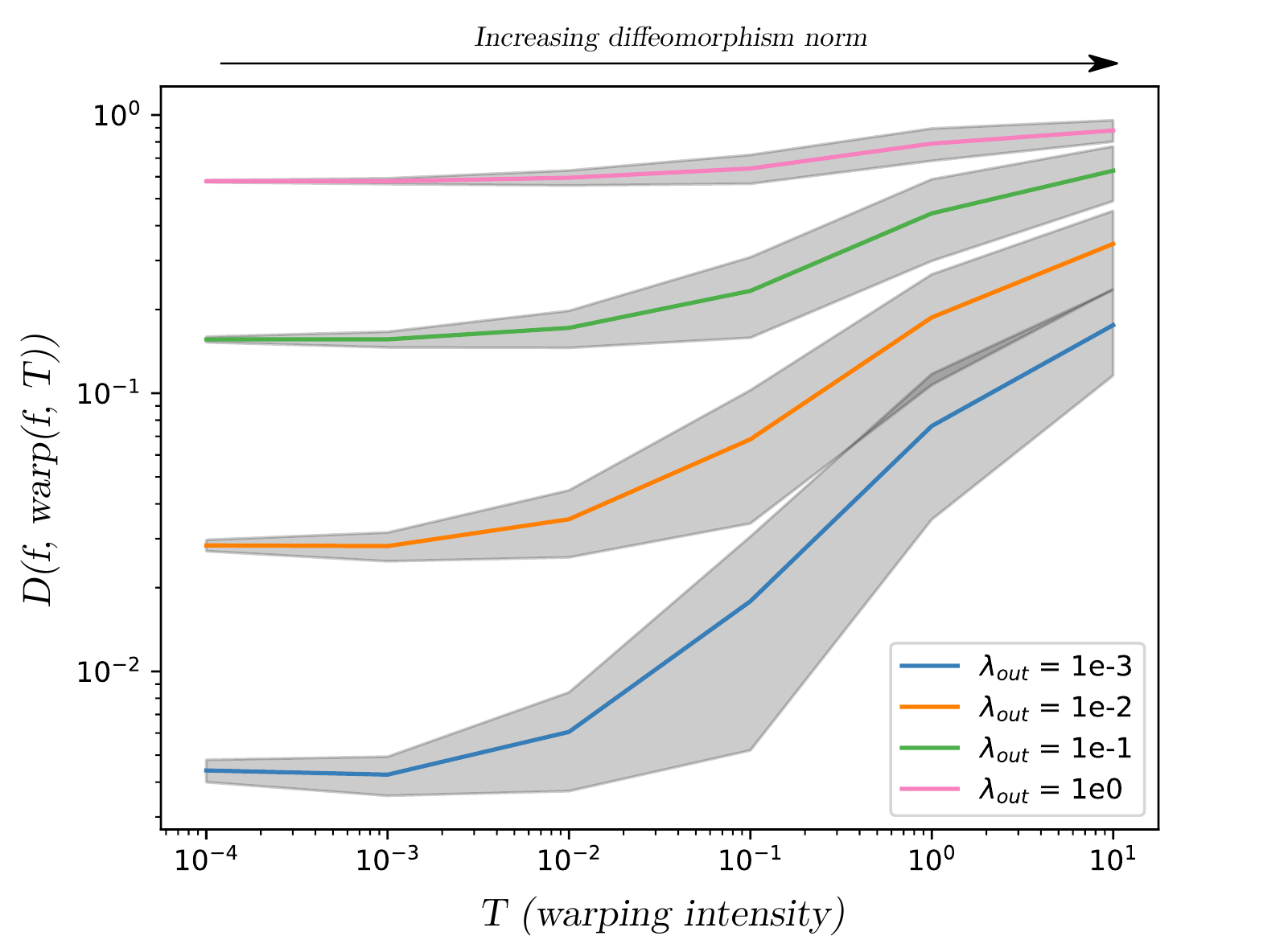}
    \caption{\emph{Effect of regularization.} We consider the same image and setting as in \cref{fig:warping} and vary $T$ and $\lambda$ (shaded areas are std deviation over $50$ warps). Observe: (1) $\widehat D_\la$ increases as a function of $T$ (as the norm of the diffeomorphism increases); (2) $\widehat D_\la$ is  proportional to $\lambda$. Both of these phenomena are predicted by \cref{theorem:main-theorem}. See \cref{sec:regularization} for more details.} 
    \label{fig:regularization}
\end{figure}

\section{Experiments}\label{sec:experiments}

This section investigates the empirical performance of \Diffy. We observe that, in line with our results in \cref{theorem:main-theorem}, when $g=f\circ Q$ the resulting $D_\la$ is small, while it is consistently large for signals that are not diffeomorphic versions of $f$.

\subsection{Implementation details}\label{sec:details}

We implemented $\widehat{D}_\la$ as in \cref{eq:widehatD-finite-dimensional} as described in the end of \cref{sec:approximation} using standard linear algebra routines such as matrix and matrix-vector products, matrix inversions, and eigendecompositions. The Python source code used for the experiments presented here is freely available at \href{https://github.com/theophilec/diffy}{https://github.com/theophilec/diffy}, depends on Numpy and Pytorch and supports using GPUs. 

\textbf{Choice of mask.}
We choose $\mu$ to be a Blackman window, a standard windowing function in image processing. In 1-D, the Blackman window is defined for any $0 \leq t \leq 1$ as:
$\mu(t) = 0.42 - 0.5\cos(2\pi t) + 0.08 \cos(4\pi t)$ \cite{oppenheim99}.
We generalize it to higher-dimension by considering its tensor-product over dimensions.

\textbf{Choice of kernel.}
Because we work with images in the experiments we present, $X = \R^2$ (coordinate space) and $Y = \R^3$ (color space). We consider the Gaussian kernel defined as $k(x, x^\prime) = \exp(- \|x - x'\|^2/(2\sigma^2))$ on $X$ and the Laplace kernel defined as $k(y, y^\prime) = \exp(-a \| y - y^\prime \|)$ on $Y$.

\textbf{Datasets.}
We rely on images from Imagenet (more precisely from the \texttt{Imagenette} subset), example images from the Matlab software (\texttt{peppers}), and finally images taken with our personal devices for illustrations (\texttt{raccon, flowers, objects}). All images are made available with the source code.

\textbf{Diffeomorphism generation.} Diffeomorphism are obtained either by affine transformations or by generating warpings. In particular, for warpings, we use the code from \cite{wyartdiffeo}. We generate random transformations of images by displacing each of its coordinates independently (while enforcing zero displacement at the edge of the grid) then interpolating the colors.  We choose the standard deviation of the displacements of each pixel so as to obtain a transformation of given (average) displacement norm. This can be controlled by two parameters $T$ (a temperature, between $10^{-4}$ and $10$) and $c$ (a cut-off parameter, taking $c=2$). We denote $\text{warp}(f, T)$ such a (random) warp. Examples of warps for various $T$ parameters (and samples) are provided in \cref{sec:appendix-warping}.

\subsection{Illustrative examples}\label{sec:illustration}

In \cref{fig:example}, we show how the $h$ and $q$ that optimize \Diffy concentrate on related regions on for different scenes which which are related by a diffeomorphism, in particular rigid-body and perspective for \texttt{objects} and keystone deformation for \texttt{raccoon}. We present supplementary illustrative examples in \cref{app:additional-experiments}.

\subsection{Invariance to warping}\label{sec:invariance-warping}
Diffeomorphisms (even infinitely regular) are a much wider class of transformations than rigid body transformations such as scale, rotation and translation (or combinations thereof). In this experiment (\cref{fig:warping}), we evaluating \Diffy's behavior against a wide family of transformations (we call warps). Consider $f$ an image from $\texttt{Imagenette}$. For $T = 10^{k}$ for $-4 \leq k \leq 2$ and various images, we evaluate $\widehat D_\lambda(f, \text{warp}(f, T))$ as well as $\textrm{RMSE}(f, \text{warp}(f, T))$. We compare the values observed to $\widehat D_\la(f, g)$ and $\textrm{RMSE}(f, g)$ for random images $f$ and $g$. We call this the \emph{random regime} ($90\%$ confidence interval). Finally, we look at the performance of \Diffy on a fixed image (gas station), with repeating warps. \cref{fig:warping} shows that \Diffy is invariant to diffeomorphic warping for $T \leq 10^{-1}$ whereas the Euclidean distance increases exponentially with $T$ (making $\text{warp}(f, T)$ indistinguishable from $g$, a random image).

\subsection{Invariance to rotation}\label{sec:invariance-rotation}
The \texttt{peppers} image is often used to demonstrate image registration techniques. In this experiment (see \cref{fig:rotation}) we show that \Diffy is invariant to rotation using patches taken from it. Consider $f$ a patch from \texttt{peppers} and $g = \text{rotate}(f, \alpha)$, rotated version of $f$ by angle $\alpha$ (in practice, we rotate a larger patch then crop to avoid artifacts). We then compare $D(f, \text{rotate}(f, \alpha))$ and $\textrm{RMSE}(f, \text{rotate}(f, \alpha))$. We show that while the Euclidean distance is \emph{constant} for $\alpha \neq 0$, \Diffy is \emph{invariant}. We compare \Diffy's and the Euclidean distance values with their values for random patches from the image. As before, we call this the \emph{random regime} ($90\%$ confidence interval). This shows that the Euclidean distance is not able to distinguish between a random patch and a rotated version of the same patch, while \Diffy can. See \cref{fig:rotation} for the results of the experiments.

\subsection{Effect of regularization}\label{sec:regularization}
In order to understand the effect of regularization, we reuse the setup from \cref{sec:invariance-warping} with a single image, with varying $\lambda$. In \cref{fig:regularization}, we observe two phenomena: (1) as $T$ increases, so does $D$; (2) as $\lambda$ decreases, so does $D$. This shows that \Diffy behaves close to what is predicted by \cref{theorem:main-theorem}. Indeed, $D$ seems proportional to $\lambda$. Also, as $T$ increases, so does the norm of the transformation between $f$ and $\text{warp}(f, T)$. This makes the upper bound of \cref{theorem:main-theorem} increase in turn.

\subsection*{Acknowledgments}

This work was funded in part by the French government under management of Agence Nationale de la Recherche as part of the “Investissements d’avenir” program, reference ANR-19-P3IA-0001 (PRAIRIE 3IA Institute). We also acknowledge support from the European Research Council (grant REAL 947908). 
Benjamin Guedj acknowledges partial support by the U.S. Army Research Laboratory and the U.S. Army Research Office, and by the U.K. Ministry of Defence and the U.K. Engineering and Physical Sciences Research Council (EPSRC) under grant number EP/R013616/1. Benjamin Guedj acknowledges partial support from the French National Agency for Research, grants ANR-18-CE40-0016-01 and ANR-18-CE23-0015-02.
Carlo Ciliberto acknowledges the support of the Royal Society (grant SPREM RGS/R1/201149) and Amazon.com Inc. (Amazon Research Award – ARA 2020)

\bibliographystyle{plainnat}
\bibliography{biblio}

\newpage

\appendix
\addcontentsline{toc}{section}{Appendix}
\section*{Appendix}

\section{Background} \label{app:background}

We recall here the classical version of change of variable theorem for $\R^d$, see e.g. Thm. 40.7 of \citet{aliprantis1998principles}.
\begin{theorem}[Change of Variables]\label{thm:change-of-variable}
Let $V$ be an open set in $\R^d$ and $Q:\R^d \to \R^d$ be an injective continuosly differentiable map. Let $f \in L^1(V)$. Denote by $\nabla Q(x) \in \R^{d\times d}$ the gradient of $Q$ for any $x \in \R^d$. Then, we have
$$\int_{Q^{-1}(V)} f(Q(x)) |\nabla Q(x)| dx = \int_{V} f(y) dy. $$
\end{theorem}

\subsection{Sobolev spaces}

We recall some basic properties of Sobolev spaces. The Sobolev space $H^m(Z)$ for $m > 0$ and an open set $Z \subseteq \R^d$ is defined as follows \cite{adams2003sobolev}

\begin{align*}
H^m(Z) = \big\{f \in L^2(\R^d) ~\big|~ \|f\|_{H^m(Z)}\big\} < \infty\big\},\\\|f\|^2_{H^m(Z)} = \sum_{\alpha_1+\dots+\alpha_d \leq m} \int_{Z} \left|\frac{\partial^{\alpha_1 + \dots + \alpha_d} f(x)}{\partial x_1^{\alpha_1} \dots \partial x_d^{\alpha_d}}\right|^2 dx.
\end{align*}
Analogously, we define the space $H^m(Z, \R^t)$ for functions with output in $\R^t$, $t \geq 1$, with norm $\|f\|_{H^m(Z,\R^t)}^2 = \sum_{j=1}^t \|f_j\|_{H^m(Z)}^2$.
In the following theorem we collect some important properties of $H^m(Z)$ that will be useful for the proof of \cref{theorem:main-theorem}.
\begin{theorem}\label{thm:good-rkhs}
Let $m > d/2$ and $Z$ be an open set with locally Lipschitz continuous boundary. The following properties hold
\begin{enumthm}
    \item \label{thm:sobolev-containCinfty} $H^m(Z)$ is a Reproducing Kernel Hilbert space. Moreover, $H^m(Z) \subset C(Z)$, and, when $Z$ is bounded we have $f|_Z \in H^m(Z)$, for any $f \in C^m(\R^d)$. 
    \item \label{thm:sobolev-restriction-extension} {\em Restriction and extension.}  
    For any $u \in H^m(\R^d)$, it holds that $u|_Z \in H^m(Z)$ and $\|u|_Z\|_{H^m(Z)} \leq c_1 \|u\|_{H^m(\R^d)}$. Moreover, for any $u \in H^m(Z)$ there exists a function $E_Z[u] \in \hh(\R^d)$ such that  $(E_Z[u])|_Z = u$ and $\|E_Z[u]\|_{H^m(\R^d)} \leq c_2\|u\|_{H^m(Z)}$. The constants $c_1,c_2$ depend only on $Z, m, d$. 
    \item \label{thm:sobolev-pointwise-product} {\em Pointwise product.} 
    For any $u, v \in H^m(Z)$ we have $u \cdot v \in H^m$. In particular, there exists $c_0 > 0$ depending only on $Z, m, d$ such that $\|u \cdot v\|_{H^m(Z)} \leq c_0 \|u\|_{H^m(Z)}\|v\|_{H^m(Z)}$. 
\end{enumthm}
\end{theorem}
\begin{proof}
The space $H^m(\R^d)$ is a RKHS due to the characterization of the norm with respect to the Fourier transform \cite{wendland2004scattered}. The space $H^m(Z)$ is a RKHS since any restriction of a RKHS to a subset of the set of definition is still a RKHS \cite{aronszajn1950theory}. The inclusions derive directly by the definition of the $H^m$ norm \citep[see][for more embeddings of Sobolev spaces]{adams2003sobolev}.

The second point is a classical result on Sobolev space and is derived in \citet{adams2003sobolev}. The third point is equivalent to showing that the Sobolev spaces with $m > d/2$ are Banach algebras and it is derived in \citet{adams2003sobolev}. For the case $Z=\R^d$ an explicit derivation based on the Fourier transform is done in Lemma 10 of \citet{rudi2020finding}. 
\end{proof}

\section{Proof of the more general version of Theorem~\ref{theorem:main-theorem}}

In the next subsection we introduce some preliminary results that will be necessary to prove the more general version of Theorem~\ref{theorem:main-theorem}, which is in the subsection \cref{sec:general-theorem}. 

\subsection{Preliminary result on composition of Sobolev functions}\label{sec:functions}
The following theorem quantifies the fact that a Sobolev space with $m > d/2$ is closed with respect to composition with diffeomorphisms. There exists many abstract results about the closure of composition in Sobolev space in the literature  \citep[see \emph{e.g.}][]{bruveris2017regularity}. Here, however, we want a quantitative bound. We base our result on the explicit bound in \citet{bourdaud2011composition}. To obtain a final readable form we have to do a bit of slalom between restriction and extension between $Z$ and $\R^d$. Indeed, we need functions that are equivalent to the functions of interest on $Z$, but whose norm does not diverge when going to $\R^d$. For example, constant functions don't belong to $H^m(\R^d)$, but for us it is enough to have a function that is equal to a constant on $Z$ and that goes to zero at infinity fast enough to have finite $H^m(Z)$ norm.

\begin{theorem}[Smooth composition]\label{thm:sobolev-smooth-composition}
 Let $m > d/2$. Let $Q:\R^d \to \R^d$ be an invertible $m$-times differentiable map whose inverse has continuous Lipschitz derivative. Let $Z$ be open bounded set with Lipschitz boundary and a compact $\Omega$ such that $\Omega \subset Z$ and $Q^{-1}(\Omega)\subset Z$. Assume also, without loss of generality, that $0$ is in the interior of $\Omega$. For any $h \in H^m(\R^d)$ supported on $\Omega$ the following holds:
 \begin{enumerate}
     \item there exists $b \in H^m(\R^d)$ satisfying $b(x) = h(Q(x))$ for any $x \in Q^{-1}(\Omega)$. 
     \item if $h(x) = 0$ for any $x \in \R^d\setminus \Omega$, then $b(x) = 0$ for any $x \in \R^d \setminus Q^{-1}(\Omega)$,
     \item the norm of $b$ is controlled by
\eqal{
\|b\|_{H^m(\R^d)} & ~\leq~ C\, \|h\|_{H^m(Z)}\, d_{\Omega, Q}^{-m-d/2} \, (1 + \|Q\|_{H^m(Z, \R^d)} + D^m + L^m),
}
\end{enumerate}
where $d_{\Omega,Q} := \min[1,~d_H(Q^{-1}(\Omega), Q^{-1}(Z) \cap Z)]$, $d_H(A,B)$ is the Haussdorff distance between two sets $A, B$ and $D := \textrm{diam}(Z)$, $L = \max_{x \in Z} \|\nabla Q (x)\|$, while $C$ depends only on $d, m, Z$.
\end{theorem}
\begin{proof}
The fact

To apply \citet{bourdaud2011composition} we need a function such that $h(0) = 0$. With this aim, we rewrite $h \circ Q$ as 
$$h \circ Q = h(0) + ((h-h(0)) \circ Q).$$

\paragraph{Step 1. Construction of $b$.}

Denote by $U$ the open set $U = Q^{-1}(Z) \cap Z$. Note that the set is not empty since, by construction, $Q^{-1}(\Omega) \subset U$.
Define $\tilde{Q} = E_{Z}[Q|_{Z}]$, i.e. the extension to the whole $\R^d$ of the restriction of $Q$ on the set $Z$ (restriction and extension done componentwise). By the first point we have that $Q|_{Z}$ belongs to $H^m(Z,\R^d)$ and, by the second point, that $\tilde{Q}$ belongs to $H^m(\R^d, \R^d)$ and moreover $\tilde{Q}(x) = Q(x)$ for any $x \in Z$.
Denote by $s = h(0) \in C^\infty(\R^d)$ the constant function equal to $h(0)$ everywhere on $\R^d$. In particular, note that $s|_Z \in H^m(Z)$ via \cref{thm:sobolev-restriction-extension}. Denote by $\tau$, the extension of $s|_Z$ to $\R^d$, i.e., $\tau = E_Z[s|_Z]$. Denote by $u$, the function $u = h|_Z - s|_Z$ and by $\tilde{u} \in H^m(\R^d)$ the function $\tilde{u} = E_{Z}[h|_Z - s|_Z]$. Note that $\tilde{u}(x) = h(x) - h(0)$ for any 
$x \in Z$, and, in particular for any $x \in \Omega$. Define by $\rho$ the $C^\infty(\R^d)$ function that is $1$ on $Q^{-1}(\Omega)$ and $0$ on $\R^d \subseteq U$.

Now define,
$$b ~~=~~ \rho ~\cdot~ (\tau ~+~ \tilde{u} \circ \tilde{Q}).$$
Denote by $\tilde{b}$ the function $\tilde{b} = \tau + \tilde{u} \circ \tilde{Q}$. We have that $\tilde{b}(x) = h(Q(x))$ for any $x \in U$. Since $Q(U) \subseteq Z$ and that $\tilde{h}(x) = h(x), \tilde{Q}(x) = Q(x)$ for any $x \in Z$. In particular, since $\tilde{b}(x) = 0$ for all $x \in U \setminus Q^{-1}(\Omega)$ and by definition of $\rho$, we have: (a)  $\rho \cdot \tilde{b} = \tilde{b} = h(Q(x))$ on $Q^{-1}(\Omega)$, (b) $\rho \cdot \tilde{b} = 0$ on $U Q^{-1}(\Omega)$; (c) $\rho \cdot \tilde{b} = 0$ and $0$ on $\R^d \setminus U$. Then $b = \rho \cdot \tilde{b} = h(Q(x))$ for any $x \in \R^d$.

\paragraph{Step 2. Bound of $b$.}
Now, let us bound the norm of $b$. By applying \cref{thm:sobolev-pointwise-product} we have 
\eqal{\label{eq:bound-composition}
\|b\|_{H^m(\R^d)} & \leq c \|\rho\|_{H^m(\R^d)}(\|\tau\|_{H^m(\R^d)} ~+~ \|\tilde{u} \circ \tilde{Q}\|_{H^m(\R^d)}).
}
The bound for $\tau$ is obtained applying \cref{thm:sobolev-restriction-extension} and the definition of $H^m(Z)$ norm, as follows,
$$\|\tau\|_{H^m(\R^d)} \leq c_2(Z) \|s|_Z\|_{H^m(Z)} = c_2 h(0) \textrm{vol}(Z)^{1/2}.$$
Now we bound the norm $\|\tilde{u} \circ \tilde{Q}\|_{H^m(\R^d)}$ with respect to the norms of $\tilde{u}$ and $\tilde{Q}$. We use a result on the composition of Sobolev functions \citep[][Theorem 27]{bourdaud2011composition} that is highly technical, but allows to highlight the quantities of interests for us. For a more extensive treatment of the topic see for example \citet{runst2011sobolev}. Since $\tilde{u}(0) = 0$ by construction, by applying Theorem 27 of \citet{bourdaud2011composition} with $p=2$ and considering that their norm $\|\cdot\|_{\dot{W}_{E^p}^m}$ is bounded by $\|\cdot\|_{H^m(\R^d)}$ and analogously $\|\cdot\|_{\dot{W}^1_{mp}} \leq \|\cdot\|_{\dot{W}^1_{\infty}} \leq \|\cdot\|_{W^1_\infty(\R^d)}$,
\eqal{
\|\tilde{u} \circ \tilde{Q}\|_{H^m(\R^d)} \leq c (\|\tilde{u}\|_{H^m(\R^d)} + \|\tilde{u}\|_{\dot{W}^1_\infty(\R^d)})(\|\tilde{Q}\|_{H^m(\R^d, \R^d)} + \|\tilde{Q}\|^m_{\dot{W}^1_\infty(\R^d, \R^d)})
}
Here $\|z\|_{\dot{W}^1_\infty(A, \R^d)} = \sup_{x \in A, j \in \{1,\dots,t\}} \|\nabla z_j\|$ for any differentiable $z:A \to \R$ and open set $A$.
Now to conclude, note that
$$\|\tilde{u}\|_{H^m(\R^d)} = \|E_{Z}[h|_Z - s|_Z]\|_{H^m(\R^d)} \leq c_2 \|h|_Z - s|_Z\|_{H^m(Z)} \leq c_2 \|h\|_{H^m(Z)} + c_2 h(0) \textrm{vol}(Z)^{1/2},$$
moreover
$$\|\tilde{Q}\|_{H^m(\R^d,\R^d)} = \|E_Z(Q|_Z)\|_{H^m(\R^d, \R^d)} \leq c_2 \|Q|_Z\|_{H^m(Z, \R^d)}.$$
Note also that for the Sobolev space $W^1_\infty$ there exists the same type of result as \cref{thm:sobolev-restriction-extension} and for the same extension operator defined in \cref{thm:sobolev-restriction-extension} (which is a {\em total extension operator}, see e.g. the Stein extension theorem, Thm 5.4 page 154 of \citealp{adams2003sobolev}), so, as above, we have 
$$\|\tilde{u}\|_{W^1_\infty(\R^d)} \leq c_2 \|h\|_{W^1_\infty(Z)} + c_2 h(0),$$
and also 
\eqals{
	\|\tilde{Q}\|_{W^1_\infty(\R^d,\R^d)}  &\leq c_2 \|Q|_Z\|_{W^1_\infty(Z, \R^d)}  \\ &= c_2\sup_{x \in Z} \max(\|Q(x)\|, \|\nabla Q(x)\|) \\&\leq  c_2\textrm{diam}(Z) + c_2 \sup_{x \in Z} \|\nabla Q (x)\|.
}
Substituting the six bounds above in \cref{eq:bound-composition}, we obtain
\eqals{
	\|b\|_{H^m(\R^d)} \leq c &\|\rho\|_{H^m(\R^d)}(c_2 h(0) \textrm{vol}(Z)^{1/2} \\&~+~ c' ( \|h\|_{H^m(W)} + \|h\|_{W^1_\infty(Z)} + h(0) c'')(\|Q|_Z\|_{H^m(Z, \R^d)} + D^m + L^m)),
}
where $D =\textrm{diam}(Z)^m $, $L = \sup_{x \in Z} \|\nabla Q (x)\|$ and $c' = c c_2^2 (1 + 2^m c_2^{m-1})$, $c''=1+\textrm{vol}(Z)^{1/2}$, with $c, c_2$ depending only on $d, m, Z$.
The final result is obtained considering that $x A + d (B + y A) R \leq (1+ d + d y)(B+A)(1+R)$ for any $x,y,d,A,B, R \geq 0$, and applying this result with $x=c_2(Z)\textrm{vol}(Z)^{1/2}$, $A = h(0)$, $d = c'$, $B = \|h\|_{H^m(Z)} + \|h\|_{W^1_\infty(Z)}$, $y= c''$, $R = \|Q|_Z\|_{H^m(Z, \R^d)} + D^m + L^m$. In particular, in the final result, the constant $C$ corresponds to $C = 3(1+ d + d y)$ and we used the fact that $h(0) \leq \|h\|_{W^1_\infty(Z)} \leq \|h\|_{H^m(Z)}$, then $\|h\|_{H^m(Z)} + \|h\|_{W^1_\infty(Z)} + h(0) \leq 3 \|h\|_{H^m(Z)}$.

\paragraph{Step 3. The norm of $\rho$}.  Let $A_{t}$ be the set $A_{t} = \{x ~|~ \min_{y \in Q^{-1}(\Omega)} \|x-y\| \leq t\}$. Let $\eta = d_H(Q^{-1}(\Omega), \overline{U})$, i.e., the Haussdorff distance between the sets $Q^{-1}(\Omega)$ and $\overline{U}$, corresponding to the largest $\eta$ for which $A_\eta \subseteq U$. Note that $\eta > 0$ since $Q^{-1}(\Omega)$ is compact, while $U$ is open and $Q^{-1}(\Omega) \subset U$. The fact that $\eta > 0$ implies that for any $\eta' < \eta$ it holds that $A_{\eta'} \subset U$. 

The function $\rho$ can be obtained by the convolution of the indicator function $1_{A_{\eta/2}}$ with the bump function $\psi_{\eta/2}(x) =  (\eta/2)^{-d} \psi(x/(\eta/2))/S$ where $S = \int_{\R^d} \psi(x) dx$ and $\psi$ is an infinitely smooth non-zero non-negative function that is $0$ on $\|x\| \geq 1$ as for example $\psi(x) = \exp(-1/(1-\|x\|^2)_+)$ for any $x \in \R^d$, where $(z)_+ = \max(0,z)$. In particular, since (a) $\|f(y-\cdot)\|_{H^m(\R^d)} = \|f\|_{H^m(\R^d)}$ for any $y \in \R^d$, by construction of the norm $H^m(\R^d)$, and (b) $\|t^{-d} f(\cdot/t)\|_{H^m(\R^d)} \leq c_6 t^{-d/2} \max(1,t^{-m}) \|f\|_{H^m(\R^d)}$ (see, e.g., Proposition 3 of \citealp{runst2011sobolev}) we have
\eqals{
\|\rho\|_{H^m(\R^d)} &\leq \int_{A_{\eta/2}} \|\psi_\eta(y-\cdot)\|_{H^m(\R^d)} dy  \leq \textrm{vol}(A_{\eta/2}) \|\psi_\eta\|_{H^m(\R^d)} \\
& \leq c_6 \|\psi\|_{H^m(\R^d)} ~ \textrm{vol}(A_{\eta/2}) ~ (\eta/2)^{-d/2} \max(1,(\eta/2)^{-m}).
}
To conclude note that $\textrm{vol}(A_\eta) \leq \textrm{vol}(Z)$, since $A_{\eta/2} \subset U \subseteq Z$.
\end{proof}

\begin{lemma}[Existence and norm of $\tilde{q} \in H^m(\R^d)$]\label{lm:existence-q}
Let $\mu$ be an infinitely differentiable function, with compact support $\Omega \subset \R^d$. Let $Q:\R^d \to \R^d$ be a $C^{m+1}$ diffeomorphism. Let $Z$ be an open bounded set with Lipschitz boundary and such that $\Omega \subset Z$ and $Q^{-1}(\Omega) \subset Z$. Moreover, assume without loss of generality that $0$ is in the interior of $\Omega$.
For any $h \in H^m(\R^d)$, there exists a function $\tilde{q} \in H^m(\R^d)$ satisfying
\eqal{\label{eq:q-equivalence}
 \tilde{q}(y) = \begin{cases} h(Q(y)) \mu(Q(y)) |\nabla Q(y)| &  y \in Q^{-1}(\Omega) \\
 0 & y \in \R^d \setminus Q^{-1}(\Omega)
 \end{cases}.
}
In particular, let $b \in H^m(\R^d)$ be defined according to \cref{thm:sobolev-smooth-composition}, then
$$ \|q\|_{H^m(\R^d)} \leq C' \|h\|_{H^m(\R^d)}\|\mu\|_{H^m(\R^d)} d_{\Omega,Q}^{-m-d/2} \|\nabla Q\|_{H^m(Z, \R^d)}^d (1 + \|Q\|_{H^m(Z, \R^d)} + D^m + L^m),
$$
the constant $C$ depends only on $d, m, Z$, while $d_{\Omega, Q} := \min[1, d_H(Q^{-1}(\Omega), Q^{-1}(Z) \cap Z)$ and $d_H(A, B)$ is the Haussdorff distance between two sets $A, B$.
\end{lemma}
\begin{proof}
Let $h \in H^m(\R^d)$ and $\tilde{\mu}$ to be the extension of $\mu$ on $\R^d$ (which corresponds to $\tilde{\mu}(x) = \mu(x)$ for any $x \in \Omega$ and $\tilde{\mu}(x) = 0$ for $x \in \R^d \setminus \Omega$). Define the function $s(x) = h \cdot \tilde{\mu}$ and note that $s \in H^m(\R^d)$ since it is the product of two functions in $H^m(\R^d)$ (see \cref{thm:sobolev-pointwise-product}). 

Second, note that the function $r = (|\nabla Q|)|_Z \in H^m(Z)$ since (a) the map $\nabla Q$ belongs to $C^{m+1}(\R^d, \R^d)$ so its entries $(\nabla Q)_{i,j}|_Z$ belong to $H^m(Z)$ (see \cref{thm:sobolev-containCinfty}); and (b) the determinant of a matrix, by the Leibniz formula, is defined in terms of sums and products of its entries and $H^m(Z)$ is closed with respect to multiplication, by \cref{thm:change-of-variable}. To quantify its norm let's write explicitly the Leibniz formula \cite{trefethen1997numerical},
$$ |\nabla Q| = \sum_{\sigma \in S_d} sgn(\sigma) \prod_{i=1}^d e_{\sigma_i}^\top\frac{\partial Q(x)}{\partial x_i},$$
where $sgn(\sigma) \in \{-1,1\}$, $S_n$ is the set of permutations of of $d$ elements and $e_1,\dots,e_d$ is the canonical basis of $\R^d$. Note, in particular, that, by the equation above, since $|S_d| = d!$ and that $\|e_{\sigma_i}^\top\frac{\partial Q(x)}{\partial x_i}\|_{H^m(Z)} \leq \|\nabla Q\|_{H^m(Z,\R^d)}$, we have that
$$\|r\|_{H^m(Z)} \leq d! \|\nabla Q\|_{H^m(Z, \R^d)}^d.$$

Now consider the function $b \in H^m(\R^d)$ defined according to \cref{thm:sobolev-smooth-composition} we have that $b(x) = s(Q(x))$ for any $x \in \R^d$. Now, define $\tilde{q}$ as follows
$$ \tilde{q} = b  \cdot  E_Z[r].$$
The function $\tilde{q}$ is in $H^m(Z)$, since it is the product of two functions in $H^m(Z)$ (see \cref{thm:sobolev-pointwise-product} and $r \in H^m(Z)$ (see \cref{thm:sobolev-restriction-extension}). Note, in particular, that \cref{eq:q-equivalence} holds, by construction. To conclude, note that, by applying \cref{thm:sobolev-pointwise-product} and \cref{thm:sobolev-restriction-extension}, we have 
\eqals{
	\|q\|_{H^m(\R^d)} &\leq c_0 \|E_Z[r]\|_{H^m(\R^d)} \|b\|_{H^m(\R^d)} \\&\leq c_0 c_2 \|r\|_{H^m(Z)} \|b\|_{H^m(\R^d)} \\& \leq d! c_0 c_2 \|\nabla Q\|_{H^m(Z,\R^d)}^d \|b\|_{H^m(\R^d)}.
}
To conclude, we bound $b$ according with \cref{thm:sobolev-smooth-composition}, obtaining
$$\|b\|_{H^m(\R^d)} \leq C\, \|s\|_{H^m(W)}\, d_{\Omega, Q}^{-m-d/2}(1 + \|Q\|_{H^m(Z, \R^d)} + D^m + L^m), $$
and $\|s\|_{H^m(W)} \leq \|s\|_{H^m(\R^d)} \leq c_0 \|h\|_{H^m(\R^d)}\|\mu\|_{H^m(\R^d)}$, by \cref{thm:sobolev-pointwise-product}.
\end{proof}

\subsection{Proof of the general version of Theorem~\ref{theorem:main-theorem}}\label{sec:general-theorem}

\cref{theorem:main-theorem} is a particular case of the following theorem

\begin{theorem}\label{thm:general-theorem}
Let $X \subset \R^d$ be an open bounded set with Lipschitz boundary. Let $\mu \in C^\infty(\R^d)$ with compact support $\Omega \subset X$. Let the RKHS $\hh$ be $\hh = H^m(\R^d)$, i.e. the Sobolev space of smoothness $m$, with $m > d/2$. Denote by ${\cal F}$ the RKHS induced by the kernel $k_Y$ on $Y$, that we assume uniformly bounded. Then for any $C^{m+1}$ diffeomorphism $Q$ on $\R^d$ satisfying $Q^{-1}(\Omega) \subset X$, we have that for all $f, g$ measurable functions,
\eqals{
 g(x) = (f \circ Q)(x), ~~ \forall x \in Q^{-1}(\Omega) \qquad \textrm{implies} \qquad D(f, g) ~~\leq~~ \lambda ~ C_{\mu} C_{Q}, 
}
where $C_\mu = \|\mu\|_{H^m(\R^d)}$, and $C_Q$ is defined in \cref{eq:def-CQ} below and depends only on $\Omega, X, Q, d, m$.
\end{theorem}
\begin{proof} 
Now we have all the elements to prove the main theorem of the paper. Let $\tilde{q}$ as defined in \cref{lm:existence-q}, with $Z = X$. Moreover, let
$v \in {\cal F}$, where ${\cal F}$ is the reproducing kernel Hilbert space associated to the kernel $k_Y$ which is bounded by assumption. Then for any continuous $f$, we have that $v \circ f$ is continuous and bounded.
Denote by $\Theta_{f,g}(h,v,q)$ the quantity
$$\Theta_{f,g}(h,v,q) := \int_X v(f(x)) h(x) \mu(x) dx - \int_X v(g(y)) q(y) dy.$$
\paragraph{Step 1. Simplifying $\Theta_{f,g}(h,v,\tilde{q})$.} 
Since $\mu$ is supported on $\Omega \subseteq X$, by assumptions, we have
$$\int_X v(f(x)) h(x) \mu(x) dx = \int_\Omega v(f(x)) h(x) \mu(x) dx.$$
Moreover, by expanding the characterization of $\tilde{q}$ in \cref{eq:q-equivalence}, we have that $\tilde{q}(x) = 0$ for any $x \in \R^d \setminus Q^{-1}(\Omega)$, since $Q^{-1}(\Omega) \subseteq X$, we have
\eqals{
\int_X v(g(y)) \tilde{q}(y) dy &= \int_{Q^{-1}(\Omega)} v(g(y)) h(Q(y)) \mu(Q(y)) |\nabla Q(y)| dy,\\
& = \int_{Q^{-1}(\Omega)} v(f(Q(y))) h(Q(y)) \mu(Q(y)) |\nabla Q(y)| dy,
}
where we used in the last step that $g(y) = f(Q(y))$ for any $y \in Q^{-1}(\Omega)$, which is now the domain of integration.
\paragraph{Step 2. Applying the Change of Variable theorem.}
Note that the function $\tilde{q}$ is continuous since $H^m(\R^d)$ is subset of continuous functions. By applying the change of variable theorem we have
$$\int_{Q^{-1}(\Omega)} v(f(Q(x))) h(Q(x)) \mu(Q(x)) |\nabla Q(x)| dx = \int_{\Omega} v(f(y)) h(y) \mu(y) dy.$$
Then, by using the characterizations in Step 1, we have
\eqals{
\Theta_{f, g}(h,v,\tilde{q}) & = \int_X v(f(x)) h(x) \mu(x) dx - \int_X v(g(y)) \tilde{q}(y) dy \\
& = \int_\Omega v(f(x)) h(x) \mu(x) dx - \int_{Q^{-1}(\Omega)} v(f(Q(y))) h(Q(y)) \mu(Q(y)) |\nabla Q(y)| dy\\
& = 0.
}

\paragraph{Step 3. Bound on $D_\la(f, g)$.} 
Now, denote by $\hh$ the set $H^m(\R^d)$. Since $\tilde{q} \in \hh$, and $\Delta_{f, g}(h,v,\tilde{q}) = 0$, we have that
\eqal{
D_\la(f, g) = \max_{\|h\|_{\hh} \leq 1} \min_{q \in \hh} \max_{\|v\|_{\cal F} \leq 1} |\Theta_{f, g}(h,v,q)|^2 + \la \|q\|^2_{\hh} & \leq \max_{\|h\|_{\hh} \leq 1} \max_{\|v\|_{\cal F} \leq 1} |\Theta_{f, g}(h,v,\tilde{q})|^2 + \la \|\tilde{q}\|^2_{\hh} \\
& = \max_{\|h\|_{\hh} \leq 1} \la \|\tilde{q}\|^2_{\hh}.
}

\paragraph{Step 4. Simplifying $\|\tilde{q}\|_{H^m(\R^d)}$.}
We bound $\|\tilde{q}\|_{H^m(\R^d)}$ as in \cref{lm:existence-q}, where $b$ is bounded according to \cref{thm:sobolev-smooth-composition},
$$\|\tilde{q}\|_{H^m(\R^d)} \leq C'\|h\|_{H^m(\R^d)} C_\mu C_Q,$$
where $C_\mu := \|\mu\|_{H^m(\R^d)}$ and 
\eqal{\label{eq:def-CQ}
C_Q  ~~:= ~~ d_{\Omega, Q}^{-m - d/2} ~ \|\nabla Q\|_{H^m(X, \R^d)}^d (1 + \|Q\|_{H^m(X, \R^d)} + D^m + L^m),
}
where $D := \textrm{diam}(X)$,  $L = \max_{x \in X} \|\nabla Q (x)\|$ and $d_{\Omega, Q} := \min[1, d_H(Q^{-1}(\Omega), Q^{-1}(X) \cap X)]$ and $d_H(A, B)$ is the Haussdorff distance between two sets $A, B$. 

The final result is obtained by considering that we are optimizing with the constraint $\|h\|_{H^m(\R^d)} \leq 1$.
\end{proof}

\section{Proof of closed form of $D_\la$ and computational results}\label{sec:additional-proofs}

In the next lemma, we prove some important properties useful for the characterization of $D_\la$ in terms of $F_\mu, G$.

\begin{lemma}\label{lm:explicit-Fmu-G}
Let $X \subset \R^d$ be an open bounded set. The linear operators $F_\mu, G$ defined above are compact and trace class. 
Moreover, for any $h, q \in \hh$ and $v \in {\cal F}$, we have
\eqals{
\scal{v}{F_\mu h}_{\cal F} &= \int_X v(f(x)) h(x) \mu(x) dx, \\
\scal{v}{G q}_{\cal F} &= \int_X v(f(x)) q(x) dx.
}
\end{lemma}
\begin{proof}
Since $k_Y$ is bounded, and $f, g$ are measurable, then $\Phi(f(\cdot)):X \to {\cal F}$ is bounded and measurable. Since $\mu \in C^\infty(\R^d)$ by assumption $X$ is bounded and $h$ is bounded and continuous since it belongs to $\hh$ and $k_X$ is bounded and continuous, so $J := \int_X \|\Phi(f(x))\|_{\cal F} \|\psi(x)\|_{\hh}|\mu(x)| \dd x < \infty$. This guarantees the existence of the following Bochner integral
$F_\mu:=\int_X \Phi(f(x)) \otimes \psi(x) \mu(x)\dd x ~~ \in {\cal F} \otimes {\hh}.$
In particular, denoting by $\|\cdot\|_*$ the trace norm, i.e. $\|A\|_* = \tr(\sqrt{A^*A})$, and recalling that
$\|u \otimes v\|_* = \tr(\sqrt{(u \otimes v)^*(u \otimes v)}) = \|u\|_{F}\|v\|_{\hh}$ for any $u \in {\cal U}, v \in {\cal V}$ and any two separable Hilbert spaces ${\cal U}, {\cal V}$, we have
$$\|F_\mu\|_* \leq \int_X \|\Phi(f(x)) \otimes \psi(x)\|_* |\mu(x)| dx = \int_X \|\Phi(f(x))\|_{\cal F} \|\psi(x)\|_\hh |\mu(x)| dx =: J < \infty.$$
Then $F_\mu$ is trace class.
The same reasoning hold for $G$, considering that $\int_X \|\Phi(f(x))\|_{\cal F} \|\psi(x)\|_{H} \dd x < \infty$, since $X$ is compact.
\end{proof}

\subsection{Proof of \cref{theorem:closed-form}}\label{sec:proof-closed-form}
\begin{proof}
We have seen in \cref{lm:explicit-Fmu-G}, that since $X$ is a bounded set, then $F_\mu$ and $G$ are trace class and, by the representer property  
\eqals{
\scal{v}{F_\mu h}_{\cal F} &= \int_X v(f(x)) h(x) \mu(x) dx,\\
\scal{v}{G q}_{\cal F} &= \int_X v(f(x)) q(x) dx
}
Using this result and considering the linearity of the inner product and the variational characterization of the Hilbert norm (i.e. $\|u\|_{\cal F} = \max_{\|v\|_{\cal F} \leq 1} |\scal{v}{u}|_{\cal F}$  for any $u \in {\cal F}$), we have
\eqals{
\Delta_{f,g}(h,q) &= \max_{\|v\|_{\cal F} \leq 1} |\scal{v}{F_\mu h}_{\cal F}  - \scal{v}{G q}_{\cal F}|^2 \\
& = \max_{\|v\|_{\cal F} \leq 1} |\scal{v}{F_\mu h - G q}_{\cal F}|^2 \\
& = \|F_\mu h - G q\|_{\cal F}^2,
}
for any $h, q \in \hh$. From which we characterize $D_\lambda$ as
\eqal{
D_\lambda(f, g) = \max_{\norh{h}\leq 1} \min_{q\in\hh} \norf{F_\mu h - G q}^2 + \lambda \norh{q}^2.
}
Now, we prove that the problem above has a characterization in terms of the operatorial norm of a given operator.
First, notice that $q\mapsto \norf{Fh - Gq}^2 + \lambda \norh{q}^2$ is $2\lambda$-strongly convex. It therefor has a unique global minimizer $q^*(h)$ which is also a critical point. This leads to $q^*(h) = ZFh$ where $Z = (GG^* + \lambda I)^{-1}G^*$. Here $GG^* + \lambda I$ is a positive linear operator, and therefor invertible. 

So far, we have shown that: \eqal{D(f, g) = \max_{\norh{h}\leq 1}\norf{Fh - GZFh}^2 + \lambda \norh{ZFh}^2.}

Rewriting both squared norms as scalar products in $\mathcal F$ and $\hh$ and using the adjoint operators, we have that:
$\norf{Fh - GZFh}^2 + \lambda \norh{ZFh}^2 = \langle h, Th\rangle_\hh$ with $T = F^*(I - Z^*G^*)(I-GZ)F + \lambda F^* Z^*ZF$. Thus, we have rewritten $D$ as the operator norm of $T$: \eqal{D(f, g) = \max_{\norh{h}\leq 1}\langle h, Th\rangle_\hh = \norop{T}.}

We can now simplify $T$. Recall that for any bounded operator $A$, $A(A^*A + \lambda I)^{-1} = (AA^* + \lambda I)^{-1}A$ and $(A + \lambda I)^{-1}A = I - \lambda(A + \lambda I)^{-1}.$

Thus, $GZ = Z^*G^*= G(G^*G + \la I) ^{-1} G^*= (GG^* + \la I)^{-1}GG^* = I - \la (GG^* + \la I)^{-1}.$ Similarly, 
\eqal{Z^*Z &= G(G^*G + \la I) ^{-1}(G^*G + \la I) ^{-1} G^* = (GG^* + \la I) ^{-1}GG^*(GG^* + \la I) ^{-1}\\
&= (I - \la (GG^* + \la I)^{-1}(GG^* + \la I) ^{-1}= (GG^* + \la I) ^{-1} -\la (GG^* + \la I) ^{-2}.
}

Replacing these expression in $T$, we obtain:
\eqal{
T &= \la^2F^*(GG^* + \la I)^{-1})(GG^* + \la I)^{-1})F + \lambda F^* \left[(GG^* + \la I) ^{-1} -\la (GG^* + \la I) ^{-2}\right]F\\
  &= \la F^*(GG^* + \la I)^{-1}F.
}
Finally, 
\eqal{D_\lambda(f, g) = \norop{T} = \la \norop{F^*(G^* + \la I)^{-1} F}= \la\norop{(GG^* + \la I)^{-1/2}F}^2.}
\end{proof}

\subsection{Proof of \cref{lm:widehatD}}

Before proceeding with the proof of \cref{lm:widehatD}, we introduce some operators, that will be useful also in the rest of the paper. We recall that for this set of results we are assuming that the kernel $k_X$ is universal. This implies that the kernel matrix $K_{\tilde{X},\tilde{X}}$ is invertible and so $R_{\tilde{X}}$ exists and is invertible. The same holds for $R_{\tilde Y}$.

\begin{definition}[The operators $S, V:\hh\to\R^{M_X}$ and $Z, U: {\cal F} \to \R^{M_Y}$]\label{def:operators}
First define $S:\hh \to \R^{M_X}$ as 
\eqals{
Su & = (\scal{\psi(\tilde{x}_1)}{u}_\hh, \dots, \scal{\psi(\tilde{x}_{M_X})}{u}_\hh) \in \R^{M_X}, \quad S^* \alpha = \sum_{i=1}^{M_X} \alpha_i \psi(\tilde{x}_i),
}
for all $u \in \hh$ and $\alpha \in \R^{M_X}$.
Analogously define $Z:{\cal F} \to \R^{M_Y}$ as 
\eqals{
Zv & = (\scal{\Phi(\tilde{y}_1)}{v}_{\cal F}, \dots, \scal{\Phi(\tilde{y}_{M_Y})}{v}_{\cal F}) \in \R^{M_Y}, \quad Z^* \beta = \sum_{i=1}^{M_Y} \beta_i \Phi(\tilde{y}_i),
}
for all $v \in {\cal F}$ and $\beta \in \R^{M_Y}$.
Moreover, define $V, U$ as 
$$V = R^{-\top}_{\tilde X} S, \qquad V = R^{-\top}_{\tilde Y} Z.$$
\end{definition}

\begin{remark}\label{rem:operators}
We recall the following basic facts about the operator above, together with a short proof, when needed.
\begin{enumerate} 
    \item The range of $S^*$ is $\textrm{span}(\psi(x_1),\dots,\psi(x_{M_X})$,
    \item $S$ is full rank and  $SS^* = K_{\tilde{X} \tilde{X}}$,
    \item $R_{\tilde{X}}^\top R_{\tilde{X}} = K_{\tilde{X} \tilde{X}}$, since $R_{\tilde{X}}$ is the upper-triangular Cholesky of $K_{\tilde{X} \tilde{X}}$. 
    \item  $VV^* = I$, indeed, $VV^*  = R^{-\top}_{\tilde X} S S^* R^{-1}_{\tilde X}  = R^{-\top}_{\tilde X} K_{\tilde{X} \tilde{X}} R^{-1}_{\tilde X}  = R^{-\top}_{\tilde X} R_{\tilde{X}}^\top R_{\tilde{X}} R^{-1}_{\tilde X} = I.$
    \item $V$ is a partial isometry, since $VV^* = I$ and it is full rank, since it is the product of two full rank operators.
    \item $P_{\tilde{X}} = V^*V$, indeed, $V^*V$ is a projector and the range of $V^*$ is the range of $S^*$ that is $\textrm{span}(\psi(x_1),\dots,\psi(x_{M_X})$.
\end{enumerate}
For the same reasons, we have that: (a) The range of $Z^*$ is $\textrm{span}(\Phi(\tilde{y}_1),\dots,\Phi(\tilde{y}_{M_Y})$ (b) $Z$ is full rank and $ZZ^* = K_{\tilde{Y} \tilde{Y}}$ (c) $R_{\tilde{Y}}^\top R_{\tilde{Y}} = K_{\tilde{Y} \tilde{Y}}$ (d) $UU^* = I$ (e) $U$ is a partial isometry (f) $P_{\tilde{Y}} = U^*U$.
\end{remark}

Now we are ready to state the proof of \cref{lm:widehatD}.

\begin{proof}
First, note that since $\|A\|_{op}^2 = \|A^*A\|_{op}$ for any bounded linear operator $A$, we have
$$
\widehat{D}_\la = \la \|P_{\tilde{X}} F_{\mu, N}^* P_{\tilde{Y}} (P_{\tilde{Y}} G_N P_{\tilde{X}} G_N^*P_{\tilde{Y}} + \la)^{-1} P_{\tilde{Y}} F_{\mu, N} P_{\tilde{X}} \|_{op}.
$$
From \cref{rem:operators} we recall that $P_{\tilde X} = V^*V$ where $V$ is a partial isometry defined in \cref{def:operators}. Analogously $P_{\tilde Y} = U^*U$ where $V$ is a partial isometry defined in \cref{def:operators}.
Now, we have $U(U^*BU + \la I)^{-1}U^* = (B + \la I)^{-1}$ for any positive semidefinite operator $B \in \R^{M_Y \times M_Y}$, since $UU^*=I$ and so $U^*UU^* = U^*$, indeed
\eqals{
U(U^*BU + \la I)^{-1}U^*(B + \la I) &= U(U^*BU + \la I)^{-1}U^*(B + \la I)UU^* \\
& = U(U^*BU + \la I)^{-1}(U^*BU + \la U^*U)U^* \\
& = U(U^*BU + \la I)^{-1}(U^*BU + \la I)U^* \\&~~-  \la U(U^*BU + \la I)^{-1}(I - U^*U)U^* \\
& = UU^* -  \la U(U^*BU + \la I)^{-1}(U^* - U^*UU^*) = I.
}
In particular, we will use now the result above. Let $C = U G_N P_{\tilde{X}} G_N^* U^*$. So,
\eqals{
\widehat{D}_\la & = \la \|V^* VF_{\mu, N}^* U^*U(U^*CU + \la I)^{-1} U^*U F_{\mu, N} V^* V \|_{op}\\
& = \la \|V^* A^* (C + \la I)^{-1} A V \|_{op}\\
& = \la \|A^* (C + \la I)^{-1} A \|_{op}
}
where $A = U F_{\mu, N} V^*$ and we used the fact that $\|V^* T U\|_{op} = \|T\|_{op}$ for any couple of partial isometries such that $VV^* = I$ and $UU^* = I$.
By applying the definition of $U, V$ and $F_{\mu, N}$, we see that $A = \widehat{A}$ as in \cref{eq:widehatA}. 
Indeed, by expanding the definitions of $U, V$ from \cref{def:operators} and denoting by $c_i \in \R^{M_Y}$ and $d_i \in \R^{M_X}$ respectively the vectors
$c_i = Z\,\Phi(f(x_i)) = \left(k_Y(\tilde{y}_1,f(x_i)), \dots,  k_Y(\tilde{y}_{M_Y},f(x_i)) \right)$ and 
$d_i = S\psi(x_i) = \left(k_X(\tilde{x}_1,x_i), \dots,  k_X(\tilde{x}_{M_X},x_i) \right)$, we have
\eqals{
U F_{\mu, N} V^* &= \frac{v_X}{N}\sum_{i=1}^N (U\Phi(f(x_i))) \otimes (V\psi(x_i)) \mu(x_i)\\
& = \frac{v_X}{N}\sum_{i=1}^N R^{-\top}_{\tilde Y} \left((Z\Phi(f(x_i))) \otimes (S\psi(x_i))\right)R^{-1}_{\tilde X} \mu(x_i)\\
& = \frac{v_X}{N}\sum_{i=1}^N R^{-\top}_{\tilde Y} \left(c_i d_i^\top\right)R^{-1}_{\tilde X} \mu(x_i).
}
Note now, that by construction $c_i$ is the $i$-th column of $K_{\tilde{Y},f}$ while $d_i$ is the $i$-th row of the matrix $K_{X,\tilde{X}}$ for $i=1,\dots,N$. Denoting by $\diag{\hat{\mu}}$ the diagonal matrix whose $i$-th element of diagonal is $\mu(x_i)$, we have 
$$ \sum_{i=1}^N \mu(x_i) \, c_i d_i^\top   = K_{\tilde{Y},f}\diag{\hat{\mu}}K_{X,\tilde{X}}.$$
From which have 
$$
U F_{\mu, N} V^* = \frac{1}{N}R^{-\top}_{\tilde Y} K_{\tilde{Y},f}\diag{\hat{\mu}}K_{X,\tilde{X}} R^{-1}_{\tilde X} = \widehat{A}.
$$
To conclude, note that 
$$C = U G_N P_{\tilde{X}} G_N^* U^* = (U G_N V^*) (V G_N^* U^*) = (U G_N V^*)\,(U G_N V^*)^*  ,$$
Analogously as we proved that $A = \widehat{A}$, we have that $U G_N V^* = \widehat{B}$, where $\widehat{B}$ is defined in \cref{eq:widehatB}. Then
$$\widehat{D}_\la = \la \|A^* (C + \la I)^{-1} A \|_{op} = \la \|\widehat{A}^* (\widehat{B}\widehat{B}^* + \la I)^{-1} \widehat{A} \|_{op}.$$
\end{proof}

\section{Proof of \cref{thm:appr-error-widehatD}}

Before proving the theorem, we need some preliminary lemmas

\begin{lemma}\label{lm:nystrom}
Let $\delta \in (0,1)$. Let $X \subseteq \R^d$ be an open bounded set with locally Lipschitz boundary. Let $k$ be a Sobolev kernel of smoothness $m$, with $m > d/2$ on $X$ and denote by $\hh$ and $\psi:X\to\hh$ the associated RKHS and canonical feature map. Let $\tilde{X} = \{\tilde{x}_1,\dots,\tilde{x}_M\} \subset X$ be $M$ points sampled independently and uniformly at random in $X$. 
Denote by $P_{\tilde{X}}$ the projection operator whose range corresponds to $\textrm{span}\{\psi(\tilde{x}_1),\dots,\psi(\tilde{x}_M)\}$.
There exists $M_0$ such that for all $M \geq M_0$, the following holds with probability at least $1-\delta$:
\eqals{
\sup_{x \in X}\|(I-P_{\tilde{X}})\psi(x)\|_{\hh} \leq C M^{-m/d + 1/2} (\log \tfrac{C' M}{\rho})^{m/d},
}
where $C,C'$ are constants depending only on $X, m, d$.
\end{lemma}
\begin{proof}
To prove this result we use the same reasoning of Theorem C.3 of \citet{rudi2021psd}, but applied to the Sobolev kernel. First, by applying, first Lemma C.2 of \citet{rudi2021psd}  we have that
$$\sup_{x \in X}\|(I-P_{\tilde{X}})\psi(x)\|_{\hh} \leq \sup_{\|f\|_{\hh} \leq 1} \|f-P_{\tilde{X}}f\|_{L^\infty(X)}.$$
Now, denote by $\eta$ the so called {\em fill distance} \citep{narcowich2005sobolev} defined as $\eta = \sup_{x \in X}\min_{i \in {1,\dots, M}} \|x - \tilde{x}_i\|$.
By applying Proposition 3.2 of \citet{narcowich2005sobolev} with $\alpha = 0, q = \infty, \tau = m_X$, we have that  there exists an $\eta_0$ such that when $\eta \geq \eta_0$ then
$$
\|f-P_{\tilde{X}}f\|_{L^\infty(X)} \leq  C \eta^{-m+d/2}\|f\|_\hh, \quad \forall ~ f \in \hh,
$$
where $\eta_0$ and $C$ are constants depending only on $d,X,m$.
To conclude, note that, by using Lemma~11 and 12 of \citet{vacher2021dimension},
$$\eta \leq (C_1 M^{-1} \log(C_2 M/\delta))^{1/d},$$
with probability $1-\delta$, where $C_1,C_2$ depend only on $X,d$. The result is obtained by combining the three inequalities above and selecting $M_0$ as the minimum integer satisfying $(C_1 M_0^{-1} \log(C_2 M_0/\rho))^{1/d} \leq \eta_0$. 
\end{proof}

Now we are ready to prove \cref{thm:appr-error-widehatD}.

\paragraph{Proof of Theorem~\ref{thm:appr-error-widehatD}.}
\begin{proof}
Let $\rho = \delta/4$.Denote by $\kappa_X$ and $\kappa_Y$ the constants bounding the kernel $k_X, k_Y$ (which are Sobolev kernels of smoothness $s$ and $z$, see \citet{wendland2004scattered} for the explicit definition of such kernel). Note that $\kappa_X, \kappa_Y$ are constants depending, respectively, only on $s, d$ and on $z, p$.
We recall here that $v_X := \textrm{vol}(X) = \int_X dx$.

\paragraph{Step 1. }
We recall that $x_1,\dots,x_N$ are independently and uniformly distributed with uniform measure over $X$. 
Define the random variable $\zeta_i \in {\cal F} \otimes \hh$ as  
$$\zeta_i = v_X \Phi(f(x_i)) \otimes \psi(x_i) \mu(x_i)$$
for $i=1,\dots,n$. Note now, that
$$F_{\mu, N} = \frac{1}{N} \sum_{i=1}^n \zeta_i, \quad F_\mu = \mathbb{E} \zeta_1.$$
Note moreover that 
$$\|\zeta_i\| ~\leq~ v_X \kappa_X \kappa_Y \|\mu\|_{L^\infty} ~=:~ L.$$
Denote by $\scal{A}{B}_{HS}$ the Hilbert-Schmidt inner product defined as $\scal{A}{B}_{HS} = \tr(A^*B)$. We recall that the space of $HS(\hh,{\cal F})$ with finite $HS$ norm, is a separable Hilbert space. We recall also that $\|\cdot\|_{op} \leq \|\cdot\|_{HS} \leq \|\cdot\|_*$ where the last is the trace norm and that $F_{\mu, N}$ has finite trace norm.
By applying the Bernstein inequality for random vectors (see, e.g. Prop. 11 of \citet{rudi2015less} and references therein), we have that the following holds with probability $1-\rho$
\eqal{\label{eq:bound-Fmu-FmuN}
\|F_{\mu, N} - F_{\mu}\|_{HS} \leq \frac{4 L}{\sqrt{N}} \log \frac{2}{\rho}.
}
Applying the same reasoning for $G$, we obtain
\eqal{\label{eq:bound-G-GN}
\|G_N - G\|_{HS} \leq \frac{4 L'}{\sqrt{N}} \log \frac{2}{\rho},
}
with probability $1-\rho$, where $L' := \textrm{vol}(X)\kappa_X \kappa_Y.$

\paragraph{Step 2.}
Now, recall that $P_{\tilde{X}}$ is a projection operator whose range is $\textrm{span}\{\psi(\tilde{x}_1),\dots, \psi(\tilde{x}_{M_X})\}$, $X$ is bounded with Lipschitz boundary and $k_X$ is a Sobolev kernel of smoothness $m_X$. 
By applying \cref{lm:nystrom}, we have that there exists $C_0$ such that, when $M_X \geq C_0$, then with probability $1-\rho$
\eqal{\label{eq:bound-nystr-X}
\sup_{x \in X}\|(I-P_{\tilde{X}})\psi(x)\|_{\hh} \leq C_1 M_X^{-s/d} (\log \tfrac{C_2 M_X}{\rho})^{s/d+1/2},
}
where $C_0, C_1, C_2$ depend only on $X, s, d$. Applying the same reasoning on $P_{\tilde{Y}}$, we have that there exists $C'_0$ such that, when $M_Y \geq C'_0$, then with probability $1-\rho$
\eqal{\label{eq:bound-nystr-Y}
\sup_{y \in Y}\|(I-P_{\tilde{Y}})\psi(y)\|_{\cal F} \leq C'_1 M_Y^{-z/p} (\log \tfrac{C'_2 M_Y}{\rho})^{z/p+1/2},
}
where $C'_0, C'_1, C'_2$ depend only on $Y, z, p$.

\paragraph{Step 3.}
Now we can estimate the distance between $F_\mu$ and $P_{\tilde{Y}}F_{\mu,N} P_{\tilde{X}}$ and, analogously between $G$ and $P_{\tilde{Y}}G_N P_{\tilde{X}}$. In particular, we can rewrite $F_\mu - P_{\tilde{Y}}F_{\mu,N} P_{\tilde{X}}$ as
\eqals{
F_\mu - P_{\tilde{Y}}F_{\mu,N} P_{\tilde{X}} \leq (F_\mu - F_{\mu,N}) + (I -  P_{\tilde{Y}})F_{\mu,N} + P_{\tilde{Y}}F_{\mu,N}(I - P_{\tilde{X}}),
}
from which, using $\|\cdot\|_{op} \leq \|\cdot\|_{HS}$, we derive
\eqals{
\|F_\mu - P_{\tilde{Y}}F_{\mu,N} P_{\tilde{X}}\|_{op} \leq \|F_\mu - F_{\mu,N}\|_{HS} + \|(I -  P_{\tilde{Y}})F_{\mu,N}\|_{op} + \|P_{\tilde{Y}}\|_{op}\|F_{\mu,N}(I - P_{\tilde{X}})\|_{op}.
}
Now, the term $\|F_\mu - F_{\mu,N}\|_{HS}$ is already studied in \cref{eq:bound-Fmu-FmuN}. For the second term, note that by expanding the definition of $F_{\mu,N}$ and using \cref{eq:bound-nystr-X} we obtain,
\eqals{
\|(I -  P_{\tilde{Y}})F_{\mu,N}\|_{op} &\leq \frac{v_X}{N} \sum_{i=1}^N \|(I -  P_{\tilde{Y}})(\Phi(f(x_i)) \otimes \psi(x_i))\|_{op} |\mu(x_i)|  \leq \\
& \frac{v_X}{N} \sum_{i=1}^N \|(I -  P_{\tilde{Y}})\Phi(f(x_i))\|_{\cal F} \|\psi(x_i)\|_\hh |\mu(x_i)|\\
& \leq \kappa_X v_X \|\mu\|_{L^\infty} C'_1 M_Y^{-z/p} (\log \tfrac{C'_2 M_Y}{\rho})^{z/p+1/2},
}
with probability $1-\rho$.
Applying the same reasoning to the third term, we obtain
\eqals{
\|F_{\mu,N}(I - P_{\tilde{X}})\|_{op} \leq \kappa_Y v_X \|\mu\|_{L^\infty} C_1 M_X^{-s/d} (\log \tfrac{C_2 M_X}{\rho})^{s/d+1/2},
}
with probability $1-\rho$.
Combining all the terms and considering that $\|P_{\tilde{X}}\|_{op} = 1$ since it is a projection, we have
\eqal{\label{eq:BB}
\|F_\mu - P_{\tilde{Y}}F_{\mu,N} P_{\tilde{X}}\|_{op} \leq \beta
}
with 
\eqals{ 
	\beta := \frac{4 L}{\sqrt{N}} \log \frac{2}{\rho} &+ \kappa_X \|\mu\|_{L^\infty} C'_1 M_Y^{-z/p} (\log \tfrac{C'_2 M_Y}{\rho})^{z/p+1/2} \\&+ \kappa_Y \|\mu\|_{L^\infty} C_1 M_X^{-s/d} (\log \tfrac{C_2 M_X}{\rho})^{s/d+1/2}.
}

To conclude, note that
$$\|F_{\mu}\|_{op} \leq \int \|\Phi(f(x))\|_{op}\|\psi(x)\|_{op} |\mu(x)| dx \leq v_X \kappa_X \kappa_Y \|\mu\|_{L^\infty(X)} = L,$$
and with the same reasoning we have $\|F_{\mu,N}\| \leq L$. Then, by considering that $\|AA^* - \hat{A}\hat{A}^*\| \leq (\|A\|_{op} + \|\hat{A}\|_{op})\|A-\hat{A}\|_{op}$ for any bounded operators $A,A^*$ between the same two Hilbert spaces, we have
$$
\|F_\mu F_{\mu}^* - P_{\tilde{Y}}F_{\mu,N} P_{\tilde{X}} F_{\mu,N}^*P_{\tilde{Y}}\|_{op} \leq 2L \beta.
$$
Repeating the same reasoning of the beginning of Step 3 for $G$ and $G_N$ we obtain
\eqal{\label{eq:QQ}
\|GG^* - P_{\tilde{Y}}G P_{\tilde{X}} G^*P_{\tilde{Y}}\|_{op} \leq 2L' \beta',
}
where
\eqals{
	\beta' := \frac{4 L'}{\sqrt{N}} \log \frac{2}{\rho} &+ \kappa_X \textrm{vol}(X) C'_1 M_Y^{-z/p} (\log \tfrac{C'_2 M_Y}{\rho})^{z/p+1/2} \\&+ \kappa_Y \textrm{vol}(X) C_1 M_X^{-s/d} (\log \tfrac{C_2 M_X}{\rho})^{s/d+1/2}.
}

\paragraph{Step 4.}
Before deriving the final result, we need an algebraic inequality between bounded operators. Let $B,\hat{B},Q,\hat{Q}$ be bounded operators and assume that $Q,\hat{Q}$ are also symmetric and invertible. We recall that $\|A\|^2_{op} = \|A^*A\|_{op} = \|AA^*\|_{op}$ for any bounded operator $A$. We have
\eqals{
\|Q^{-1/2} B\|^2_{op} - \|\hat{Q}^{-1/2} \hat{B}\|^2_{op} = (\|Q^{-1/2} B\|^2_{op} - \|\hat{Q}^{-1/2} B\|^2_{op}) +(\|\hat{Q}^{-1/2} B\|^2_{op} - \|\hat{Q}^{-1/2} \hat{B}\|^2_{op}).
}
Then, using the equality $A^{-1} - B^{-1} = B^{-1}(A - B)A^{-1}$, valid for any bounded and invertible operator, we have
\eqals{
|\|Q^{-1/2} B\|^2_{op} - \|\hat{Q}^{-1/2} B\|^2_{op}| &= |\|B^* Q^{-1} B\|_{op} - \|B^*\hat{Q}^{-1} B\|_{op}| \\
& \leq \|B^* (Q^{-1} - \hat{Q}^{-1}) B\|_{op} = \|B^* \hat{Q}^{-1} (Q - \hat{Q})Q^{-1} B\|_{op}\\
& \leq \|B^*\|_{op} \|\hat{Q}^{-1}\|_{op} \|Q - \hat{Q}\|_{op} \|Q^{-1}\|_{op} \|B\|_{op}.
}
moreover, we have 
\eqals{
|\|\hat{Q}^{-1/2} B\|^2_{op} - \|\hat{Q}^{-1/2} \hat{B}\|^2_{op}| & = |\|\hat{Q}^{-1/2} BB^*\hat{Q}^{-1/2}\|^2_{op} - \|\hat{Q}^{-1/2} \hat{B}\hat{B}^*\hat{Q}^{-1/2} \|_{op}| \\
& \leq \|\hat{Q}^{-1/2} (BB^* - \hat{B}\hat{B}^*) \hat{Q}^{-1/2} \|_{op} \\
& \leq \|\hat{Q}^{-1/2}\|^2_{op} \|BB^* - \hat{B}\hat{B}^*\|_{op}.
}
Now, noting that $\|\hat{Q}^{-1/2}\|^2_{op} = \|\hat{Q}^{-1}\|_{op}$ and combining the inequalities above, we obtain
\eqal{\label{eq:alg-ineq}
|\|Q^{-1/2} B\|^2_{op} - \|\hat{Q}^{-1/2} \hat{B}\|^2_{op}| \leq \|B\|^2_{op} \|\hat{Q}^{-1}\|_{op}\|Q^{-1}\|_{op} \|Q - \hat{Q}\|_{op} + \|\hat{Q}^{-1}\|_{op} \|BB^* - \hat{B}\hat{B}^*\|_{op}.
}

\paragraph{Step 5.}
With the tools derived above we can proceed to bound $\widehat{D}_\la(f,g) - D_\la(f,g)$. 
First, note that, by \cref{lm:widehatD}, $\widehat{D}_\la(f,g)$ of \cref{eq:widehatD-finite-dimensional} is equivalent to 
\eqals{
\widehat{D}_\la(f,g) := \la \|(P_{\tilde{Y}} G_N P_{\tilde{X}} G_N^*P_{\tilde{Y}} + \la)^{-\frac{1}{2}} P_{\tilde{Y}} F_{\mu, N} P_{\tilde{X}} \|^2_{op}.
}
Then,
\eqals{ 
|D_\la(f,g) - \widehat{D}_\la(f,g)| \leq = \la | \|(GG^* + \la)^{-\frac{1}{2}} F_{\mu}\|^2_{op} - \|(P_{\tilde{Y}} G_N P_{\tilde{X}} G_N^*P_{\tilde{Y}} + \la)^{-\frac{1}{2}} P_{\tilde{Y}} F_{\mu, N} P_{\tilde{X}}\|^2_{op}|.
}
By applying \cref{eq:alg-ineq} with $Q = GG^* + \la$, $\hat{Q} = P_{\tilde{Y}} G_N P_{\tilde{X}} G_N^*P_{\tilde{Y}} + \la$, $B = F_{\mu}$ and $\hat{B} = P_{\tilde{Y}} F_{\mu, N} P_{\tilde{X}}$ and noting that $Q \succeq \la I$ and $\hat{Q} \succeq \la I$, then $\|Q^{-1}\|_{op} \leq \la^{-1}$ and $\|\hat{Q}^{-1}\|_{op} \leq \la^{-1}$,
and that $\|F_{\mu}\|_{op} \leq L$, we have
\eqals{
|D_\la(f,g) - \widehat{D}_\la(f,g)| &\leq \la \|B\|^2_{op} \|\hat{Q}^{-1}\|_{op}\|Q^{-1}\|_{op} \|Q - \hat{Q}\|_{op}  + \|\hat{Q}^{-1}\|_{op} \|BB^* - \hat{B}\hat{B}^*\|_{op} \\
& \leq L^2 \la^{-1} \|Q - \hat{Q}\|_{op}  + \|BB^* - \hat{B}\hat{B}^*\|_{op}.
}
Now, since $\|Q - \hat{Q}\|_{op} = \|GG^* - P_{\tilde{Y}}G_{N} P_{\tilde{X}}\|_{op}$ and  $\|BB^* - \hat{B}\hat{B}^*\|_{op} = \|F_\mu - P_{\tilde{Y}}F_{\mu,N} P_{\tilde{X}}\|_{op}$, which are bounded in \cref{eq:QQ,eq:BB}
\eqals{
|D_\la(f,g) - \widehat{D}_\la(f,g)| \leq 2L'L^2 \la^{-1} \beta'  + 2L\beta.
}
\end{proof}

\section{Computing the optimal $h$ and $q$}\label{sec:appendix-hq}

Using the finite-rank approximate described in the paper, we observe $h$ and $q$ using the following formulas, where $[h] = [h(x_1), \ldots, h(x_N)]$ and $[q] = [q(x_1), \ldots, q(x_N)]$, the values at grid points (discretization of the integral):
\eqals{
[h] &= K_{X \widetilde X} R_X^{-1/2}\widetilde h,\\
[q] &= K_{X \widetilde X} R_X^{-1/2}\widehat G^* (\widehat G\widehat G^* + \lambda I)^{-1}\widehat F\widetilde h.
}
where $\widetilde h$ is computed as a byproduct of the computation of $\widehat D$ (for example, power-iteration).
\section{Information for reproducing experiments}

\paragraph{Dependencies}
We conduct our experiments using Pytorch version 1.9.0 (torchvision 0.10.0) and Numpy 1.20.1 (but our implementation does not require any special functions so should generalize to any recent version). We also make use of Kornia version 0.5.7 and PIL version 8.2.0 for IO.

\paragraph{Code}

Our source code is organized as follows:

\bi 
    \item \texttt{warping} (dir): contains the necessary code for generating random warps. Code modified from \citet{wyartdiffeo} at \href{https://github.com/pcsl-epfl/diffeomorphism}{https://github.com/pcsl-epfl/diffeomorphism}.
    \item \texttt{did} (dir): implements objects necessary for the computation of \Diffy.
    \item \texttt{imagenet.py}: implements as ImageFolder torchvision dataset for Imagenet (and Imagenette).
    \item \texttt{scenes} (dir), \texttt{perspective} (dir), \texttt{peppers.mat}: provide images taken us for the experiments (the \texttt{.mat} is a Matlab binary object which can be read thanks to \texttt{scipy}).
    \item \texttt{demo\char`_XXX.py} are used for the demonstrations in \cref{fig:example}. They are the easiest way to get familiar with \Diffy.
    \item \texttt{exp\char`_XXX.py} are used for the different experiments quantitative experiments.
\ei

Note that the path to the Imagenette dataset must be set in each experiment file or in the \texttt{imagenet.py} source file.

\paragraph{Imagenette} We use images from the Imagenet dataset for our experiments. We use the subset called Imagenette, available at: \href{https://github.com/fastai/imagenette}{https://github.com/fastai/imagenette}.

\paragraph{ImageNet statistics}
We use the following Imagenet statistics to normalize \emph{all} images:

$$ \mu = [0.485, 0.456, 0.406]$$
$$ \sigma =[0.229, 0.224, 0.225]$$

\clearpage
\section{\Diffy in action (supplementary experiments)} \label{app:additional-experiments}

\begin{figure}[!h]
    \centering
    \includegraphics[width=0.5\textwidth]{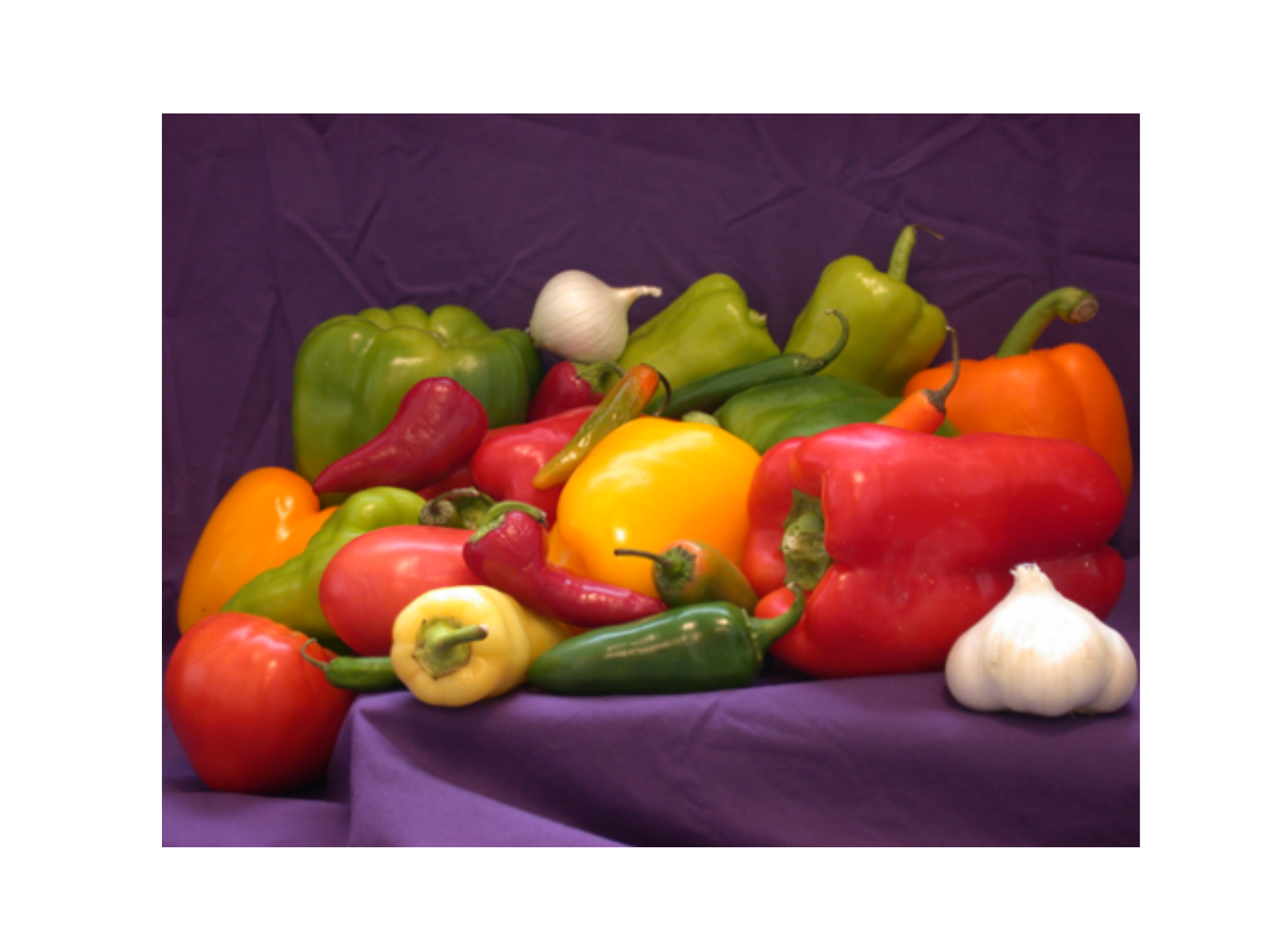}
    \caption{Peppers image from Matlab software.}
    \label{fig:peppers}
\end{figure}

We propose a collection of experiments with \Diffy on the \texttt{peppers} (see \cref{fig:peppers}) image (with random square patches). We show the values \Diffy takes and the shapes of the optimal $h$ and $q$ for a choice of $\lambda$.

\begin{figure}[!h]
    \centering
    \begin{subfigure}{0.60\textwidth}
    \includegraphics[width=\textwidth]{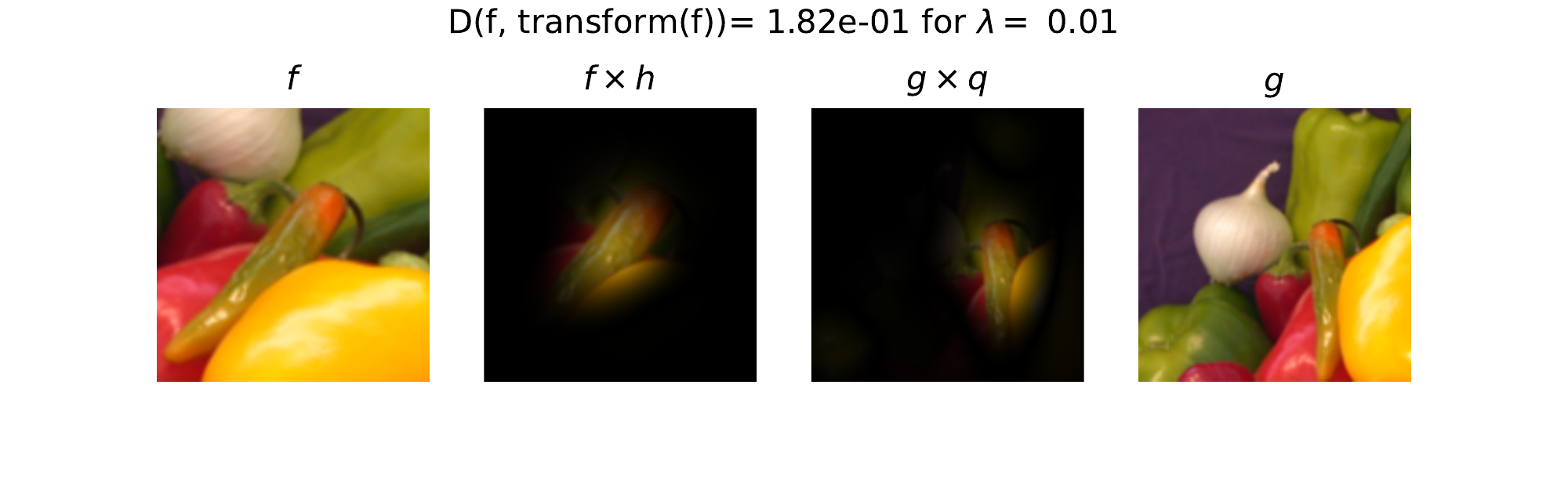}
    \end{subfigure}
    \begin{subfigure}{0.60\textwidth}
    \includegraphics[width=\textwidth]{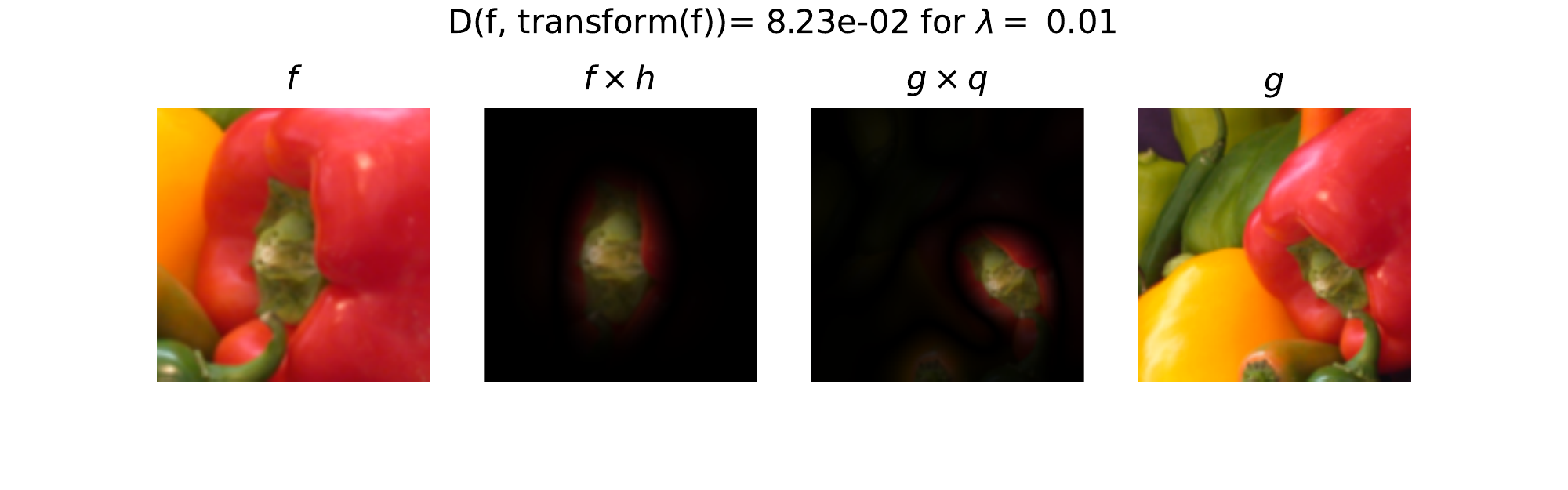}
    \end{subfigure}
    \begin{subfigure}{0.60\textwidth}
    \includegraphics[width=\textwidth]{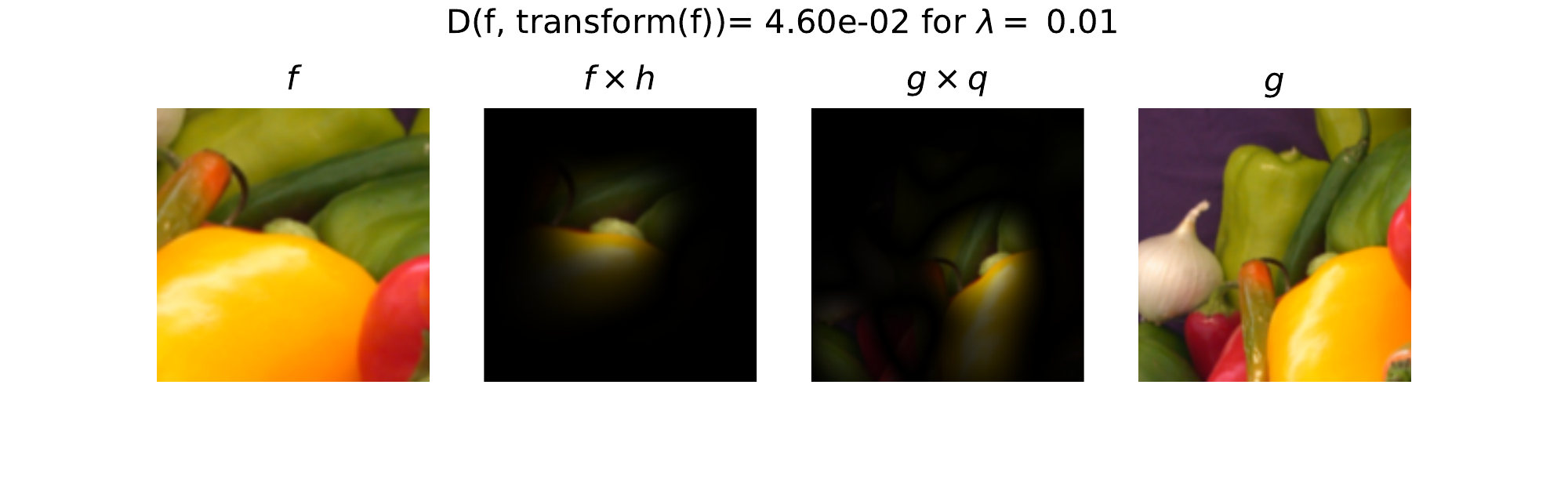}
    \end{subfigure}
    \begin{subfigure}{0.60\textwidth}
    \includegraphics[width=\textwidth]{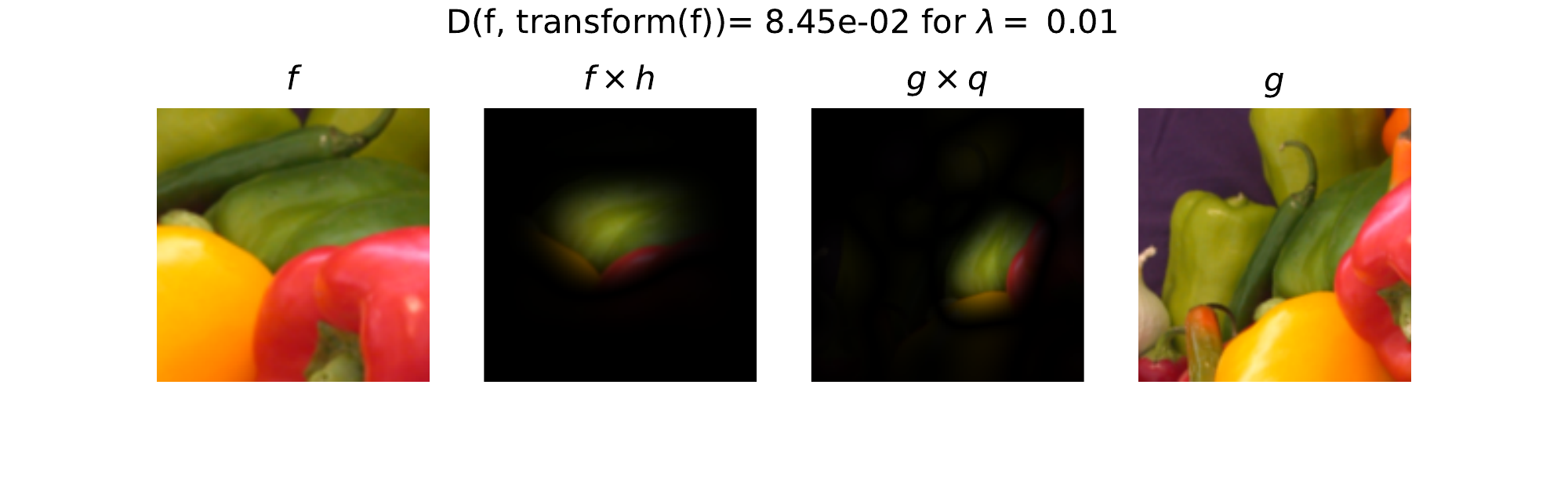}
    \end{subfigure}
    \begin{subfigure}{0.60\textwidth}
    \includegraphics[width=\textwidth]{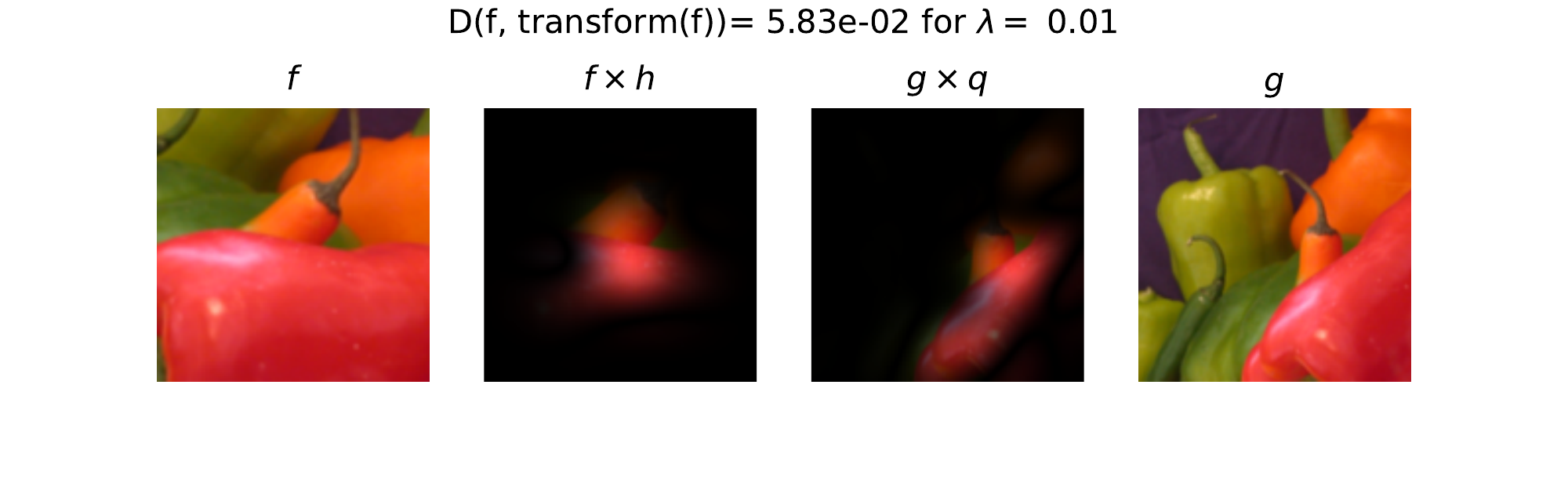}
    \end{subfigure}
    \begin{subfigure}{0.60\textwidth}
    \includegraphics[width=\textwidth]{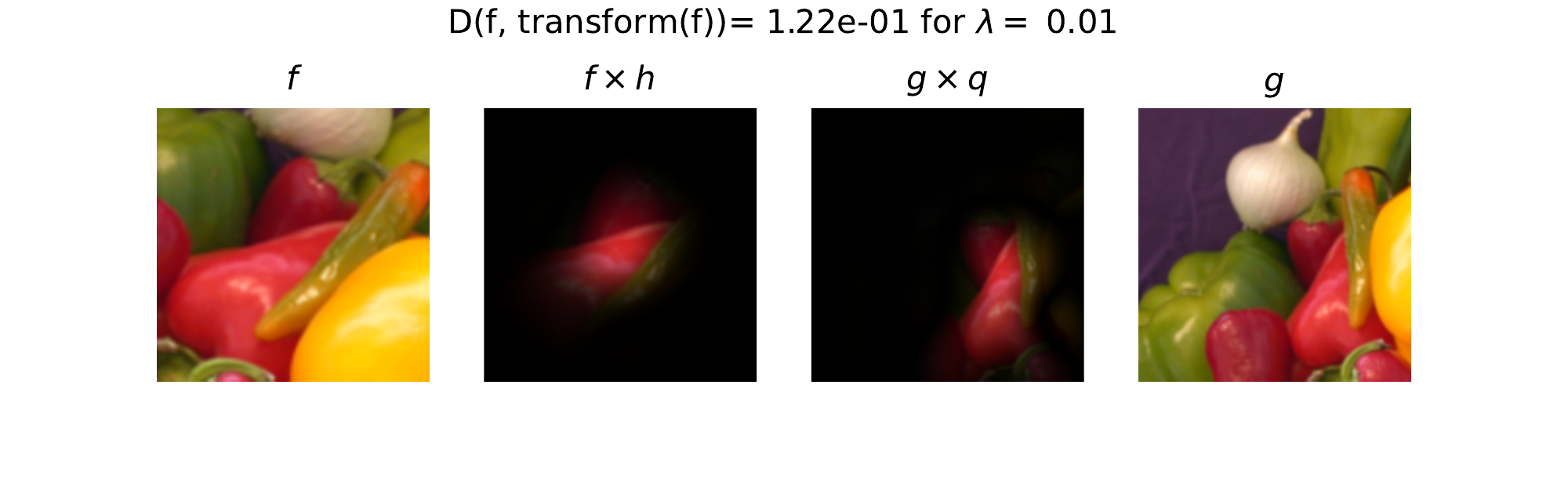}
    \end{subfigure}
    \caption{$\widehat D_\lambda(f, \textrm{transform}(f))$ for $f$ random patches from \texttt{peppers}. We show $f$, $g=\textrm{transform}(f)$ as well as the optimal functions $h$ and $q$.}
    \label{fig:appendix-peppers-matching}
\end{figure}

\Diffy has parameters: $M_X = 100$, $M_Y = 16^3$, $k_X$ Gaussian with $\sigma = 1/6$ and $k_Y$ Abel with $a=5$. We ran several experiments and chose $\lambda = 10^{-2}$ as the ``best'' value. Indeed, as shown in \cref{fig:appendix-peppers-matching}, the regions found by $h$ and $q$ for this value are coherent. 

We consider a random square area $f$ of size $150 \times 150$. We rotate, translate and scale it and denote it $g = \textrm{transform}(f)$. We then show $f$, $h$ (over $f$), $g$ and $q$ (over $g$) as well as the value of $\widehat D_\lambda(f, g)$. Experiments in \cref{fig:appendix-peppers-matching} can be reproduced with \texttt{appendix\char`_peppers\char`_match.py}.

\clearpage
\section{Warping demonstrations}\label{sec:appendix-warping}
The warps below were generated using file \texttt{appendix\char`_warp.py} in our source code, using an image from ImageNet. Rows are from different samples, columns for different warp temperatures. All waps use $c=2$.

\begin{figure}[!h]
    \centering
    \begin{subfigure}{0.18\textwidth}
    \includegraphics[width=\textwidth]{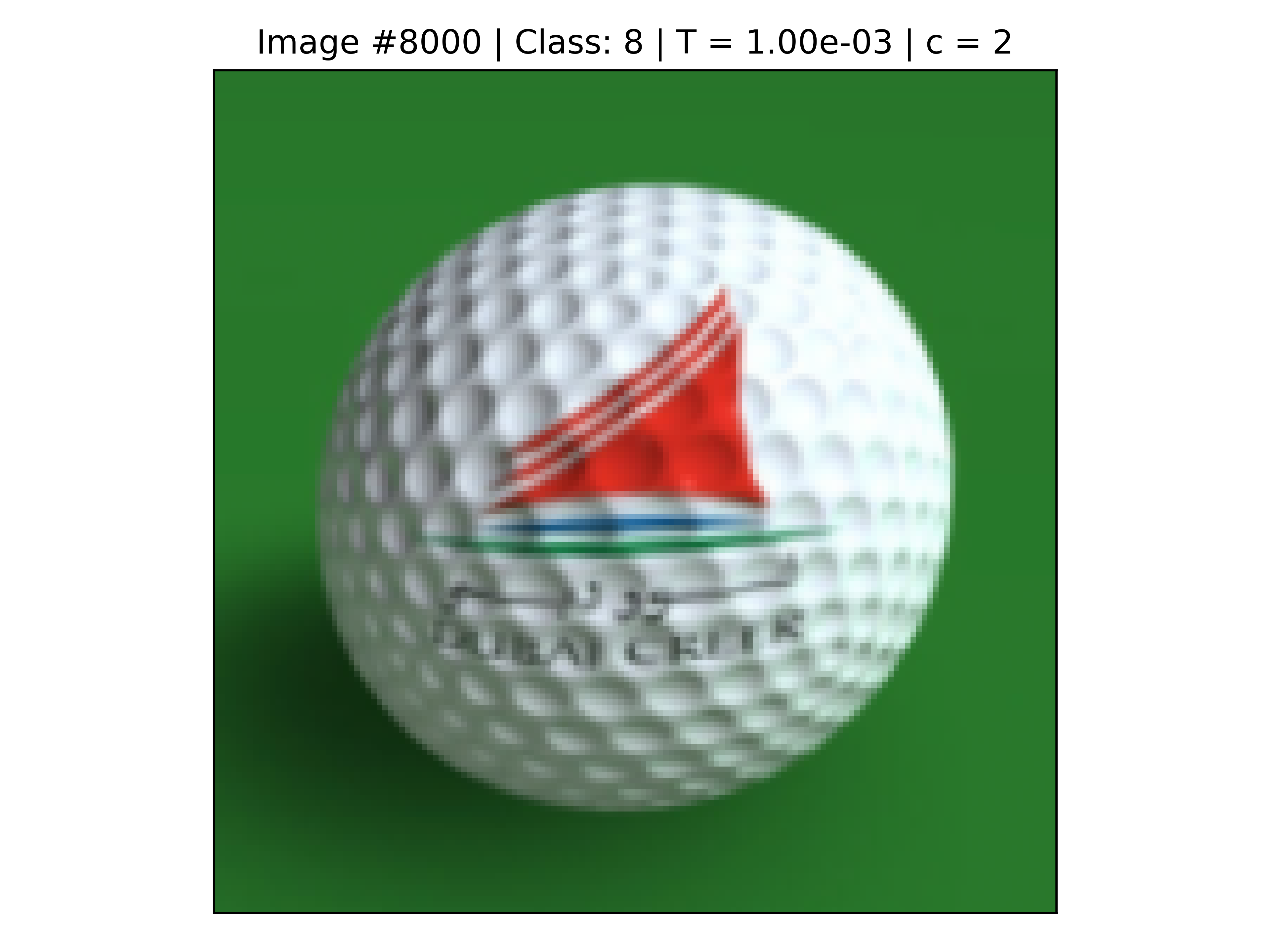}
    \caption{$T=10^{-3}$}
    \end{subfigure}
    \begin{subfigure}{0.18\textwidth}
    \includegraphics[width=\textwidth]{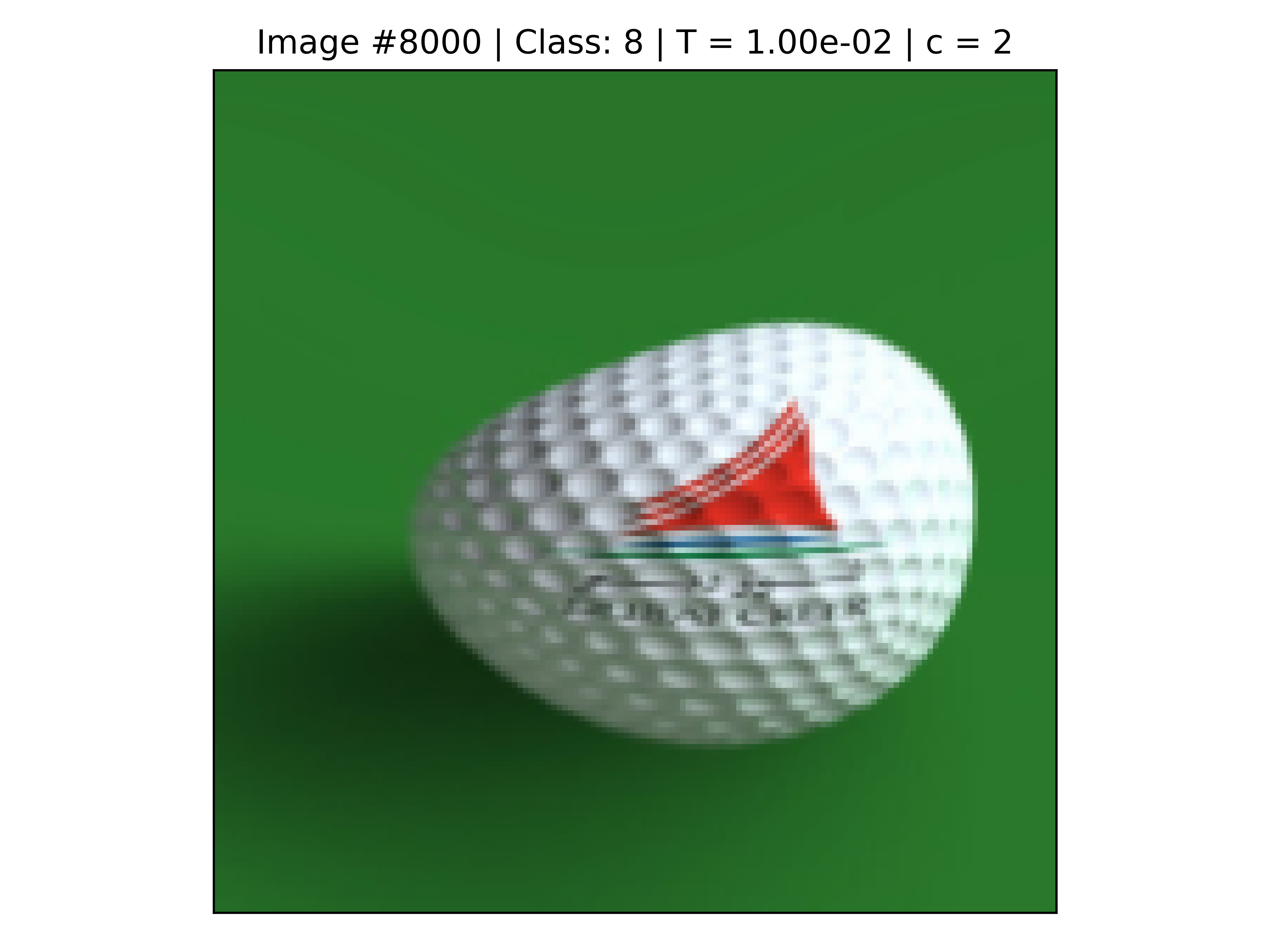}
    \caption{$T=10^{-2}$}
    \end{subfigure}
    \begin{subfigure}{0.18\textwidth}
    \includegraphics[width=\textwidth]{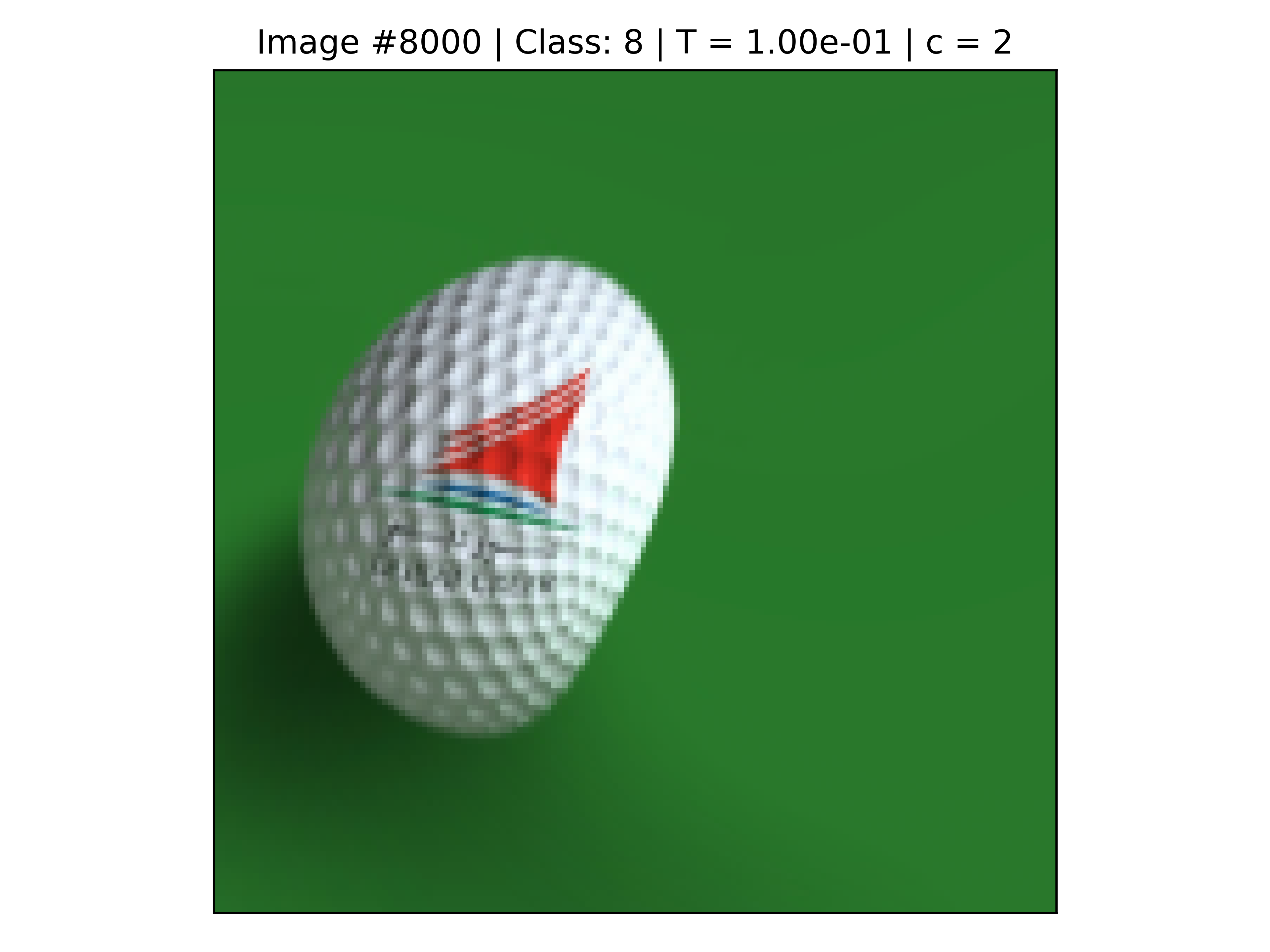}
    \caption{$T=10^{-1}$}
    \end{subfigure}
    \begin{subfigure}{0.18\textwidth}
    \includegraphics[width=\textwidth]{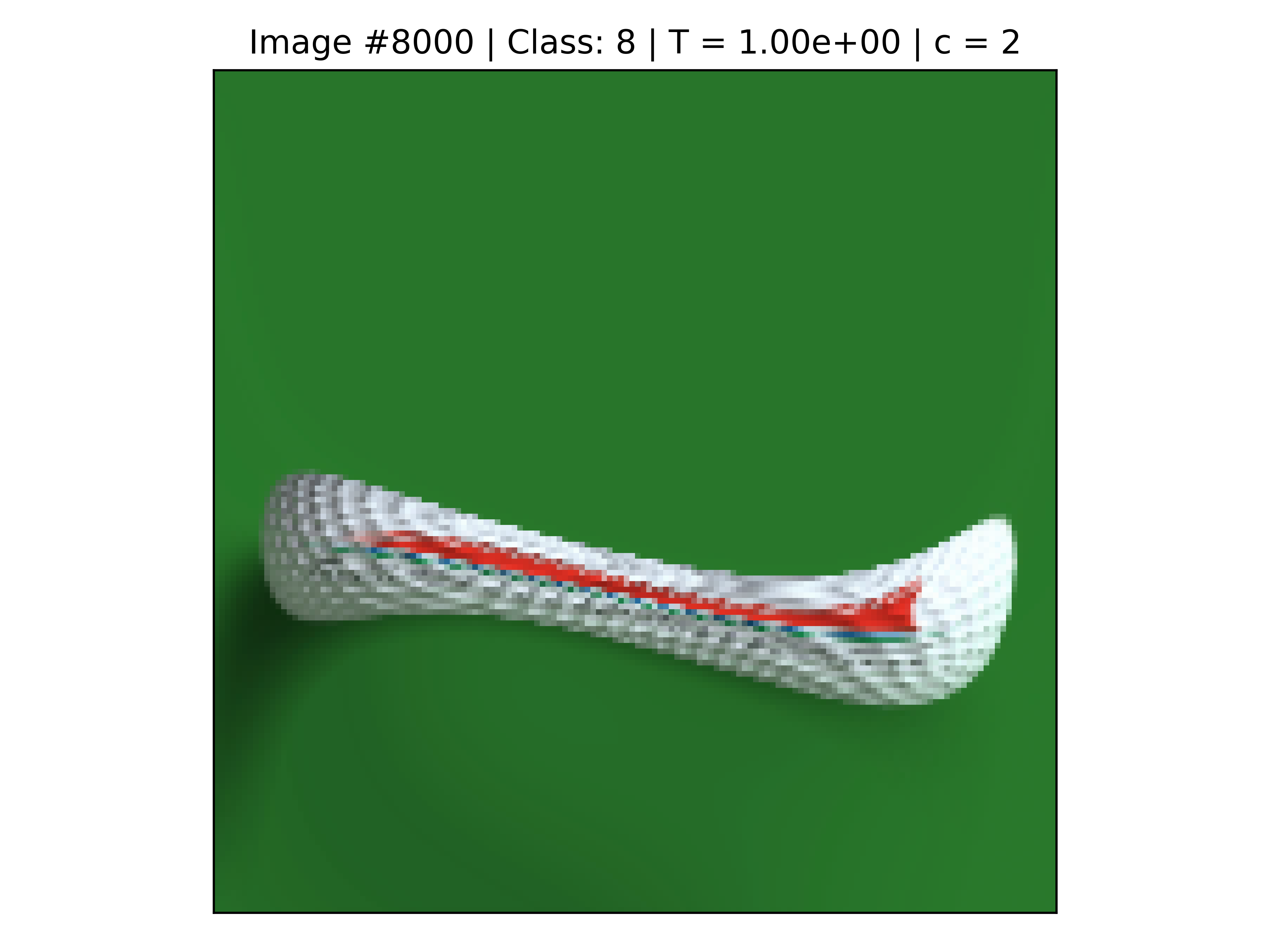}
    \caption{$T=1$}
    \end{subfigure}
    \begin{subfigure}{0.18\textwidth}
    \includegraphics[width=\textwidth]{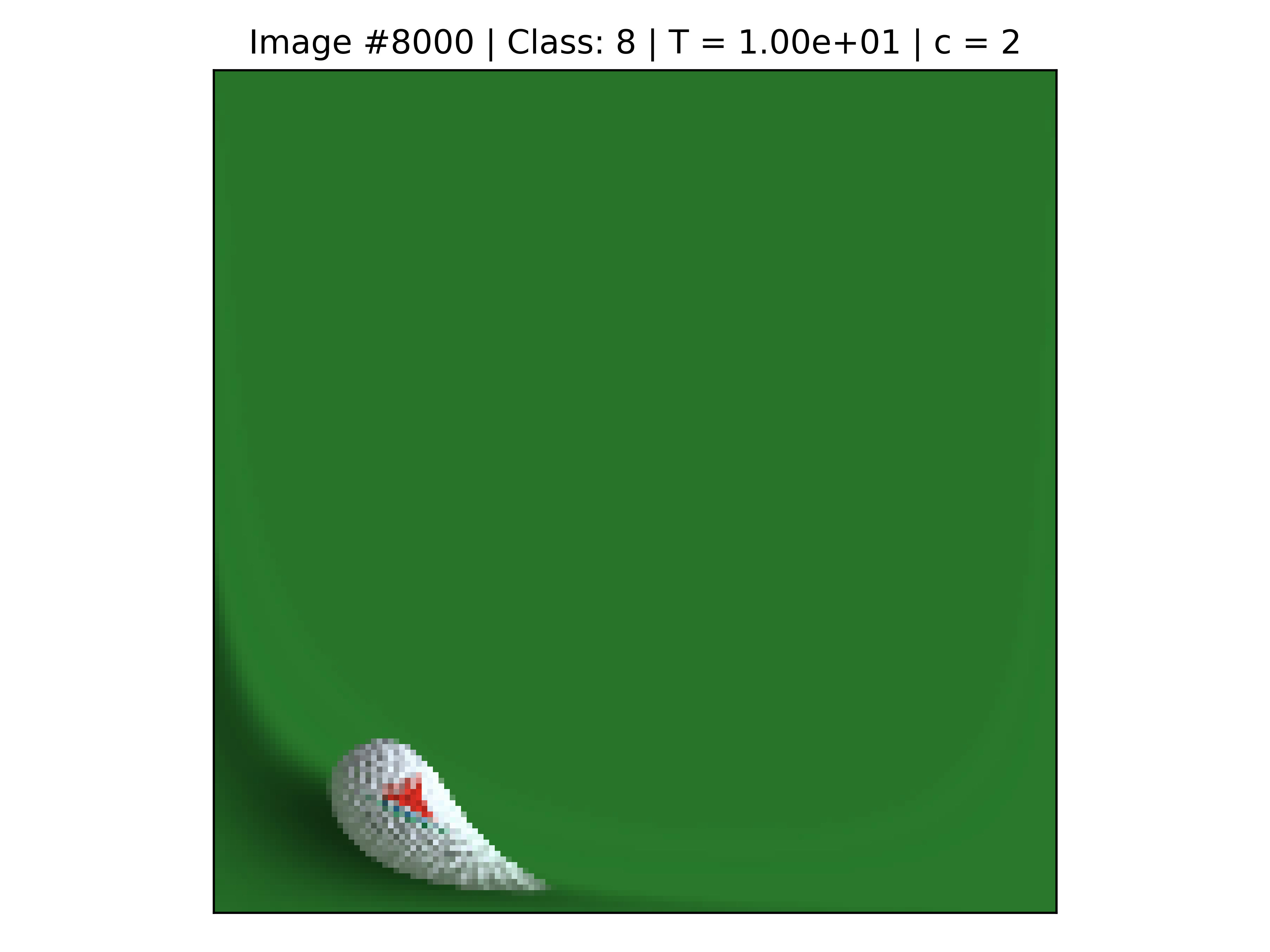}
    \caption{$T=10$}
    \end{subfigure}
    \begin{subfigure}{0.18\textwidth}
    \includegraphics[width=\textwidth]{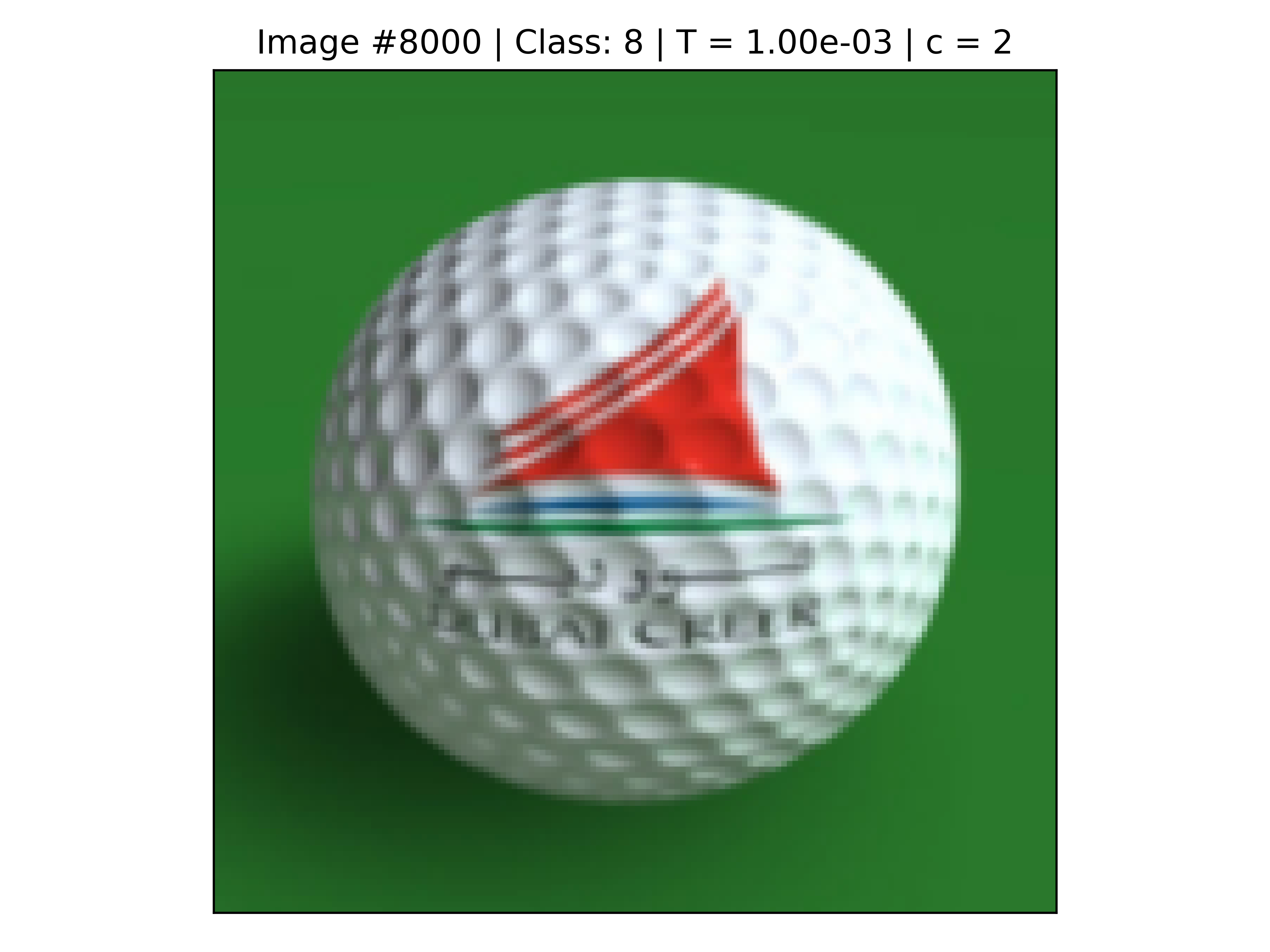}
    \caption{$T=10^{-3}$}
    \end{subfigure}
    \begin{subfigure}{0.18\textwidth}
    \includegraphics[width=\textwidth]{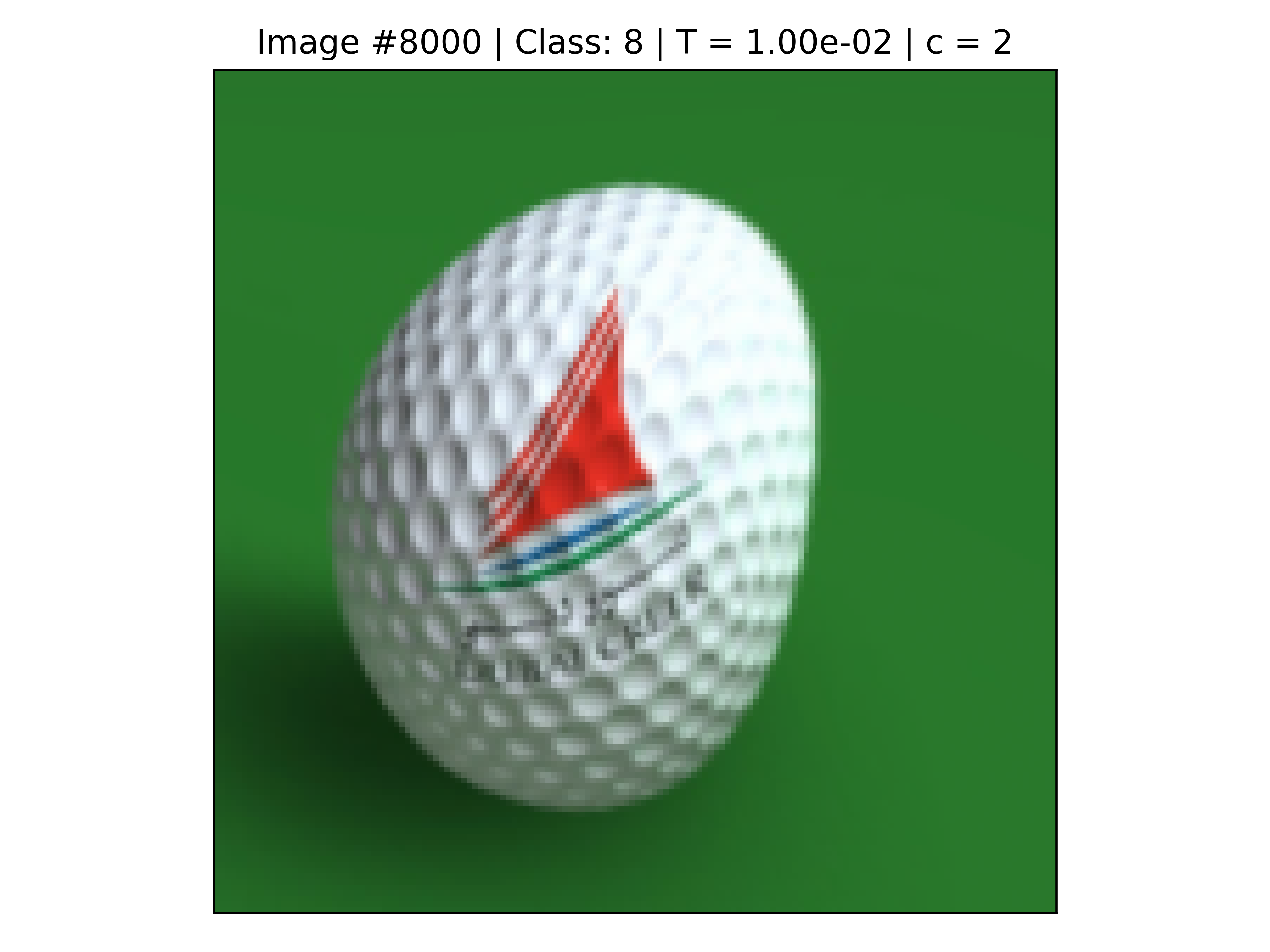}
    \caption{$T=10^{-2}$}
    \end{subfigure}
    \begin{subfigure}{0.18\textwidth}
    \includegraphics[width=\textwidth]{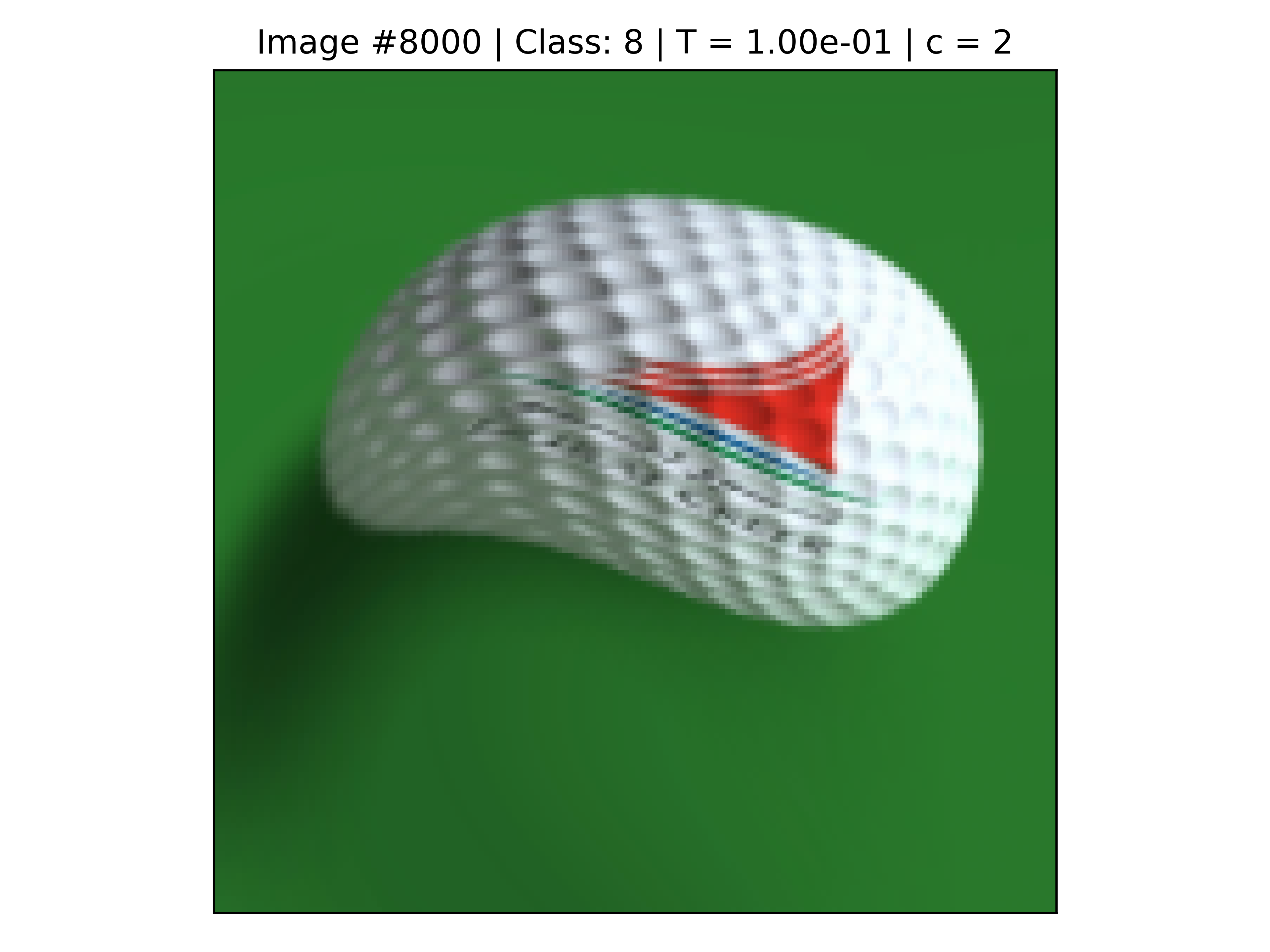}
    \caption{$T=10^{-1}$}
    \end{subfigure}
    \begin{subfigure}{0.18\textwidth}
    \includegraphics[width=\textwidth]{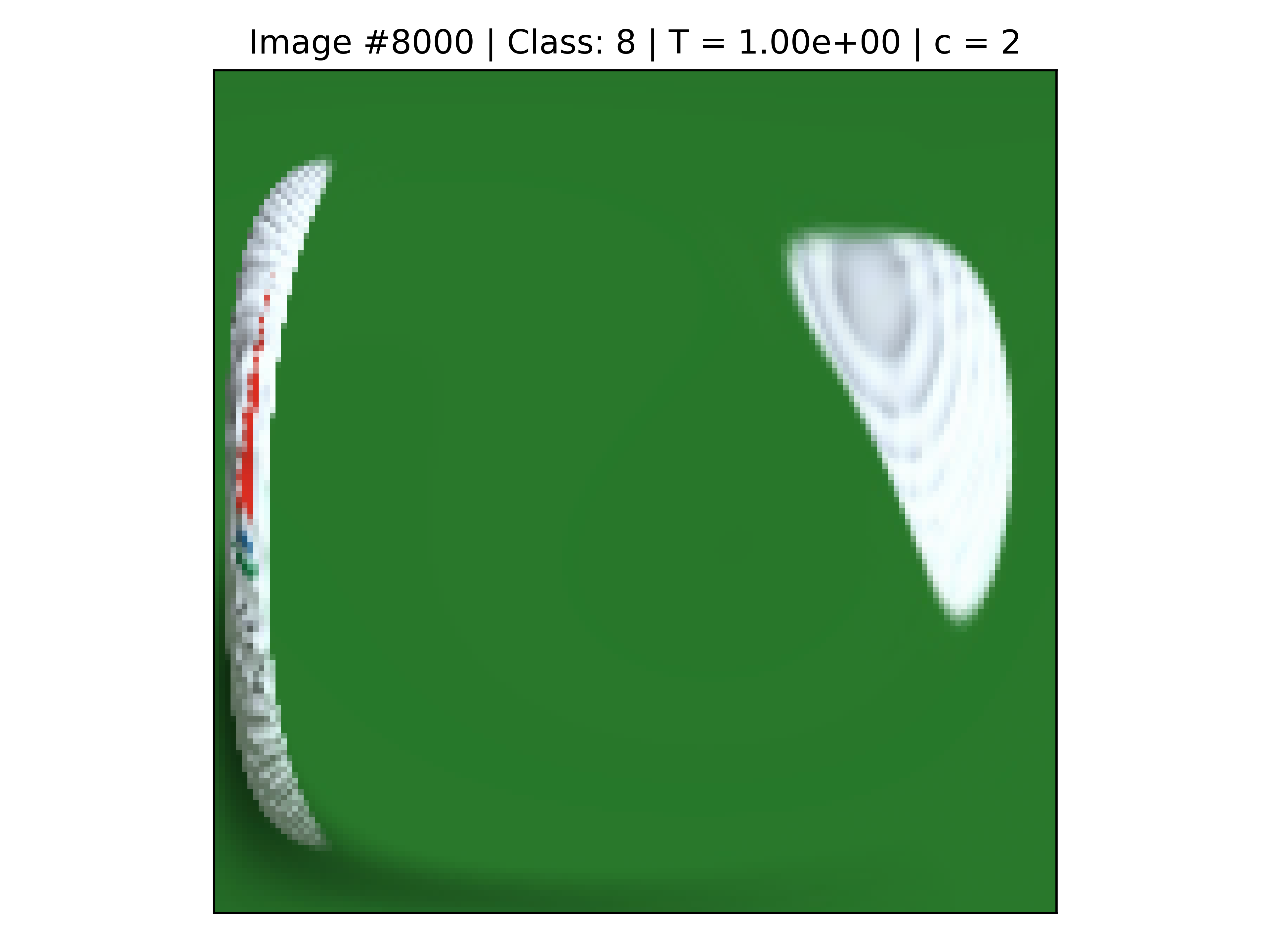}
    \caption{$T=1$}
    \end{subfigure}
    \begin{subfigure}{0.18\textwidth}
    \includegraphics[width=\textwidth]{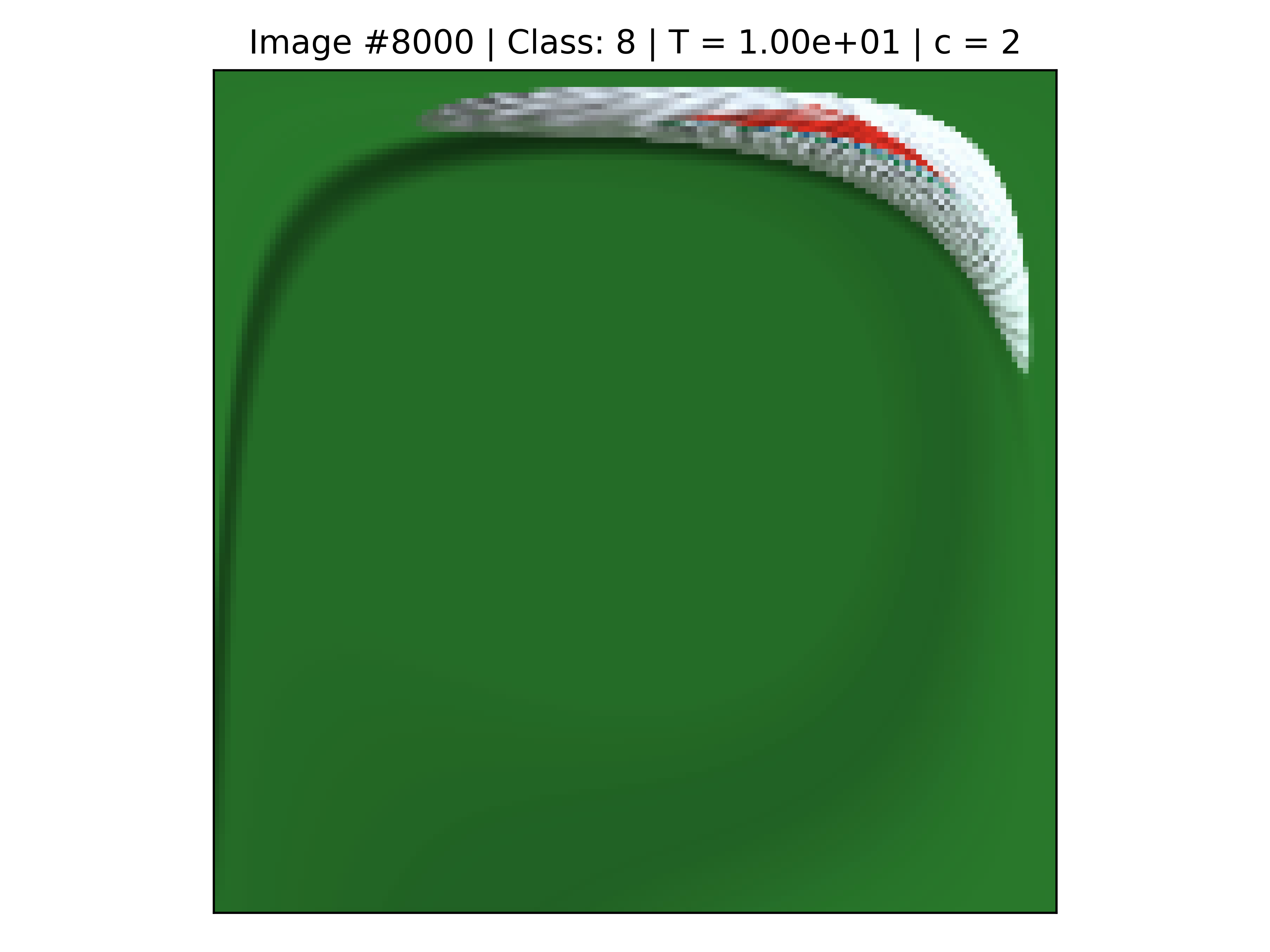}
    \caption{$T=10$}
    \end{subfigure}
    \begin{subfigure}{0.18\textwidth}
    \includegraphics[width=\textwidth]{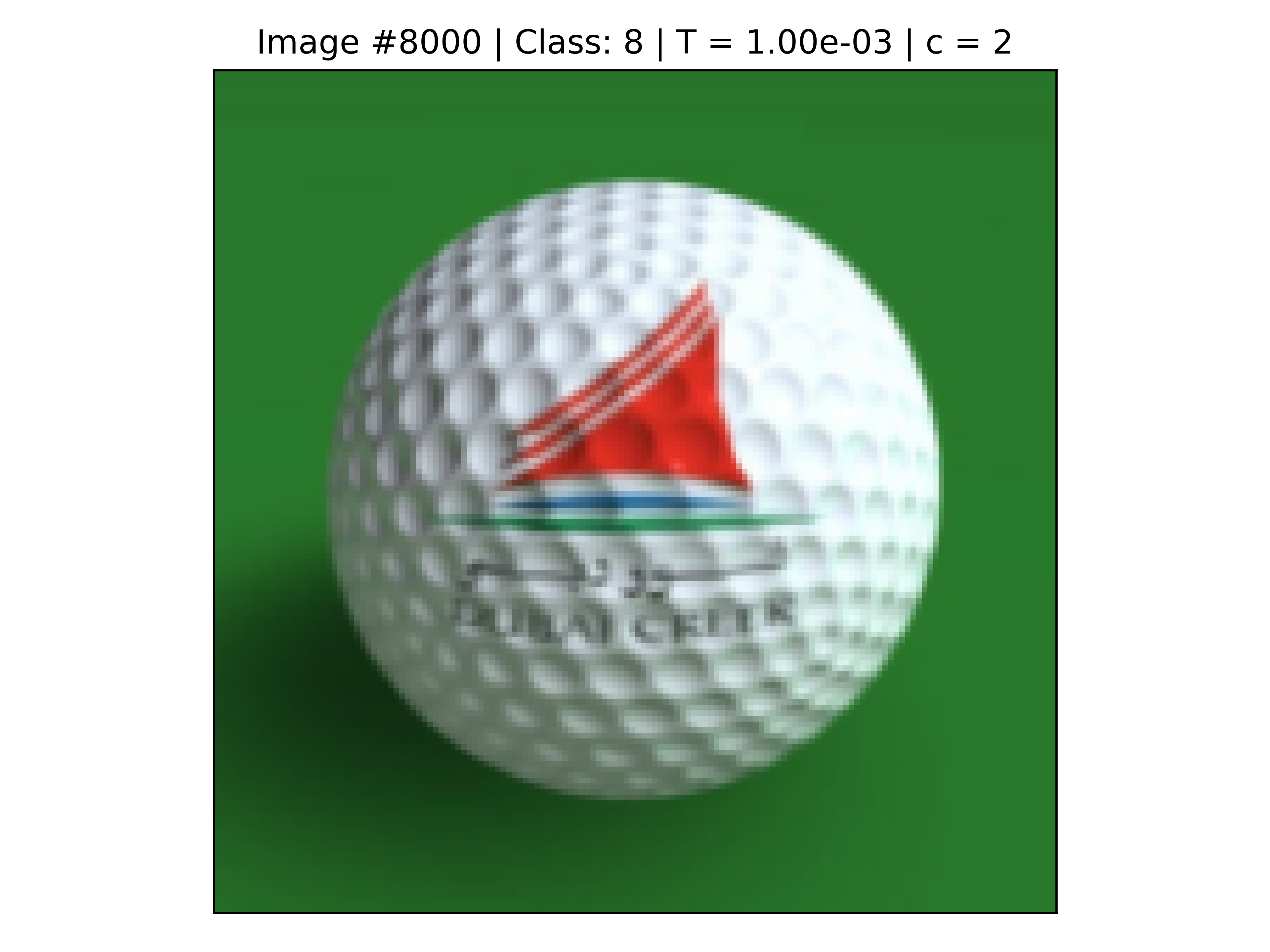}
    \caption{$T=10^{-3}$}
    \end{subfigure}
    \begin{subfigure}{0.18\textwidth}
    \includegraphics[width=\textwidth]{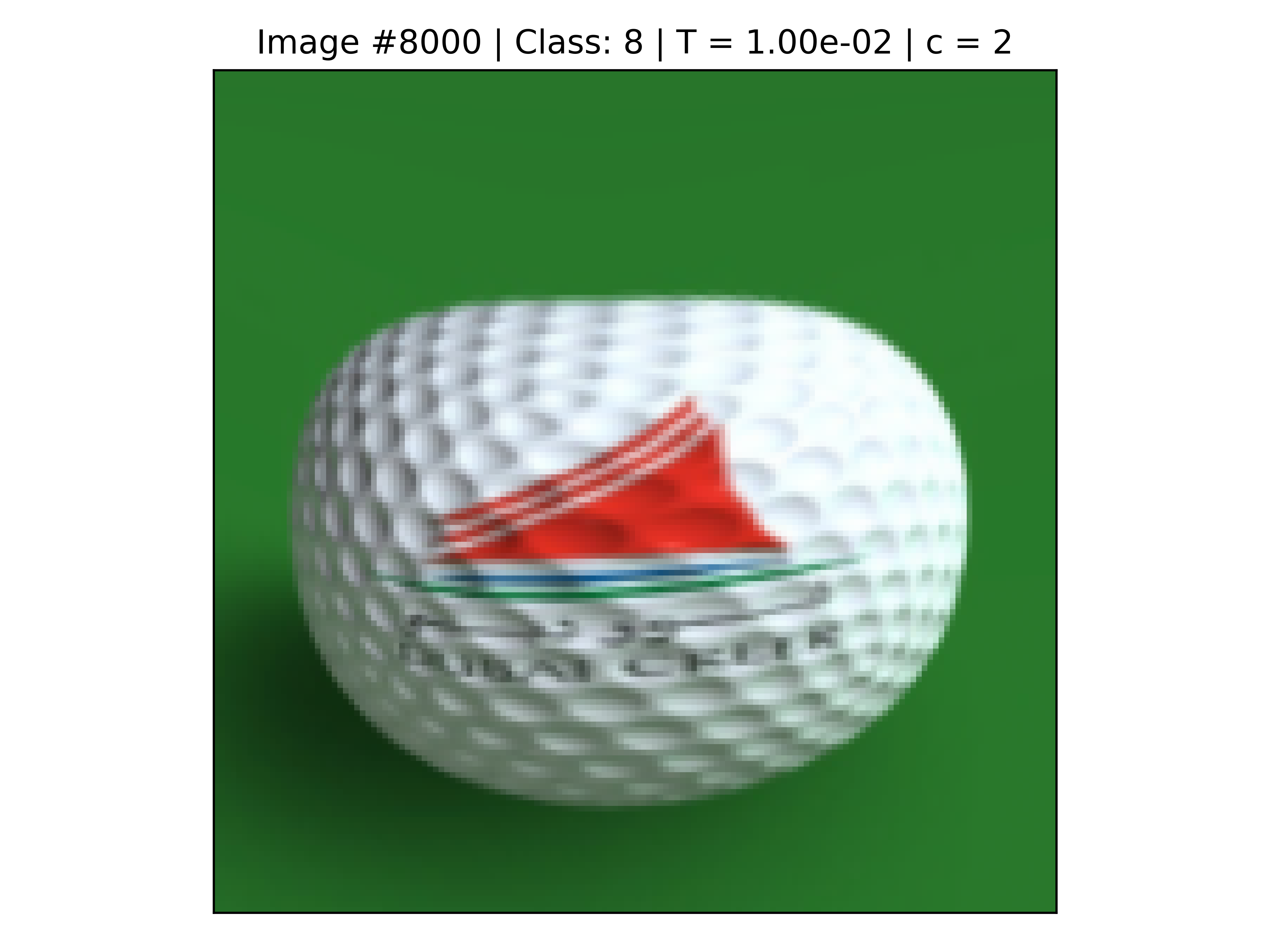}
    \caption{$T=10^{-2}$}
    \end{subfigure}
    \begin{subfigure}{0.18\textwidth}
    \includegraphics[width=\textwidth]{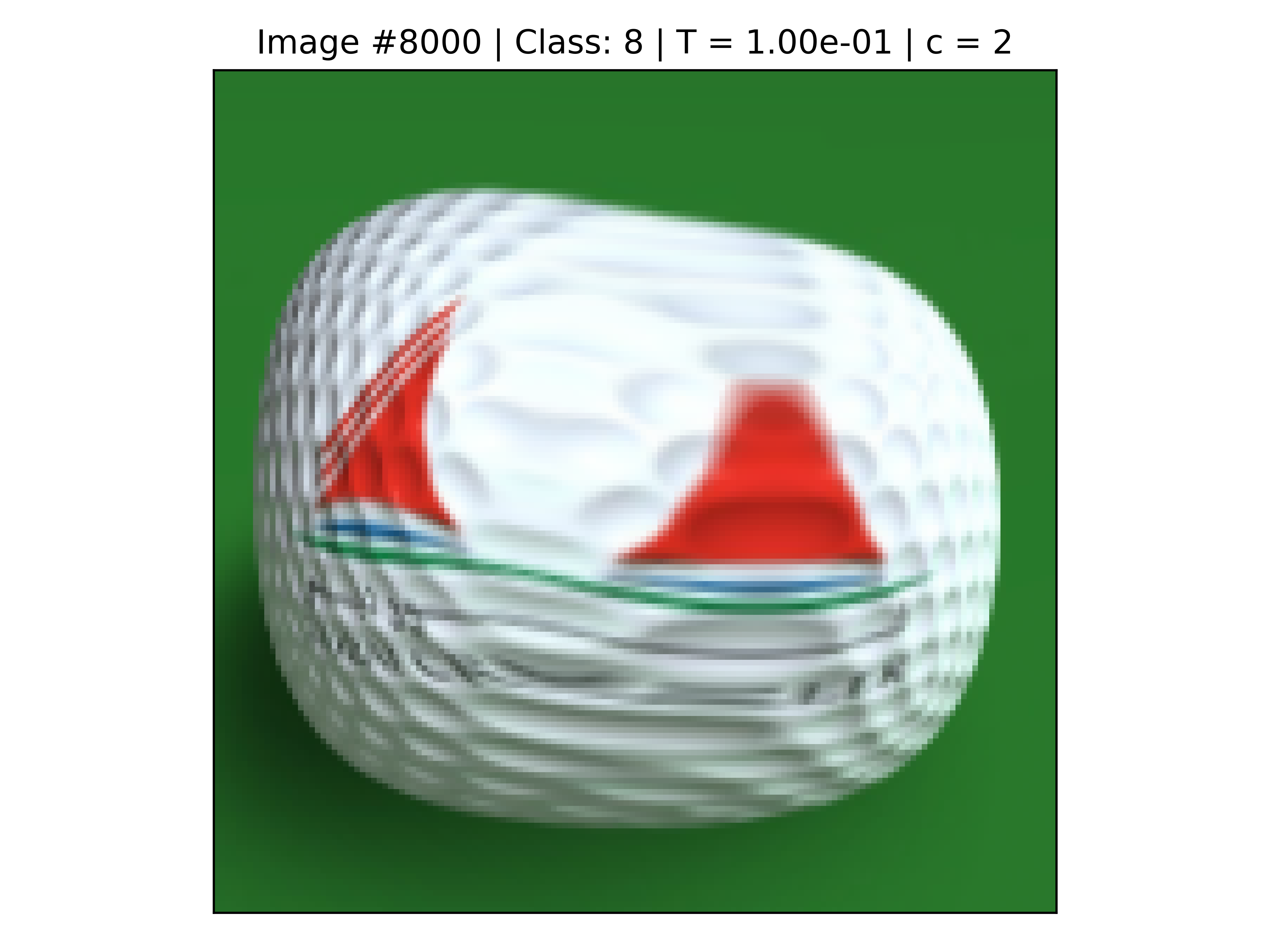}
    \caption{$T=10^{-1}$}
    \end{subfigure}
    \begin{subfigure}{0.18\textwidth}
    \includegraphics[width=\textwidth]{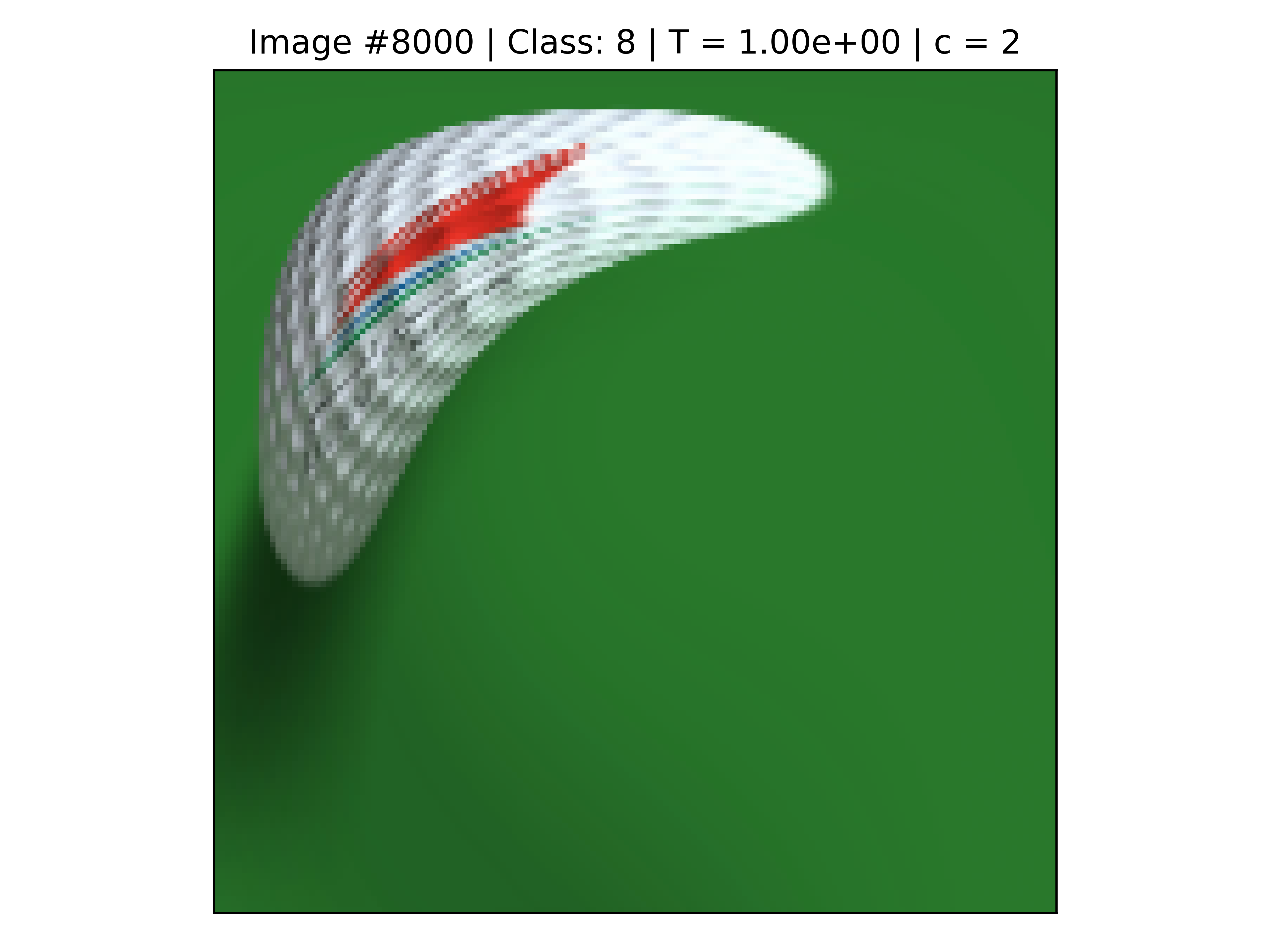}
    \caption{$T=1$}
    \end{subfigure}
    \begin{subfigure}{0.18\textwidth}
    \includegraphics[width=\textwidth]{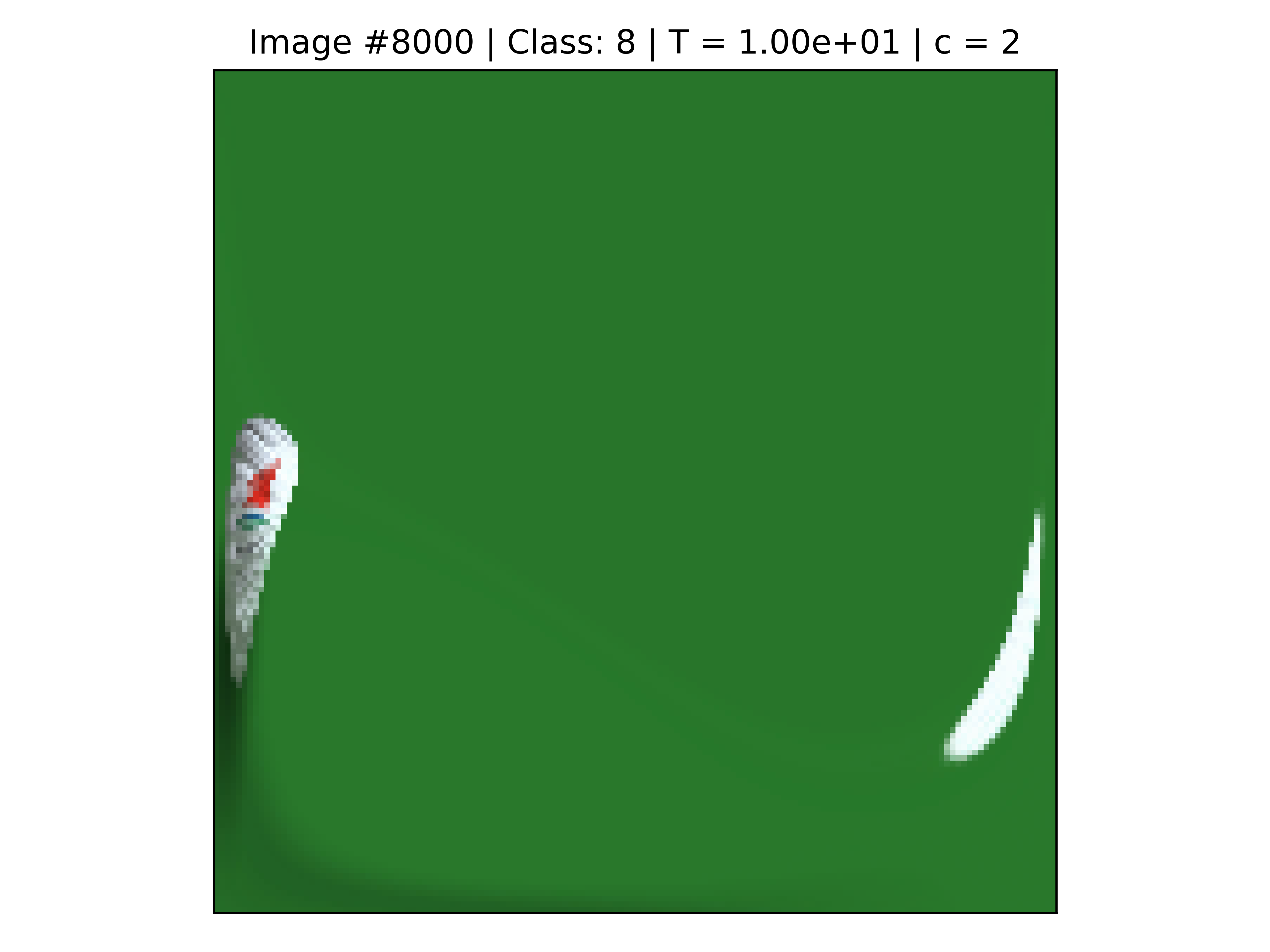}
    \caption{$T=10$}
    \end{subfigure}
    \centering
    \begin{subfigure}{0.18\textwidth}
    \includegraphics[width=\textwidth]{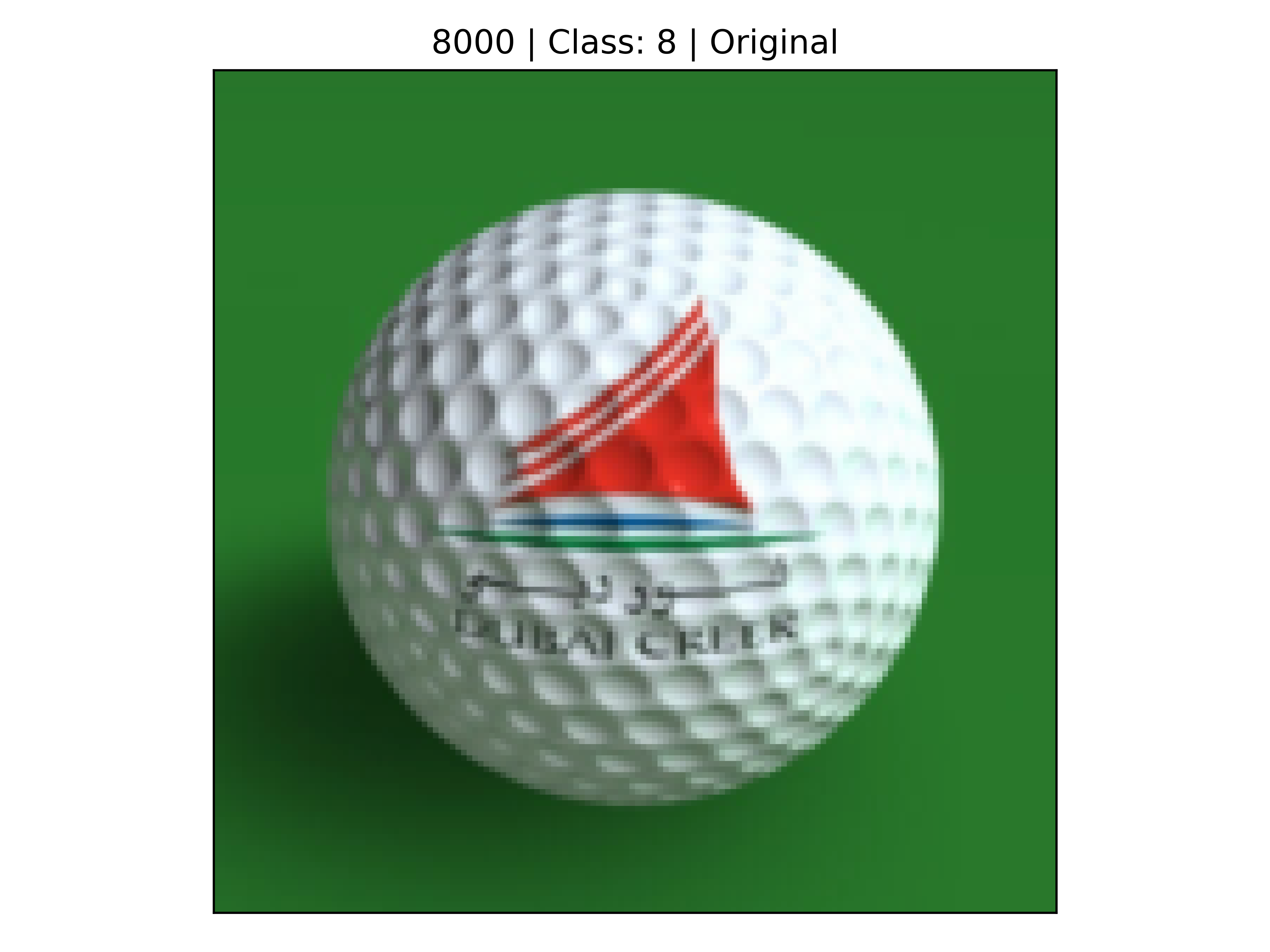}
    \caption{Original}
    \end{subfigure}
    \caption{Image \#8000. Deformations $\text{warp}(f, T)$ for different values of $T$ ($c=2$).}
\end{figure}

\end{document}